\documentclass[Afour,sageh,times]{sagej}

\usepackage{moreverb,url}
\usepackage{pifont} 

\usepackage[hidelinks,colorlinks,bookmarksopen,bookmarksnumbered,citecolor=blue,urlcolor=blue,allcolors=blue,pdfborder={0 0 0}]{hyperref}

\hypersetup{
    colorlinks=true, 
    linkcolor=blue, 
    citecolor=cyan, 
    filecolor=blue, 
    urlcolor=blue,   
    pdfborder={0 0 0} 
}
\usepackage{xcolor}
\colorlet{linkequation}{blue}
\usepackage[colorlinks]{hyperref}
\usepackage{cleveref}

\usepackage{moreverb,url}
\usepackage{xcolor}
\usepackage{color, soul}
\usepackage{caption}
\usepackage{subcaption}
\usepackage{amsmath}
\usepackage{booktabs}
\usepackage{xstring}
\usepackage{hhline}
\usepackage{siunitx}
\usepackage{physics}
\usepackage{tikz}
\usepackage{amssymb,amsfonts}
\usepackage{tcolorbox}
\usepackage{textcomp}
\usepackage{array, multirow,graphicx}
\usepackage{bm}
\usepackage[linesnumbered,ruled,vlined]{algorithm2e} 
\usepackage{dsfont}

\usepackage{todonotes}
\presetkeys{todonotes}{size=\tiny}{}

\newcommand{\argmin}{\mathop{\mathrm{argmin}}}

\def\volumeyear{2024}

\begin{document}

\runninghead{Ji Wu \emph{et al.} QuadricsReg}

\title{QuadricsReg: Large-Scale Point Cloud Registration using Quadric Primitives}

\author{Ji Wu\affilnum{1}, Huai Yu\affilnum{2}, Shu Han\affilnum{1}, Xi-Meng Cai\affilnum{1}, Ming-Feng Wang\affilnum{1}, Wen Yang\affilnum{2}, Gui-Song Xia\affilnum{1}}

\affiliation{%
\affilnum{1}School of Computer Science, Wuhan University, China\\
\affilnum{2}School of Electronic Information, Wuhan University, China
}

\begin{abstract}
In the realm of large-scale point cloud registration, designing a compact symbolic representation is crucial for efficiently processing vast amounts of data, ensuring registration robustness against significant viewpoint variations and occlusions. This paper introduces a novel point cloud registration method, {i.e.}, QuadricsReg, which leverages concise quadrics primitives to represent scenes and utilizes their geometric characteristics to establish correspondences for 6-DoF transformation estimation. As a symbolic feature, the quadric representation fully captures the primary geometric characteristics of scenes, which can efficiently handle the complexity of large-scale point clouds. The intrinsic characteristics of quadrics, such as types and scales, are employed to initialize correspondences. Then we build a multi-level compatibility graph set to find the correspondences using the maximum clique on the geometric consistency between quadrics. Finally, we estimate the 6-DoF transformation using the quadric correspondences, which is further optimized based on the quadric degeneracy-aware distance in a factor graph, ensuring high registration accuracy and robustness against degenerate structures. We test on 5 public datasets and the self-collected heterogeneous dataset across different LiDAR sensors and robot platforms. The exceptional registration success rates and minimal registration errors demonstrate the effectiveness of QuadricsReg in large-scale point cloud registration scenarios. Furthermore, the real-world registration testing on our self-collected heterogeneous dataset shows the robustness and generalization ability of QuadricsReg on different LiDAR sensors and robot platforms. The codes and demos will be released at \url{https://levenberg.github.io/QuadricsReg}.

\end{abstract}

\keywords{Point Cloud Registration, Quadric Representation, Geometric Primitives, Localization and Mapping, Neural-Symbolic AI.}

\maketitle

\section{Introduction} 
\label{sect:intro}

\begin{figure*}[th!]
  \centering
   \includegraphics[width=\linewidth]{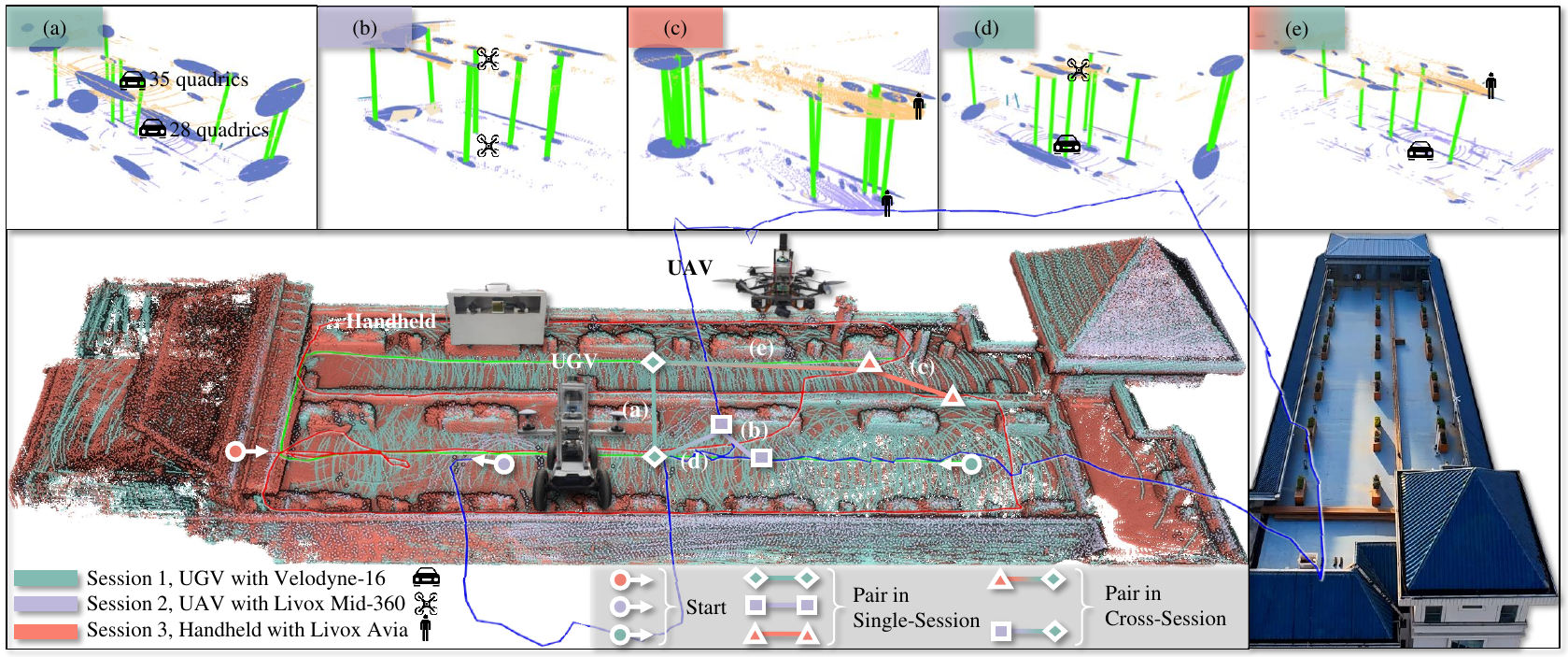}
   \caption{Global point cloud registration using QuadricsReg. The raw point clouds are collected by an Unmanned Ground Vehicle (UGV), Unmanned Aerial Vehicle (UAV), and handheld platform equipped with different LiDAR sensors in a roof garden. Accurate map integration results demonstrate the effectiveness of QuadricsReg.}   
   \label{fig:intro}
\end{figure*}

Point cloud registration is a fundamental task in 3D vision and robotics that involves aligning and integrating multiple 3D point clouds into a common coordinate system \citep{Yin2024}. This alignment is critical for creating a coherent representation of complex environments, which has significant implications for autonomous navigation \citep{zhou2023lidar}, SLAM \citep{Lim2024quatro++}, large-scale 3D mapping \citep{shiratori2015efficient}. By estimating accurate 6-Degrees of Freedom (DoF) transformation between the paired point clouds, the registration enables seamless integration of multiple views or scans, facilitating precise spatial understanding and large-scale 3D reconstructions. However, when dealing with large-scale scenes, the registration accuracy and efficiency are unsatisfactory due to the complexity of scene representation, heterogeneous data collection platforms, and various LiDAR types.

The typical framework for global point cloud registration comprises three key components: Scene representation, feature correspondence establishment, and transformation estimation. Firstly, the sheer volume of points can lead to cumbersome computations in extensive scenarios, therefore a lightweight representation is the fundamental problem for large-scale point cloud registration. Geometric features, such as keypoints \citep{9157132}, lines \citep{rs12010061}, planes \citep{8936527, 8594463}, and cylinders \citep{9787712}, have been widely used in point cloud registration for their representation simplicity. Recent approach \citep{Qiao2024G3reg} leverages the Gaussian Ellipsoid Model (GEM) to represent segmented objects as ellipsoids. However, ellipsoids cannot effectively model different types of primitive uniformly with rich geometric features. Moreover, CAD models can capture the details of known objects in indoor dense settings \citep{Antoni2021Kimera}, yet are inadequate for sparse point cloud representation in complex large-scale scenes. 
Compared to these single shapes, utilizing multiple primitives has the potential to provide a more comprehensive representation. However, different primitives lack a unified mathematical formulation, requiring complex parameter-fitting algorithms for each. Therefore, our primary motivation is to design an elegant model to represent various geometric primitives with high scene expressiveness and compact formulation.

Secondly, observed point clouds are typically partial in large-scale scenes due to varying collection viewpoints or occlusions, which leads to limited overlaps between paired point clouds, making it challenging to achieve robust feature matching. Descriptors based on the local geometric information, whether traditionally handcrafted \citep{wang2023roreg,DONG201861} or learning-based \citep{9577271,9775606}, struggle to effectively represent the complete point cloud features. Consequently, even points belonging to the same region may exhibit low similarity, making it hard to produce reliable feature correspondences. Moreover, outliers are inevitable in the matching process \citep{Chavdar2012Rigid,Yang2021TEASER}, necessitating robust methods that can tolerate geometric feature estimation noise while effectively pruning outliers. As a result, the matching robustness to viewpoint disparities or occlusions is another important issue for point cloud registration.



The final goal of registration is to estimate the 6-DoF transformation between paired point clouds. With the feature correspondences, most existing methods require minimizing a distance metric that measures alignment quality during transformation estimation, as employed by strategies such as RANSAC \citep{Fischler1981Random, Chavdar2012Rigid} and Graduated Non-Convexity (GNC) \citep{Yang2021TEASER, Lim2024quatro++}. However, these metrics are typically constrained by the expressiveness of the scene representation, being limited to measuring single geometric structures like point-to-point distances. Such simplistic metrics overlook the geometric degeneracy within local regions of point clouds, and the remaining outliers lead to unclear geometric structures after registration, such as planes acquiring thickness or becoming curved surfaces \citep{Zhen2022Unified}. Therefore, the last objective is to obtain a robust transformation estimation method and maintain the local fine geometric properties after registration.

Quadrics, as a form of second-order curved surfaces, offer distinct advantages in representing point clouds in a concise and unified manner \citep{Zhen2022Unified, Wu2024QuadricsNet}. They are a natural choice for modeling complex geometries due to their ability to fit a variety of shapes with high accuracy. As a unified symbolic representation of 3D geometry, quadrics have the potential to overcome the challenges mentioned above by providing a robust and comprehensive framework for point cloud analysis. Their symbolic nature allows efficient modeling of scenes, their geometric characteristics enable robust feature matching, and their mathematical properties preserve the geometric shapes after the integration of paired point clouds, making them an attractive option for large-scale point cloud registration tasks.

In this paper, we propose a novel method for global point cloud registration in large-scale scenes with concise quadric primitives, namely \emph{QuadricsReg}. Our approach initially leverages the complete mathematic model of quadrics to extract different geometric primitives and represent large-scale scenes with the primary quadric primitives. Subsequently, we initialize quadric correspondences by comparing the geometric similarity of quadrics and construct a multi-level compatibility graph set to find quadric matches using the maximum clique on geometric consistency between quadric primitives. Finally, we estimate the 6-DoF transformation matrix between paired point clouds using the quadric correspondences. The transformation matrix is further optimized in a factor graph based on the quadric degeneracy-aware distance. An example of global point cloud registration for multi-session mapping using quadrics is shown in Fig. \ref{fig:intro}. This method explores the symbolic quadric representation of 3D scenes and designs the point cloud registration pipeline based on quadrics, which is applied in real-world localization and mapping tasks using heterogeneous robots with different LiDAR sensors.

The contributions of this work are listed as follows:
\begin{itemize}
    \item We propose \emph{QuadricsReg}, a systematic framework for global point cloud registration with quadric primitives, which can efficiently handle large-scale scenes with robustness to viewpoint variations and occlusions. 
    \item We present the novel quadric model, which can concisely represent diverse common geometric primitives using only 10 parameters, having a strong surface fitting ability for lightweight representation of large-scale point clouds.
    \item We build a matching graph for establishing correspondences between quadric primitives, which relies on the intrinsic similarity of quadrics and the geometric consistency between quadrics, having robustness to outliers and efficiency.
    \item We design a novel degeneracy-aware quadric distance to estimate 6-DoF transformation between paired point clouds in a factor graph, which preserves the geometric shapes of primitives after the integration of point clouds.
    \item We comprehensively test \emph{QuadricsReg} on KITTI, KITTI-360, Apollo-SouthBay, Waymo, nuScenes, and self-collected heterogeneous datasets. The exceptional registration success rates and minimal registration errors demonstrate the effectiveness and robustness of \emph{QuadricsReg} on registration of large-scale scenes under efficient quadric representation. 
\end{itemize}

The remainder of this paper is organized as follows. Section \ref{sec2:relatedwork} reviews the related work about scene representation, 3D matching, and 6-DoF transformation estimation. Section \ref{sec3:preliminary} presents a general description of the problem formulation and system overview. Section \ref{sec4:method} details the methodology of point cloud registration based on quadrics. Section \ref{sec5:experiment} illustrates the experimental results and analyses. Finally, section \ref{sec6:conclusion} concludes the paper with future directions.

\section{Related Works}
\label{sec2:relatedwork}
This section reviews the related works for explicit 3D scene representation and global point cloud registration. We mainly discuss three key components for the registration task: Scene representation for LiDAR, 3D correspondence estimation and outlier pruning, and correspondence-based 6-DoF transformation estimation. 
\subsection{Scene Representation for LiDAR}
As a fundamental front-end task, 3D scene representation involves modeling the 3D information of an environment, which provides a crucial foundation for subsequent backend tasks, such as global point cloud registration and object detection. Explicit scene representations can generally be categorized into two types: Low-level spatial representations and high-level geometric primitive-based representations. The low-level spatial representations are directly derived from raw LiDAR data, providing a dense representation that captures detailed spatial information of the environment, typically including points \citep{zhang2014loam,liosam2020shan,Xu2022FAST-LIO2}, meshes \citep{Ruan2023SLAMesh,Lin2023ImMesh}, voxels \citep{hornung2013octomap}, and surfels \citep{chen2019SuMa++}. However, in large-scale outdoor scenarios, as the map incrementally grows over time, the number of candidate frames for registration increases exponentially, resulting in significant computational and storage overheads when employing low-level spatial representations for global registration. 

In contrast, as a robotic localization task, the global point cloud registration primarily relies on a subset of semantically or geometrically distinctive key elements (landmarks) within the scene, rendering much of the information present in dense representations redundant. Recent studies 
\citep{Dubé2017SegMatch,Kong2020Semantic,Cramariuc2021SemSegMap,Yin2023Segregator} compact scene representation by clustering distinct semantic type points to identify segments as key landmarks. CAD models \citep{Antoni2021Kimera} show versatile representation capabilities in small indoor scenes. High-level geometric primitives, such as lines \citep{zuo2017robust,yu2020monocular} and planes \citep{Chen2022VIP-SLAM}, can effectively represent structures of a scene. More complex primitives, such as cuboids \citep{Yang2019CubeSLAM} and ellipsoids \citep{nicholson2018quadricslam,Qiao2024G3reg}, combined with semantic information, have recently been employed in object-based SLAM. These high-level geometric primitives have proven to efficiently represent the distribution of key objects in the scene in a lightweight, vectorized manner. Nevertheless, for large-scale scenes characterized by complex and heterogeneous structures, a single primitive-based representation is insufficient to capture all the geometric intricacies comprehensively \citep{Wu2024QuadricsNet}. Although hybrid representations incorporating multiple types of geometric primitives offer enhanced geometric diversity, extracting and fitting multiple parameterized models necessitate different fitting procedures, significantly increasing algorithmic complexity \citep{schnabel2007efficient,Lingxiao2019Supervised,Gopal2020ParSeNet}. Thus, a balance between expressiveness and simplicity in scene representation must be sought, as this trade-off substantially influences the precision and efficiency of point cloud registration.

Quadrics can represent diverse common geometric primitives using only 10 parameters, demonstrating their expressiveness for LiDAR-based 3D structured modeling \citep{Wu2024QuadricsNet} and SLAM \citep{Zhen2022Unified,Xia2023Quadric} applications. We extend this concept to large-scale scenes by designing a unified and concise representation for multiple geometric primitives. Building on this representation, we further develop a global registration framework to enhance the robustness and efficiency of scene alignment.

\subsection{3D Correspondence Establishment and Outlier Pruning}
Global point cloud registration methods can be broadly categorized into correspondence-based and correspondence-free methods. This work focuses on the former, which are commonly applied in large-scale scenes. These methods first detect key elements and establish initial correspondences by comparing descriptors between point clouds. These initial correspondences inevitably contain outliers, which need to be pruned before solving the transformation. Typically, at least three valid correspondences are required, otherwise matching degeneracy may occur \citep{Lim2024quatro++}.

Key elements in a scene can be represented by key points, with their descriptors derived from local geometric features. 
USC \citep{Tombari2010Unique} employs covariance matrices of point pairs, while SHOT \citep{salti2014shot} constructs a 3D histogram of normal vectors. PFH \citep{rusu2008aligning} and FPFH \citep{Rusu2009FPFH} create oriented histograms based on pairwise geometric properties. Deep learning methods such as FCGF \citep{Choy2019FCGF} and D3Feat \citep{Bai2020D3Feat} have been explored to extract dense geometric features and predict key point scores, thereby enhancing the identification of matchable points. 
However, in cases of significant viewpoint variations or occlusions, low overlap between point clouds makes reliable correspondence challenging when relying on local descriptors at the point level, even for identical parts of the scene. Furthermore, point-level features can exhibit representational redundancy in global registration tasks. Thus, descriptors based on high-level representations have been developed. SegMatch \citep{Dubé2017SegMatch} retains only key segments and computes feature values and histograms at the segment level. BoxGraph \citep{Pramatarov2022BoxGraph} models objects of the scene using bounding boxes and establishes correspondences based on axis-aligned elongation. G3Reg \citep{Qiao2024G3reg} represents key objects with Gaussian models, employing the Wasserstein distance for similarity measurement. However, these descriptors rely on statistical analysis from the raw point clouds to derive attributes such as centroids and sizes. When the overlap between point clouds is limited, partial observations can make it challenging to estimate accurate geometric features, leading to unreliable similarity measures and inaccuracies in pose estimation.

Outliers must be rejected to ensure registration success, as most of the initial correspondences can be erroneous. RANSAC \citep{Fischler1981Random,Chavdar2012Rigid} and its variants repeatedly sample correspondences, generating and evaluating geometric models until a satisfactory solution is found. GNC \citep{Yang2020Graduated} simultaneously estimates the pose and rejects outliers. Graph-based methods \citep{lusk2021clipper,Zhang20233DMAC,Yang2024MAC} have recently gained prominence due to their robustness, even in the presence of high outlier rates. These methods filter correspondences using geometric consistency checks, construct a compatibility graph, and identify inliers through maximum clique detection. TEASER \citep{Yang2021TEASER} utilizes length invariance under transformation to establish check criteria. Segregator \citep{Yin2023Segregator} models the invariance by a Gaussian distribution and performs checks based on discrepancies between these distributions. Nevertheless, the optimal threshold for geometric consistency checks depends on the uncertainty boundaries of matching point distributions, which are challenging to quantify. The multi-level threshold strategies  \citep{Qiao2023Pyramid,Qiao2024G3reg} address this by constructing a pyramid compatibility graph, ultimately selecting the optimal transformation by evaluating the transformations from each level.

We employ a complementary approach combining learning-based and statistics-based methods to fit high-level quadrics from partial point clouds, ensuring reliable estimation of geometric attributes. Using these estimations, we establish correspondences to achieve geometric consistency under significant viewpoint variations. Additionally, a multi-level compatibility graph strategy accounts for estimation uncertainty, ensuring sufficient correspondences and enhancing the robustness of the registration algorithm.

\subsection{Correspondence-based 6-DoF Transformation Estimation}
Estimating the transformation between point clouds is the primary objective of point cloud registration. It is demonstrated that a closed-form solution for the 6-DoF transformation between two point clouds exists when correspondences are known, and the points are affected by isotropic zero-mean Gaussian noise \citep{Horn1987Closed,Arun1987Least}. However, even after correspondence pruning, it is challenging to guarantee the correctness of all established correspondences. The commonly used robust methods for handling mismatches include RANSAC-based approaches \citep{Fischler1981Random,Chavdar2012Rigid}, which are effective in scenarios with low outlier ratios, and GNC-based methods \citep{zhou2016fast,Yang2020Graduated,Yang2021TEASER,Lim2024quatro++}, which improve robustness under higher outlier ratios. Branch-and-bound (BnB)-based algorithms offer guaranteed theoretical optimality but are often too slow for practical use \citep{Olsson2009Branch}. Additionally, data-driven methods \citep{Choy2020DGR,Bai2021Pointdsc,Cattaneo2022LCDNet} leverage deep networks to directly estimate the transformation in an end-to-end manner, typically relying on learned features. 

Typically, transformation estimation methods require an error function to evaluate the alignment quality of the estimated transformation, generally based on distances between corresponding points, as such metrics constrain the transformation in three directions. In large-scale environments, certain regions may lack sufficient geometric features, such as those predominantly composed of planar, cylindrical, or spherical structures \citep{Chen2020SLOAM,Zhen2022Unified,Chen2022VIP-SLAM}. These geometric structures exhibit degeneracy in specific directions, resulting in insufficient constraints. Thus, relying on conventional point-to-point metrics during registration can lead to substantial errors in alignment. Prior research has proposed using point-to-point \citep{Besl1992ICP}, point-to-line \citep{Censi2008point-to-line}, and point-to-plane \citep{low2004linear} distances as error functions, each offering directionally specific constraints suitable for different geometric configurations. However, these transformation estimation algorithms are typically decoupled from the scene representation, often requiring additional feature detection methods, such as smoothness analysis \citep{zhang2014loam}, to determine geometric types. The independent use of these error metrics results in a complex optimization process, potentially increasing computational overhead.

Based on the quadric representation of scenes, we develop a unified distance metric and optimization strategy, which addresses geometric degeneracy to reduce registration errors. Additionally, we validate and select the optimal transformation from multiple candidates using this metric, effectively suppressing the impact of potential correspondence outliers.
\section{Problem Formulation}
\begin{figure*}[h]
  \centering
   \includegraphics[width=\linewidth]{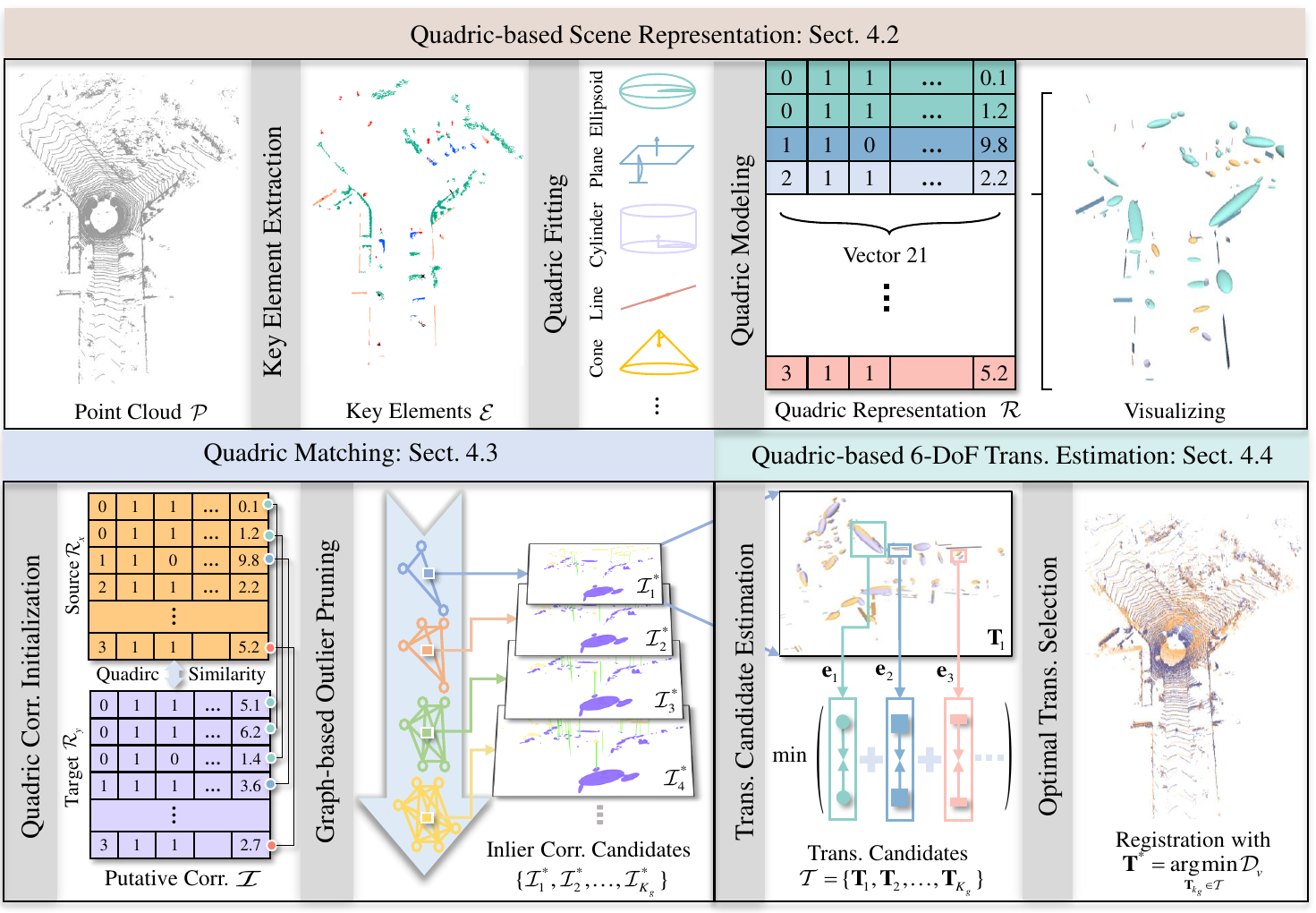}
   \caption{The pipeline of QuadricsReg mainly consists of three parts: Quadric-based scene representation, quadric matching, and quadric-based 6-DoF transformation estimation.
   }   
   \label{fig:pipeline}
\end{figure*}

\label{sec3:preliminary}
\subsection{Problem Definition}
Let $\mathcal{P}_x$ and $\mathcal{P}_y$ represent two partially overlapping 3D LiDAR point clouds captured from different viewpoints of a scene. Global registration aims to estimate the optimal relative rigid transformation $\mathbf{T}^* \in SE(3)$, consisting of a rotation matrix $\mathbf{R}^* \in SO(3)$ and a translation vector $\mathbf{t}^* \in \mathbb{R}^3$, to align the source point cloud $\mathcal{P}_x$ with the target point cloud $\mathcal{P}_y$, without requiring an initial guess. 

We focus on a correspondence-based global registration method, which can be divided into three steps. The first step is the scene representation with symbols or elements, where key elements $\{\mathbf{e}_i\}$ have distinctive features.  $\mathcal{E}_x = \{\mathbf{e}^1_x, \mathbf{e}^2_x, \dots, \mathbf{e}^{N_x}_x\}$ and $\mathcal{E}_y = \{\mathbf{e}^1_y, \mathbf{e}^2_y, \dots, \mathbf{e}^{N_y}_y\}$ are extracted elements (\emph{e.g.}, edge points, lines, planes and objects) from the point cloud $\mathcal{P}_x$ and $\mathcal{P}_y$, respectively. The scene representations are then generated from key elements, denoted as $\mathcal{R}_x = \{\mathbf{r}_x^1, \mathbf{r}_x^2, \dots, \mathbf{r}_x^{N_x}\}$ and $\mathcal{R}_y = \{\mathbf{r}_y^1, \mathbf{r}_y^2, \dots, \mathbf{r}_y^{N_y}\}$, where $\mathbf{r}$ is the mathematical representation of an element $\mathbf{e}$. The second step involves establishing correspondences by comparing the similarities between the scene representations, forming an initial correspondence set between $\mathcal{R}_x$ and $\mathcal{R}_y$. Outliers are then pruned using a geometric consistency check, resulting in an inlier set $\mathcal{I}^*$. The third step is the transformation estimation, where the initial transformation is estimated based on $\mathcal{I}^*$, followed by a more detailed transformation optimization. 

The global registration can be formulated as an optimization problem, where the objective is to minimize the distance between corresponding element representations in $\mathcal{R}_x$ and $\mathcal{R}_y$. It can be expressed as
\begin{equation}
\label{eq: formulation}
    \mathbf{R}^*, \mathbf{t}^* = 
    \argmin_{\mathbf{R} \in \mathrm{SO}(3), \mathbf{t}\in \mathbb{R}^3} \sum_{\left(\mathbf{r}_x^k, \mathbf{r}_y^k\right) \in \mathcal{I}^*} \rho\left( e\left( f_t\left( \mathbf{r}_x^k \mid \mathbf{R}, \mathbf{t} \right),\mathbf{r}_y^k \right) \right),
\end{equation}
where $\rho \left( \cdot \right)$ is the robust kernel function, $e\left( \cdot \right)$ is the error function, $f_t\left( \cdot \right)$ denotes the rigid transformation.
\subsection{System Overview}

The pipeline of our proposed global point cloud registration framework, QuadricsReg, is illustrated in Fig. \ref{fig:pipeline}. It comprises three steps: Quadric-based scene representation, quadric matching, and quadric-based transformation estimation. Given the paired point clouds $\mathcal{P}_x$ and $\mathcal{P}_y$, we first extract distinctive primitives and fit them as quadrics. Subsequent registration between $\mathcal{P}_x$ and $\mathcal{P}_y$ is conducted based on their concise quadric representations, $\mathcal{R}_x$ and $\mathcal{R}_y$. Initial correspondences $\mathcal{I}$ are established based on the proposed similarity metric for quadrics. Multi-threshold consistency checks are then applied to construct multi-level compatibility graphs for outlier pruning among the correspondences. The maximum clique from each graph level is identified to generate inlier correspondence candidates $\{\mathcal{I}_1^*,\mathcal{I}_2^*,\dots,\mathcal{I}_{K_g}^*\}$. For each correspondence candidate, an initial transformation is computed using Singular Value Decomposition (SVD). A degeneracy-aware distance metric is subsequently designed to further refine the transformation, resulting in transformation candidates $\mathcal{T} = \{ \mathbf{T}_1, \mathbf{T}_2, \dots, \mathbf{T}_{K_g}\}$. 
Finally, the transformation with the minimum distance between $\mathcal{R}_x$ and $\mathcal{R}_y$ after registration is selected as the optimal transformation $\mathbf{T}^*$.
\section{Methodology}
\label{sec4:method}
\subsection{Quadrics}
\label{sect: Quadrics}
In this section, we begin by defining the quadrics, a parametric representation of surfaces that offers a compact mathematical form capable of representing various common primitive types. We then decompose the quadrics to clarify the geometric meaning of the components in their compact representations. Following that, we provide a detailed explanation of how to infer geometric attributes (type, scale, and pose) from the mathematical representation. Finally, we discuss the degeneracy in quadrics.

\subsubsection{Quadrics Basics.}
Quadrics are a class of surfaces defined implicitly by a second-degree polynomial equation:
\begin{equation}
\begin{aligned}
  f_q(\mathbf{x,q})=&Ax^2+By^2+Cz^2+2Dxy+2Exz\\
  &+2Fyz+2Gx+2Hy+2Iz+J=0,
\end{aligned}
\label{eq: quadrics equation}
\end{equation}
where $\mathbf{x}$ represents a 3D point with homogeneous coordinate $[x,y,z,1]^{\mathrm{T}}$, $\mathbf{q}=[A,B,C,D,E,F,G,H,I,J]$ and the quadratic term is not all zero. The compact matrix form is $\mathbf{x}^{\mathrm{T}}\mathbf{Q}\mathbf{x}=0$, where
\begin{equation}
\setlength{\arraycolsep}{4.5pt}
\mathbf{Q}=\left[\begin{array}{@{}cccc@{}}
A & D & E & G \\
D & B & F & H \\
E & F & C & I \\
G & H & I & J
\end{array}\right],\nabla\mathbf{Q}=2\left[\begin{array}{@{}cccc@{}}
A & D & E & G \\
D & B & F & H \\
E & F & C & I
\end{array}\right].
\label{eq: Q and normal}
\end{equation}
The gradient at $\mathbf{x}$ is given by $\nabla f_q(\mathbf{x,q})=\nabla\mathbf{Q}\mathbf{x}$, which also represents the direction of its normal $\mathbf{n} \in \mathbb{R}^{3} $.

Despite being defined by just 10 parameters, quadrics can uniformly represent 17 geometric primitives, such as points, lines, planes, spheres, cylinders, and cones, covering the majority of objects and structures in urban environments.

\begin{figure}[t]
  \centering
   \includegraphics[width=\linewidth]{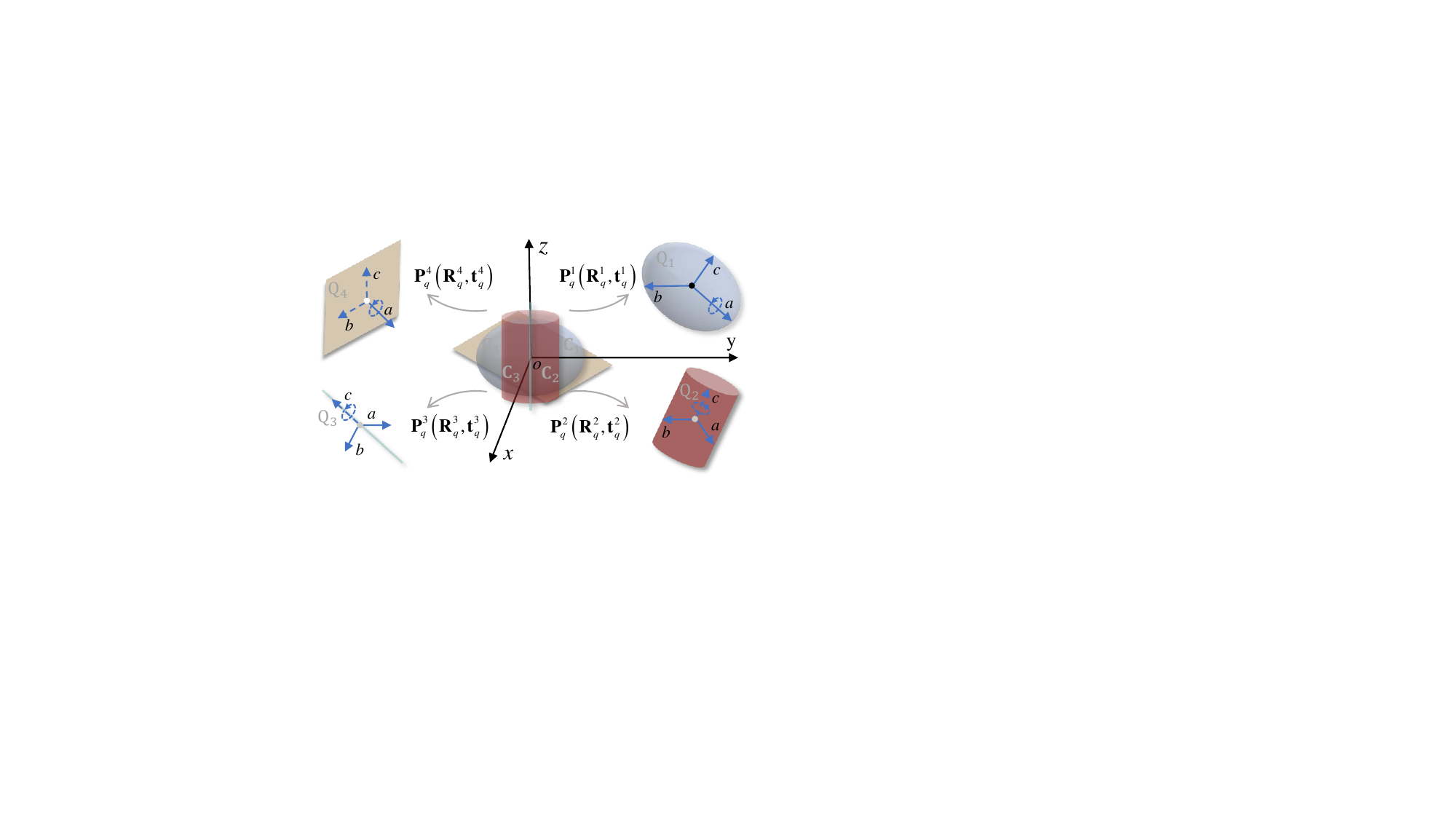}
   \caption{Illustration of quadric derivation and degeneracy. Ellipsoid $\mathbf{Q}_{1}$, cylinder $\mathbf{Q}_{2}$, line $\mathbf{Q}_{3}$, and plane $\mathbf{Q}_{4}$ are transformed from canonical quadrics $\mathbf{C}_{1}$, $\mathbf{C}_{2}$, $\mathbf{C}_{3}$, and $\mathbf{C}_{4}$. For $\mathbf{Q}_{1}$, rotation around axis $a$ is degenerate due to symmetry along axes $b$ and $c$. For $\mathbf{Q}_{2}$, scale and translation along axis $c$, and rotation around it, are degenerate due to openness along $c$ and symmetry on axes $a$, $b$. The other quadrics can be similarly analyzed.
   }   
   \label{fig: illustration of quadrics derivation and degeneracy}
\end{figure}
\subsubsection{Quadric Decomposition.}
\label{sect: Quadrics Decomposition}
As shown in the Fig. \ref{fig: illustration of quadrics derivation and degeneracy}, in 3D space, any quadric $\mathbf{Q}$ can be considered as the result of applying a rigid transformation to a canonical quadric $\mathbf{C}$, whose axes are aligned with the coordinate axes. This process can be formulated as
\begin{equation}
\mathbf{Q} = \mathbf{P}_q^{-\mathrm{T}} \mathbf{C} \mathbf{P}_q^{-1} =
\left[ \begin{smallmatrix}
\mathbf{R}_q & \mathbf{t}_q \\
\mathbf{0}^\mathrm{T} & 1
\end{smallmatrix} \right]^{-\mathrm{T}}
\left[ \begin{smallmatrix}
\mathbf{\Lambda}_q & \mathbf{0} \\
\mathbf{0}^\mathrm{T} & c_{44}
\end{smallmatrix} \right]
\left[ \begin{smallmatrix}
\mathbf{R}_q & \mathbf{t}_q \\
\mathbf{0}^\mathrm{T} & 1
\end{smallmatrix} \right]^{-1},
\label{eq: Q=PCP}
\end{equation}
where $\mathbf{P}_q(\mathbf{R}_q, \mathbf{t}_q) \in SE(3)$ represents the transformation matrix from $\mathbf{C}$ to $\mathbf{Q}$, describing the pose of a quadric.  $\mathbf{R}_q \in SO(3)$ and $\mathbf{t}_q \in \mathbb{R}^{3}$ are the rotation and center (translation) blocks of $\mathbf{P}_q$. For the canonical quadric, $\mathbf{Q}$ is reduced to a diagonal matrix
$\mathbf{C}$. The diagonal blocks are given by $\mathbf{\Lambda}_q \in \mathbb{R}^{3\times 3}$ and $c_{44}$, where the values of $\mathbf{\Lambda}_q$ determine the scale of a quadric. More explicitly, $\mathbf{Q}$ can be decomposed as
\begin{equation}\label{eq: quadrics further expanded}
\mathbf{Q} 
 =\left[\begin{array}{@{}cc@{}}
\mathbf{R}_q \mathbf{\Lambda}_q \mathbf{R}_q^{\mathrm{T}} & -\mathbf{R}_q \mathbf{\Lambda}_q \mathbf{R}_q^{\mathrm{T}}\mathbf{t}_q \\
* & k
\end{array}\right]=\left[\begin{array}{@{}cc@{}}
\mathbf{Q}_{33} & \mathbf{l} \\
* & k
\end{array}\right].
\end{equation}

Although the quadric representation is highly compact, its decomposition shows that $\mathbf{Q}$ can be formed by matrices with clear geometric meanings, which is crucial for robust estimation of its parameters.

\subsubsection{Quadric Geometry Analysis.}
\label{sect: Quadrics Geometry Analysis}
The geometric attributes of a quadric, including type, scale, and pose, can be inferred from its mathematical representation $\mathbf{Q}$. In Eq. \ref{eq: Q=PCP}, the matrix $\mathbf{C}$ form $\mathbf{I_C} \in \{0,1,-1\}^{4}$ determines the type of a quadric, which is illustrated in Table \ref{tab: Characteristics of Important Quadrics}. According to Eq. \ref{eq: quadrics further expanded}, the scale and pose can be inferred through mathematical analysis of $\mathbf{Q}_{33}$. 

Before performing inference, $\mathbf{Q}$ has to be normalized to resolve the proportional ambiguity in Eq. \ref{eq: quadrics equation}:
\begin{equation}
{\mathbf{Q}} = \left\{ \begin{array}{cc}
\left| {\dfrac{{\prod {\lambda _i^{\mathbf{Q}_{33}}} }}{{\prod {\lambda _i^\mathbf{Q} }}}} \right|{\mathbf{Q}}, \ &{c_{44}} \ne 0\\
\dfrac{1}{\left \| \mathbf{Q} \right \| }{\mathbf{Q}}, \ &{c_{44}} = 0
\end{array} ,\right.
\end{equation}
where $\lambda^{\mathbf{Q}_{33}}$ and $\lambda^{\mathbf{Q}}$ are the non-zero eigenvalues of $\mathbf{Q}_{33}$ and $\mathbf{Q}$. According to Eq. \ref{eq: quadrics further expanded}, the symmetric matrix $\mathbf{Q}_{33}$ can be diagonalized as $\mathbf{Q}_{33} = \hat{\mathbf{R}}_q \hat{\mathbf{\Lambda}}_q \hat{\mathbf{R}}_q^{\mathrm{T}}$. The diagonal matrix $\hat{\mathbf{\Lambda}}_q = \mathbf{\Lambda}_q$ consists of the eigenvalues of $\mathbf{Q}_{33}$, given as $\mathrm{diag}(\lambda_a, \lambda_b, \lambda_c)$, and $\hat{\mathbf{R}}_q$ contains the corresponding eigenvectors. Without loss of generality, assume that $\lambda_a \geq \lambda_b \geq \lambda_c$, and neither is equal to zero. The scale $\mathbf{s}_q \in \mathbb{R}^{3} $, rotation $\mathbf{R}_q$ and center $\mathbf{t}_q$ can be inferred as
\begin{equation}
\left\{ 
\renewcommand{\arraystretch}{1.5}
\begin{array}{lll}
\left [s^a_q,s^b_q,s^c_q\right ]^{\mathrm{T}}&=&\mathrm{diag}(\mathbf{I_s})\sqrt{\left | \left [ \frac{1}{\lambda_a} ,\frac{1}{\lambda_b},\frac{1}{\lambda_c} \right ]\right | ^{\mathrm{T}}},\\
\left[ \mathbf{r}^a_q, \mathbf{r}^b_q, \mathbf{r}^c_q \right] &=& \pm \hat{\mathbf{R}}_q \, \mathrm{diag} (\mathbf{I}_\mathbf{R}),
\\
\left [ t^a_q,t^b_q,t^c_q\right ] ^{\mathrm{T}} &=& \mathrm{diag}(\mathbf{I_t}){\mathbf{t}_q},
\end{array} \right.
\end{equation}
where $\mathbf{I}_{\mathbf{s,R,t}} \in \{0,1\}^{3}$ indicate the degeneracy of the scale, rotation, and translation. If the elements in $\mathbf{s}$ are unequal, the quadric becomes asymmetric, such as an ellipsoid, elliptic cylinder, or elliptic cone. The direction of $\mathbf{r}_q$ can either be identical or opposite to the column of $\hat{\mathbf{R}}_q$. The center $\mathbf{t}_q$ can be inferred by solving: 
\begin{equation}
\label{eq: solve t}
\mathbf{Q}_{33} \mathbf{t}_q + \mathbf{l}= \mathbf{0}.
\end{equation}
The ranks of the coefficient matrix $\mathbf{Q}_{33}$ and augmented matrix $\left[ \mathbf{Q}_{33} \ | \ \mathbf{l} \right]$, which can be 3, 2, or 1, determine the case of solution space, classifying the quadric as central (points, spheres, cones), linear-center (lines, cylinders), or planar-center (planes).

\begin{figure*}[th!]
  \centering
   \includegraphics[width=\linewidth]{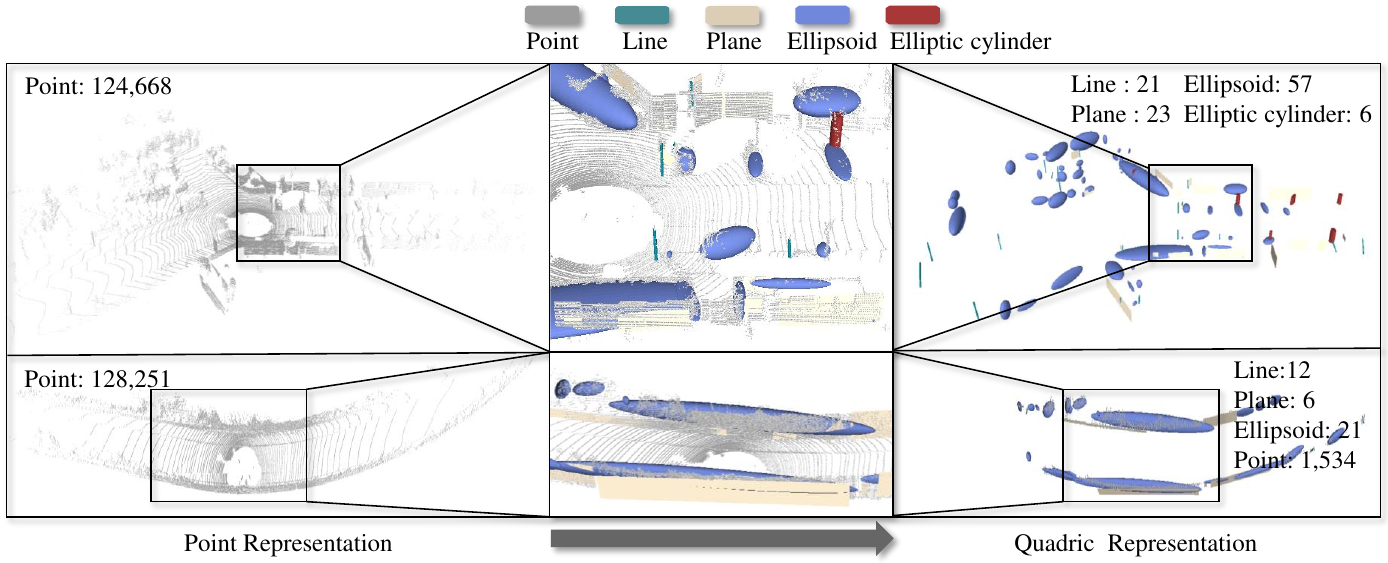}
   \caption{Quadric-based representation for LiDAR scans from KITTI dataset. The point scans are represented compactly using quadrics. Road structures like the ground, buildings, and poles can be modeled as planes and lines, while objects such as vehicles and trunks are treated as ellipsoids and elliptic cylinders. Additionally, we find that modeling vegetation as ellipsoids can effectively enrich scene representation. The top scene has an adequate number of key elements, whereas the bottom scene features extended and repetitive elements in a sparse arrangement. 
   } 
   \label{fig: Illustration of quadrics-based scene representation}
\end{figure*}

\subsubsection{Quadric Degeneracy.}
\label{sect: Degeneracy of Quadrics}

Quadric degeneracy means that scaling, rotating, and translating operations on a quadric will not affect its shape, which can be judged from $\mathbf{Q}$. If $\mathbf{Q}$ is rank-deficient, this will lead to the degeneracy of scale at certain axes. The zero values in the eigenvalues $\hat{\mathbf{\Lambda}}_q$ of $\mathbf{Q}_{33}$ make the translation along the corresponding axis degenerate, and the number of such zero values determines the type of quadric center according to Eq. \ref{eq: solve t}. If there are duplicates in $\hat{\mathbf{\Lambda}}_q$, \textit{i.e.}, the quadric is symmetric, and the rotation around the non-symmetric axis will degenerate. Fig. \ref{fig: illustration of quadrics derivation and degeneracy} illustrates this more intuitively.

The ability to describe degeneracy is an advantage of quadric representation. During transformation optimization in Eq. \ref{eq: formulation}, errors of degenerate attributes should be eliminated. For example, when registering planar scenes, only errors along the normal direction should be considered by setting the point-to-plane distance.


\begin{table}[h]
    \centering
    \scriptsize  
    \setlength{\tabcolsep}{5.5pt}  
    \caption{Characteristics of typical quadrics. Negative signs indicate negative values.}
    \label{tab: Characteristics of Important Quadrics}
    \begin{tabular}{@{}llcccc@{}}
        \toprule
        Type & $\mathrm{Diag}(\mathbf{C})$ & $\mathbf{I_C}$ & $\mathbf{I_s}$ & $\mathbf{I_R}$ & $\mathbf{I_t}$ \\ \midrule
        Point    & $[\lambda_a, \lambda_b, \lambda_c, 0]$       & $[1, 1, 1, 0]$ & $[0, 0, 0]$ & $[0, 0, 0]$ & $[1, 1, 1]$ \\
        Line     & $[\lambda_a, \lambda_b, 0, 0]$               & $[1, 1, 0, 0]$ & $[0, 0, 0]$ & $[0, 0, 1]$ & $[1, 1, 0]$ \\
        Plane    & $[\lambda_a, 0, 0, 0]$                       & $[1, 0, 0, 0]$ & $[0, 0, 0]$ & $[1, 0, 0]$ & $[1, 0, 0]$ \\
        Sphere   & $[\lambda_a, \lambda_b, \lambda_c, -1]$      & $[1, 1, 1, -1]$ & $[1, 1, 1]$ & $[0, 0, 0]$ & $[1, 1, 1]$ \\
        Cylinder & $[\lambda_a, \lambda_b, 0, -1]$              & $[1, 1, 0, -1]$ & $[1, 1, 0]$ & $[0, 0, 1]$ & $[1, 1, 0]$ \\
        Cone     & $[\lambda_a, \lambda_b, -\lambda_c, 0]$      & $[1, 1, -1, 0]$ & $[1, 1, 0]$ & $[0, 0, 1]$ & $[1, 1, 1]$ \\ 
        \bottomrule
    \end{tabular}
\end{table}

\subsection{Quadric-based Scene Representation}
In this section, we present an approach for reducing scene modeling from the dense point cloud into a quadric representation. First, we use semantic or geometric information to extract key elements from the scene, including key structures and objects. We then fit these key elements with quadrics to estimate their quadric parameters and geometric attributes. Finally, based on the extraction and fitting results, we construct a quadric-based scene representation, describing each element's semantic, scale, pose, and quadric degeneracy with only 21 parameters.

\subsubsection{Quadric Extraction.}
Human scene processing significantly relies on the semantics, geometrics, and relationships of key elements in the scene, such as key structures and objects. As illustrated in Fig. \ref{fig: Illustration of quadrics-based scene representation}, they are typically stable, distinctive, and common in scenes. These key elements are extracted as quadrics, with each fitted to a specific quadric type based on its geometric characteristics. 

Key elements are extracted from point cloud $\mathcal{P}$ as follows:
\begin{enumerate} 
    \item \textbf{Key structure extraction}: The key structures contains the ground $\mathcal{E}_g$, planes $\mathcal{E}_p=\{\mathbf{e}^1_p, \mathbf{e}^2_p, \dots, \mathbf{e}^{K_p}_p\}$, and lines $\mathcal{E}_l=\{\mathbf{e}^1_l, \mathbf{e}^2_l, \dots, \mathbf{e}^{K_l}_l\}$.  We refer to the structure extraction process of TRAVEL \citep{Oh2022TRAVEL} and G3Reg \citep{Qiao2024G3reg}: 
        \begin{enumerate}
            \item Ground extraction: The ground $\mathcal{E}_g$ is extracted by TRAVEL and labeled with ground semantics $\mathcal{L}_g$. 
            \item Plane extraction: The remaining point cloud $\mathcal{P} \setminus \mathcal{E}_g$ is voxelized, followed by eigenvalue analysis. Planes are identified based on the ratio of the smallest to second smallest eigenvalue for points in each voxel, and adjacent plane voxels are then merged using region growing. The results $\mathcal{E}_p$ are labeled with the plane semantics $\mathcal{L}_p$. 
            \item Line extraction: The remaining point cloud $\mathcal{P} \setminus (\mathcal{E}_g \cup \mathcal{E}_p)$ is segmented into clusters by TRAVEL, and then RANSAC is applied to detect lines $\mathcal{E}_l$, which are labeled with the line semantics $\mathcal{L}_l$. 
        \end{enumerate} 
    \item \textbf{Key object extraction}: Extracting key objects in the point cloud with distinct semantics (vehicles, trunks, and vegetation) is efficient, but semantic labels may be difficult to predict for some complex fields. If the semantic labels $\mathcal{L}_s$ are reliable and comprehensive, clustering is performed in each specified semantics, including vehicles, trunks, and vegetation, represented by $\mathcal{L}_{o}=\{\mathcal{L}_{s_1}, \mathcal{L}_{s_2}, \dots, \mathcal{L}_{S_{K_s}} \mid \mathcal{L}_{s_k} \subseteq \mathcal{L}_s\}$, to obtain key objects $\mathcal{E}_o = \{\mathcal{E}_{s_1}, \mathcal{E}_{s_2}, \dots, \mathcal{E}_{s_{K_s}}\}$, where $\mathcal{E}_{s_k}=\{\mathbf{e}^1_{s_k}, \mathbf{e}^2_{s_k}, \dots, \mathbf{e}^{K_o}_{s_k}\}$ are labeled with their semantics $\mathcal{L}_{s_k}$. Otherwise, the remaining clusters after line extraction can be extracted as key objects $\mathcal{E}_o = \{\mathbf{e}^1_o, \mathbf{e}^2_o, \dots, \mathbf{e}^{K_o}_o\}$, which are labeled with object semantics $\mathcal{L}_o$.
\end{enumerate}

Through the above process, the point cloud $\mathcal{P}$ is reduced to the element point clouds $\mathcal{E} = \{\mathbf{e}_k | \mathbf{e}_k \in \{\mathcal{E}_g,\mathcal{E}_p,\mathcal{E}_l,\mathcal{E}_o\}\}$ and semantic labels $\mathcal{L}_e = \{l_k | l_k \in\{\mathcal{L}_g,\mathcal{L}_p,\mathcal{L}_l,\mathcal{L}_o\}\}$ are recorded to facilitate quadric fitting, where $\left | \mathrm{unique} \left (\mathcal{L}_e\right )\right | =M_l$ denotes the the number of semantic types. They simplify the scene and preserve key elements to support the registration task.

\subsubsection{Quadric Fitting.}
The elements $\mathbf{e}$ in $\mathcal{E}$ are still represented as point clouds. In this section, each $\mathbf{e}$ in $\mathcal{E}$ will be fitted to a specific type of quadrics based on their semantic labels $\mathcal{L}_e$. We seek that even if the observed point cloud $\mathbf{e}$ of a structure or object is noisy or partially observed, the quadric that closely approximates its underlying complete surfaces can be estimated. The ultimate goal is to maintain consistency in the quadric representation of the same element as much as possible when the viewpoint varies.
\begin{figure}[h]
  \centering
   \includegraphics[width=\linewidth]{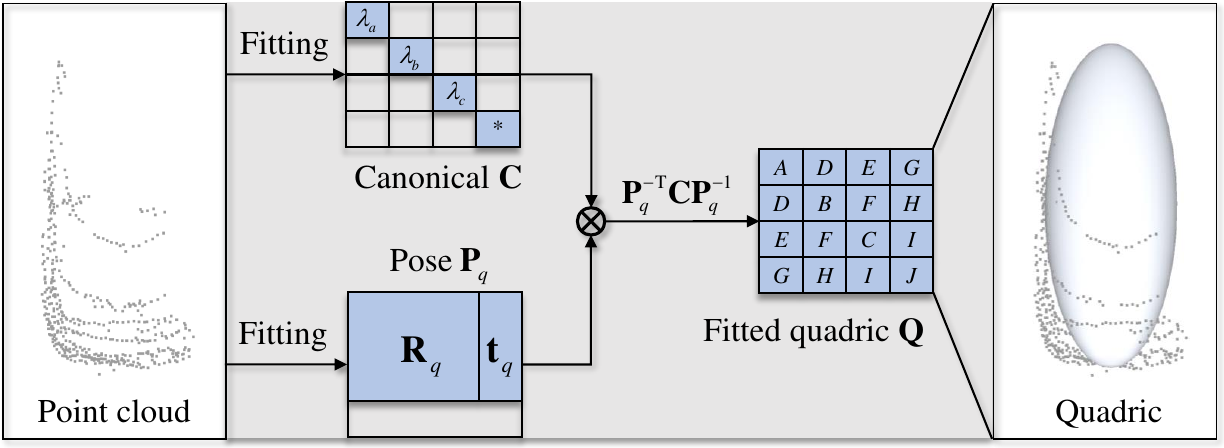}
   \caption{The quadric fitting strategy of point clouds. From a partially observed quadric point cloud, we first fit the canonical $\mathbf{C}$ and the pose $\mathbf{P}_q$, define the type and scale. Then we combine these attributes to form the quadric parameter $\mathbf{Q}$. 
   }   
   \label{fig: illustration of quadrics fitting strategy}
\end{figure}
Based on the quadric mathematical modeling, its representation is highly compact, where even a minor deviation in parameters may lead to significant differences in its geometric attributes. The relationships between quadric mathematical formulation and geometric attributes should be integrated into the fitting process to ensure robust fitting. Therefore, as illustrated in Fig. \ref{fig: illustration of quadrics fitting strategy}, instead of estimating $\mathbf{Q}$ directly, we first estimate the canonical matrix $\mathbf{C}$ and pose matrix $\mathbf{P}_q$ separately, which define the type, scale, and pose of the quadric. We seek to robustly estimate accurate geometric attributes from noisy and fragmented point clouds.
Here, the mapping function $f_l(\mathcal{L}_e)=\mathcal{L}_q$ is defined to assign semantic labels in $\mathcal{L}_e$ to specific quadric types $\mathcal{L}_q = \{l_q | l_q \in \{\mathrm{point,line,plane,ellipsoid},\dots\}\}$ for fitting, thereby determining the form $\mathbf{I_C}$ of the matrix $\mathbf{C}$. Then, $\mathbf{Q}$ is formed based on $\mathbf{C}$ and $\mathbf{P}_q$ according to the decomposition formulation in Eq. \ref{eq: Q=PCP}. 

Following these insights, we leverage QuadricsNet \citep{Wu2024QuadricsNet}, a learning-based method to fit quadrics $\mathbf{Q}$ of specified types $l_q$ from the partially observed point clouds $\mathbf{e}$. In that, $\mathbf{C}_l$ and $\mathbf{P}_l(\mathbf{R}_l, \mathbf{t}_l)$ are learned separately and then combined to form $\mathbf{Q}$. During the training, the inputs are the fragmented point clouds with noises, and the ground truths are the accurate quadric parameters of their complete case. With a large amount of training data, QuadricsNet can achieve robustness to noise and fragmentation. 

However, in some complex fields, certain patterns may inevitably fall outside the scope of the training data, resulting in reduced network performance. Complementarily, we propose a statistics-based fitting method for quadrics:
\begin{enumerate} 
    \item \textbf{Centralization}: $
    \mathbf{e}_c = \mathbf{e} - \mathbf{t}_s$, where $\mathbf{t}_s = \frac{1}{n_e}\sum_{\mathbf{x}\in\mathbf{e}} \mathbf{x}$ is the center of the quadric.
    \item \textbf{Eigenvalue decomposition}: The covariance matrix $
    \frac{1}{n_e} \mathbf{e}_{c}^{\mathrm{T}} \mathbf{e}_c = \mathbf{R}_s \mathbf{\Lambda}_s \mathbf{R}_s^{\mathrm{T}}$, where $\mathbf{\Lambda}_s$ is a diagonal matrix containing eigenvalues and $\mathbf{R}_s$ contains eigenvectors. The standard deviation vector $\mathbf{\sigma}_s = \sqrt{\mathrm{diag}(\mathbf{\Lambda}_s)}$, with values in descending order. 
    \item \textbf{Quadric computation}: The scale of the quadric can be approximated as $\mathbf{s}_s=k_s \mathbf{\sigma}_s$. Assuming the point cloud $\mathbf{e}$ follows a Gaussian distribution and to include 90\% of points, \( k_s \) is set accordingly (\( k_s \approx 1.645 \)). Denoting $\mathbf{S}_{e} = \mathrm{diag}(\left [\mathbf{s}_s,1\right ])$ as the scale matrix, the canonical matrix $\mathbf{C}_s$ and pose matrix $\mathbf{P}_s$ of the quadric can be computed as
    \begin{equation}
     \mathbf{C}_{e}= \mathbf{S}_s^\mathrm{-T} \mathbf{I_C} \mathbf{S}_s^\mathrm{-1} , \mathbf{P}_s= \left[\begin{array}{@{}cc@{}}
    \mathbf{R}_s & \mathbf{t}_s \\
    \mathbf{0}^{\mathrm{T}} & 1
    \end{array}\right],
    \end{equation}
    where $\mathbf{I_C}$ indicates the form of $\mathbf{C}_{e}$ determined by the quadric type $l_{q}$. Then, the quadric parameters can be formed as $\mathbf{Q} = \mathbf{P}_s^{-\mathrm{T}} \mathbf{C}_s \mathbf{P}_s^{-1}$.
\end{enumerate}
The function $f_{p2s}$ is used to measure the distance of the point cloud $\mathbf{e}$ to fitted surface $\mathbf{Q}$ in the unit space \citep{Taubin1991Estimation}:
\begin{equation}
 f_{p2s}(\check{\mathbf{e}},\check{\mathbf{Q}}) = \frac{1}{n_e}\sum_{\mathbf{x}\in\check{\mathbf{e}}} \left |  \frac{\mathbf{x}^\mathrm{T} \check{\mathbf{Q}}\mathbf{x}}{\mathbf{x}^\mathrm{T} \mathbf{\nabla \check{Q}}\mathbf{x}} \right |,
\end{equation}
where superscript $\vee$ indicates normalization to unit space, and $n_e$ denotes the number of points in $\mathbf{e}$. If $f_{p2s}(\check{\mathbf{e}},\check{\mathbf{Q}}_l)$ exceeds the threshold $\delta_p$ indicating that the learning-based result $\mathbf{Q}_l$ does not approximate the point cloud $\mathbf{e}$ well, so the statistics-based result $\mathbf{Q}_s$ is chosen as $\mathbf{Q}$.

Both learning-based and statistics-based methods are essential: The former handles noise and partial observation, while the latter ensures reliable fitting under out-of-distribution cases. Through two complementary processes, $\mathbf{Q}$ is robustly fitted for each element in $\mathcal{E}$. Additionally, the statistics-based method provides non-degenerate statistical scale $\mathbf{s}_s$ and pose $\mathbf{P}_s(\mathbf{R}_s, \mathbf{t}_s)$ of each element to compensate for quadric degeneracy. 

\begin{figure}[th!]
  \centering
   \includegraphics[width=\linewidth]{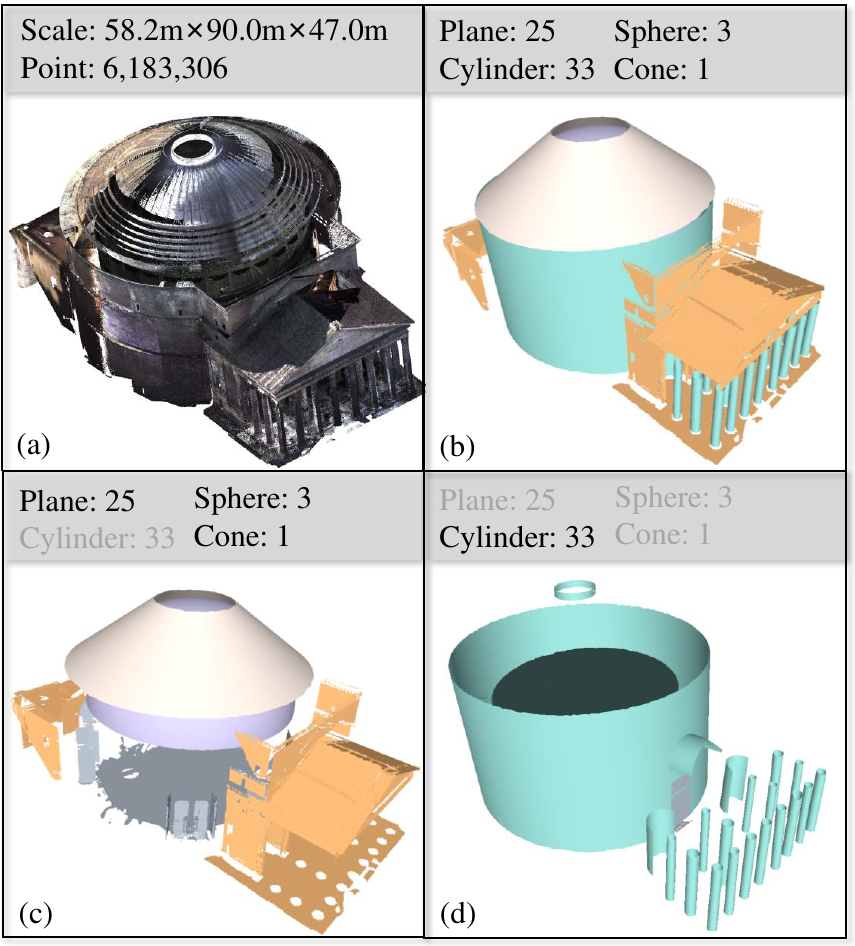}
   \caption{The quadric modeling of Pantheon in Italy. The raw 6 million point clouds are concisely represented by 33 cylinders, 25 planes, 3 spheres, and 1 cone in quadric format. }   
   \label{fig:quadricsPantheon}
\end{figure}
\subsubsection{Quadric Modeling.}
\label{sect: Quadrics Modeling}
In this section, we model the point clouds $\mathcal{P}$ with a concise quadric representation $\mathcal{R}=\{\mathbf{r}_1, \mathbf{r}_2, \dots, \mathbf{r}_N\}$ based on the extraction and fitting results. We aim to create a lightweight representation that efficiently supports global registration while minimally describing the scene. As illustrated in Fig. \ref{fig:quadricsPantheon}, a large 3D scene can be concisely represented by several quadric primitives.

\begin{algorithm}[htbp]
    \SetAlgoLined
    \NoCaptionOfAlgo
    \caption{\textbf{Struct. 1} Quadric representation for an element}
    \label{structure:1}
     $l\in \mathbb{N}_{1\times 1}$; \% Semantic label\\
     $\mathbf{q}\in \mathbb{R}_{1\times 10}$; \% Quadric parameters\\
     $\mathbf{s}_f\in \mathbb{R}^{+}_{1\times 3}$; \% Full scale\\
     $\bm{\eta}_f \in \mathbb{R}_{1\times 4}$; \% Full rotation in quaternion form\\
     $\mathbf{t}_f \in \mathbb{R}_{1\times 3}$; \% Full center\\
\end{algorithm} 

The content of each quadric representation $\mathbf{r}$ in $\mathcal{R}$ is given in Struct. \ref{structure:1}, which contains a total of only 21 parameters. The semantic label $l$ is extracted along with the element $\mathbf{e}$, which is used to reduce the computation space during registration. The parameters $\mathbf{q}$ (the vector form of $\mathbf{Q}$) fitted from $\mathbf{e}$ can be decomposed to obtain scale $\hat{\mathbf{s}}_q$, rotation $\hat{\mathbf{R}}_q$ and center $\hat{\mathbf{t}}_q$ that are robust to viewpoint changes, which are crucial for the subsequent registration process. Meanwhile, these decomposed attributes with degeneracy are indicated by $\mathbf{I_{s,R,t}}$ in certain cases, meaning that the quadric lacks the scale, rotation, or translation attributes along certain axes. For example, in the case of a plane, the scale is completely degenerate, rotation along the normal is degenerate, and the planar center is defined only by the offset at the normal. These degenerate attributes should be eliminated during transformation optimization but still provide clues for the correspondence establishment and transformation initialization between the clouds. Therefore, the statistical scale $\mathbf{s}_s$ and pose $\mathbf{P}_s(\mathbf{R}_s, \mathbf{t}_s)$ are employed as a complement to get the full attributes:
\begin{equation}
\left\{ 
\begin{array}{lll}
\mathbf{s}_f&=&\mathrm{diag}(\mathbf{I_s})\hat{\mathbf{s}}_q+\mathrm{diag}(\mathbf{1} - \mathbf{I_s})\mathbf{s}_s,\\
\mathbf{R}_f&=&\hat{\mathbf{R}}_q \, \mathrm{diag} (\mathbf{I}_\mathbf{R}) + \mathbf{R}_s \, \mathrm{diag} (\mathbf{1} - \mathbf{I_R})
,\\
\mathbf{t}_f &=& \mathrm{diag}(\mathbf{I^{'}_t}) \hat{\mathbf{t}}_q + \mathrm{diag}(\mathbf{1} - \mathbf{I^{'}_t})\mathbf{t}_s,
\end{array} \right.
\end{equation}
where $\mathbf{1} - \mathbf{I}$ denotes the negation of $\mathbf{I}$. If any element of $\mathbf{I_t}$ is zero, then $\mathbf{I^{'}_t}$ defined as the zero vector. The rotation $\mathbf{R}_f$ is simplified to a quaternion form in Struct. \ref{structure:1}. 

To further ensure a concise representation, we select the top-$K_e$ quadrics by size for each semantic type to retain in $\mathcal{R}$. The size $V_a$ of the quadric is measured as the product of the first $n$ elements in the vector $\mathbf{s}_f$, where $n=1,2,3$ depending on whether the element is a line, plane, or volume, correlating to length, area, and volume, respectively, as larger quadrics are typically more distinctive and stable. After the selection, the number of elements in 
$\mathcal{R}$ will not exceed $M_l\times K_e$, where $M_l$ is  the the number of semantic types.

The performance of global point cloud registration significantly depends on the number of inlier correspondences, typically requiring at least 3 matching pairs to avoid correspondence degeneracy. As illustrated in the bottom of Fig. \ref{fig: Illustration of quadrics-based scene representation}, for extended and repetitive scenes with sparse objects like highways or corridors, the extracted key elements are considerably reduced, which in turn decreases the number of correspondences. Moreover, extended elements like continuous walls exhibit strong viewpoint-dependent geometric attributes in such scenes, making it challenging to establish correspondences even for the same element from different viewpoints. As a result, treating each structure or object as one element may lead to correspondence degeneracy and reduced registration performance. To address this, we first check if the number of elements in $\mathcal{R}$ exceeds the threshold $\delta_a$. If not, we augment the top-$K_a$ largest elements of each semantic type $l_e \in \mathrm{unique} \left (\mathcal{L}_e\right )$ in $\mathcal{E}$ by voxel down-sampling with a voxel size $v_a$. The coordinates of down-sampled points are set as the center and other attributes degenerate, and their quadric parameters are fitted as point type statistically. 
Finally, the augmented points are also represented in the form of 21 parameters $\mathbf{r}_a$ and expanded into the representation $\mathcal{R}$.

As illustrated in Fig. \ref{fig: Illustration of quadrics-based scene representation}, based on the extraction and fitting information, the scene is modeled from the dense point cloud $\mathcal{P}$ to the quadric representation $\mathcal{R}\in \mathbb{R}^{N\times 21}$, where $N\le M_l\times K_e+N^a$ and $N^a$ is the number of the augment points. In typical scenes, the number of key elements is sufficient to represent the scene, thus $N^a=0$. In the KITTI dataset \citep{Geiger2012KITTI}, each frame captured by the 64-beam LiDAR contains approximately 120K 3D points, with around $N=100$ key elements. If each key element is represented in a unified manner by 21 parameters $\mathbf{r}$, the quadric representation $\mathcal{R}$ of the scene is comparable to about 700 3D points. Even for scenes with sparse elements, the number of points augmented by downsampling large elements $N^a$ is limited. Therefore, the quadric representation $\mathcal{R}$ of the scene is concise, while still containing the semantics, scale, rotation, center, and even degeneracy of the key elements.

\subsection{Quadric Matching}
\begin{figure}[h!]
  \centering
   \includegraphics[width=\linewidth]{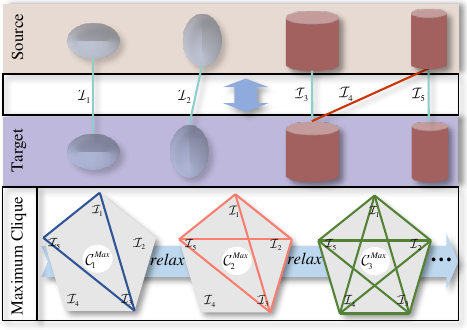}
   \caption{Quadric matching using multi-level compatibility graphs. We first establish quadric correspondences $\mathcal{I}_{1-5}$ within the shared semantics based on quadric similarity. Among them, $\mathcal{I}_{1,2,3,5}$ are inliers, while $\mathcal{I}_{4}$ is an outlier. By progressively relaxing threshold $\delta_m$, we construct multi-level compatibility graphs and determine the maximum cliques. A strict threshold-derived clique $\mathcal{C}^{Max}_{1}$ may reject $\mathcal{I}_{4}$, resulting in sparse correspondences, whereas a loose $\mathcal{C}^{Max}_{3}$ retains more but includes outliers. }   
   \label{fig: Quadrics Matching}
\end{figure}

Given the partially overlapping source point cloud $\mathcal{P}_x$ and target point cloud $\mathcal{P}_y$, model them as the concise quadric representations $\mathcal{R}_x$ and $\mathcal{R}_y$. In this section, we aim to establish correspondences between these quadrics. As illustrated in Fig. \ref{fig: Quadrics Matching}, we first initialize correspondences by comparing the geometric similarity of quadrics for each semantic type. Then, we leverage a consistency check with progressively relaxed thresholds to construct a multi-level compatibility graph set, identifying vertices of the maximum clique in each graph as the inlier correspondence candidates.
\subsubsection{Quadric Correspondence Initialization.}
We aim to establish sufficient putative correspondences to cover potential inliers. The conservative strategy is to establish all-to-all correspondences, but it contains numerous outliers leading to computationally expensive pruning. We leverage the semantic and geometric attributes of quadrics to achieve efficient and robust initialization.

Denote the semantic labels in $\mathcal{R}_x$ and $\mathcal{R}_y$ as $\mathcal{L}_x$ and $\mathcal{L}_y$. The intersection of their semantic types is $\mathcal{L}_{\cap} = \mathrm{unique}(\mathcal{L}_x)\cap \mathrm{unique}(\mathcal{L}_y)$. For each semantic type $l_{\cap} \in \mathcal{L}_{\cap}$, we extract the quadric representations of the shared semantic type from $\mathcal{R}_x$ and $\mathcal{R}_y$, denoted as $\mathcal{R}_x^{l_{\cap}}$ and $\mathcal{R}_y^{l_{\cap}}$. The similarity between the quadrics is evaluated as
\begin{equation}
 f_{qsq}(\mathbf{r}_x , \mathbf{r}_y) = - \left \| \mathrm{diag}(\mathbf{I_s})( \mathbf{s}_x^f-\mathbf{s}_y^f)\right \| ,
\end{equation}
where $\mathbf{s}^f$ is the full scale of the quadric represented in $\mathbf{r}$ and $\mathbf{I_s}$ is the indicator of scale degeneracy for its quadric type. For each element $\mathbf{r}_x \in \mathcal{R}_x^{l_{\cap}}$, select the top-$K_s$ similar correspondences from 
$\left\{ f_{qsq}(\mathbf{r}_x, \mathbf{r}_y) \mid \mathbf{r}_y \in \mathcal{R}_y^{l_{\cap}} \right\}$ forming $\mathcal{I}_x^{l_{\cap}}$. Similarly, for each element $\mathbf{r}_y \in \mathcal{R}_y^{l_{\cap}}$, select the top-$K_s$ similar correspondences from $\left\{ f_{qsq}(\mathbf{r}_x, \mathbf{r}_y) \mid \mathbf{r}_x \in \mathcal{R}_x^{l_{\cap}} \right\}$ forming $\mathcal{I}_y^{l_{\cap}}$. Then, the correspondences for semantic $l_{\cap}$ are given by $\mathcal{I}^{l_{\cap}} = \mathcal{I}_x^{l_{\cap}} \cap \mathcal{I}_y^{l_{\cap}}$. The final putative correspondences are obtained by concatenating all correspondences for each semantic type in $\mathcal{L}_{\cap}$, expressed as $\mathcal{I} = \bigcup_{l_{\cap} \in \mathcal{L}_{\cap}} \mathcal{I}^{l_{\cap}} $. For augmented points, we establish correspondences based on the similarity between point-level FPFH descriptors \citep{Rusu2009FPFH}, followed by eliminating pairs whose elements do not share the same semantic label. 

The number of correspondences in $\mathcal{I}$ does not exceed $K_e \times K_s \times M_{\cap} + N_{\cap}^a$, where $M_{\cap}=|\mathcal{L}_{\cap}|$ and $N_{\cap}^a \le \mathrm{min} (N_{x}^a,N_{y}^a)$ represents the number of elements in the intersection of augmented points. For an all-to-all correspondence strategy, the upper limit on the number of correspondences is $N_x\times N_y = (K_e\times M_x + N_x^a)\times (K_e\times M_y + N_y^a)$, where $M_{x}=|\mathcal{L}_{x}|$ and $M_{y}=|\mathcal{L}_{y}|$. This comparison highlights that matching quadrics within each semantic type by geometric similarity significantly reduces the number of putative correspondences while ensuring the inclusion of correct pairs.

\subsubsection{Graph-based Outlier Pruning.}
In the putative correspondences $\mathcal{I}$ with outliers, the pruning process aims to find the largest inlier correspondence set $\mathcal{I}^*$. We leverage the quadric representations of pairs $(r_x,r_y)$ associated in $\mathcal{I}$ to construct a compatibility graph $\mathcal{G}$ and identify the maximum clique to achieve $\mathcal{I}^*$.

For any combinations of correspondence pairs $(r_{x, i},r_{y, i})$ and $(r_{x, j},r_{y, j})$, inspired by TEASER \citep{Yang2021TEASER}, the translation and rotation invariant measurements are defined based on the full centers $\mathbf{t}^f$ of quadrics as
\begin{equation}
d_{i,j} = \left|\left\| \mathbf{t}_{x,i}^f-\mathbf{t}_{x,j}^f \right\| -\left\| \mathbf{t}_{y,i}^f-\mathbf{t}_{y,j}^f  \right\| \right|.
\end{equation}
If the two correspondence pairs $(r_{x, i},r_{y, i})$ and $(r_{x,j},r_{y,j})$ are inliers and the fitted centers are noise-free, the two will be perfectly consistent, resulting in $d_{i,j}=0$. However, due to inevitable observational errors in practical scenarios, the fitted centers may not be perfectly accurate. As such, we consider correspondences to be mutually consistent if $d_{i,j} \le \delta_m$. For each consistent combination, we represent the two correspondences as vertices and establish an edge between them. Iterating over all combinations in $\mathcal{I}$, we construct the compatibility graph $\mathcal{G}$, where the vertices and edges represent consistent correspondences within $\mathcal{I}$.

We seek to select an optimal threshold $\delta_m$ that maximizes inlier inclusion through mutual consistency while eliminating outliers. The optimal value of $\delta_m$ depends on the observational error bounds in $\mathbf{t}^f$, which vary under different observations. Inspired by a multi-level thresholding strategy \citep{Qiao2023Pyramid,Qiao2024G3reg}, as illustrated in Fig. \ref{fig: Quadrics Matching}, we begin with a strict initial threshold $\delta_m^1$, and progressively relax it to generate a set of thresholds $[\delta_m^1,\delta_m^2,\dots,\delta_m^{K_g}]$. This strategy enables us to construct a multi-level compatibility graph set $\{\mathcal{G}_1,\mathcal{G}_2,\dots,\mathcal{G}_{K_g}\}$ for the robust consistency check. The maximum clique $\mathcal{C}^{Max}_{k_g}$ represents the largest set of mutually consistent inliers. According to \citep{Rossi2015PMC,Qiao2023Pyramid,Qiao2024G3reg}, where $|\mathcal{C}^{Max}_{1}| \le |\mathcal{C}^{Max}_{2}| \le \dots \le |\mathcal{C}^{Max}_{K_g}|$, we employ the graduated PMC to find the maximum cliques. The number of vertices $|\mathcal{C}^{Max}_{k_g}|$ in $\mathcal{G}_{k_g}$ serves as a lower bound for $|\mathcal{C}^{Max}_{k_g+1}|$ in $\mathcal{G}_{k_g+1}$, accelerating the search in PMC. Finally, the vertex sets in all maximum cliques are identified as the inlier correspondence candidates $\{\mathcal{I}_1^*,\mathcal{I}_2^*,\dots,\mathcal{I}_{K_g}^*\}$. 
These candidates are used to estimate the transformation between the paired point clouds in the next section.

\subsection{Quadric-based 6-DoF Transformation Estimation}
\label{sect. Transformation Estimation}
This section presents the process of deriving transformation candidates and validating the optimal transformation. We propose a factor graph optimization method based on the quadric distance to estimate transformation candidates. Each candidate is then applied to $\mathcal{R}_x$, and the candidate that results in the minimum distance between the transformed $\mathcal{R}_x$ and $\mathcal{R}_y$ is selected as the optimal transformation.

\begin{algorithm}[htbp]
    \SetKwFunction{FInitialize}{Initialize}
    \SetKwFunction{FOptimizer}{LMOptimizer}
    \SetKwFunction{FSVD}{SVD}
    \SetKwFunction{FSNN}{SNN}
    \SetKwFunction{FAugment}{Augment}
    \SetKwFunction{FAddFactor}{AddFactor}

    \NoCaptionOfAlgo
    \caption{\textbf{Algorithm 1: Quadric-based Transformation Estimation}}
    \label{algorithm: Robust Transformation Estimation}
    \textbf{Input:} Quadric representations of source and target point clouds $\mathcal{R}_x$ and $\mathcal{R}_y$, inlier correspondence candidates set $\{\mathcal{I}_1^*,\mathcal{I}_2^*,\dots,\mathcal{I}_{K_g}^*\}$ \\
    \textbf{Output:} Optimal transformation $\mathbf{T}^*(\mathbf{R}^*,\mathbf{t}^*)$ \\
    
    \% Transformation Candidate Estimation \\
    Initialize transformation candidates set: $\hat{\mathcal{T}} = \emptyset$ \\
    \For{$\mathcal{I}^*_{k_g}$ \textnormal{in} $\{\mathcal{I}_1^*,\mathcal{I}_2^*,\dots,\mathcal{I}_{K_g}^*\}$}{
        $\mathbf{T}^0_{k_g} = \FSVD(\mathcal{R}_x,\mathcal{R}_y,\mathcal{I}^*_{k_g})$ \\
        Initialize factor graph: $\mathcal{F}_{k_g}\mathrm{.initial}(\mathbf{T}^0_{k_g})$ \\
        
        \For{$(\mathbf{r}_x,\mathbf{r}_y)$ \textnormal{in} $\mathcal{I}^*$}{
            \If{$f_{pp}(\mathbf{r}_{x,s})$ \textnormal{or} $f_{pp}(\mathbf{r}_{y,s})$}{
                \textbf{continue} \\
            }
            \FAugment{$\mathbf{r}_x,\mathbf{r}_y$} \\
            Quadric distance: $\mathbf{e} = [\mathbf{e_R},\mathbf{e_t}]^{\mathrm{T}}$ \\
            Jacobian: $\mathbf{J} = \left[\frac{\partial \mathbf{e_R}}{\partial \mathbf{R}}, \frac{\partial \mathbf{e_R}}{\partial \mathbf{t}} ; \frac{\partial \mathbf{e_t}}{\partial \mathbf{R}}, \frac{\partial \mathbf{e_t}}{\partial \mathbf{t}} \right] $ \\
            $\mathcal{F}_{k_g}\mathrm{.addFactor}(\mathbf{e}, \mathbf{J})$ \\
        }
        
        $\mathbf{T}_{k_g} = \FOptimizer(\mathcal{F}_{k_g})\mathrm{.pose}$ \\
        $\mathcal{T} = \mathcal{T} \cup \{\mathbf{T}_{k_g}\}$ \\
    }
    
    \% Optimal Transformation Selection \\
    Initialize distance set: $\mathcal{D}_v = \emptyset$ \\
    \For{$\mathbf{T}_{k_g}$ \textnormal{in} $\mathcal{T}$}{
        $\mathcal{I}_{snn} = \FSNN(f_t(\mathcal{R}_x|\mathbf{T}_{k_g}),\mathcal{R}_y)$ \\
        $d_v = []$
        
        \For{$(\mathbf{r}_x,\mathbf{r}_y)$ \textnormal{in} $\mathcal{I}_{snn}$}{
            \If{$f_{pp}(\mathbf{r}_{x,s})$ \textnormal{or} $f_{pp}(\mathbf{r}_{y,s})$}{
                \textbf{continue} \\
            }
            \FAugment{$\mathbf{r}_x,\mathbf{r}_y$} \\
            $d_v.\mathrm{append}(\rho(\left \| \mathbf{e_R} \right \| + \left \| \mathbf{e_t} \right \|))$ \\
        }
        $\mathcal{D}_v = \mathcal{D}_v \cup \left\{\sum d_v / {|d_v|}\right\}$ \\
    }

    Optimal transformation: $\mathbf{T}^* = \underset{\mathbf{T}_{k_g} \in \mathcal{T}}{\mathrm{arg\,min}} \, \mathcal{D}_v$ 
\end{algorithm}

\subsubsection{Transformation Candidate Estimation.}
For each correspondence candidate $\mathcal{I}_{k_g}^*$ in $\{\mathcal{I}_1^*,\mathcal{I}_2^*,\dots,\mathcal{I}_{K_g}^*\}$, we aim to estimate the transformation $\hat{\mathbf{T}}_{k_g}$ between the paired point clouds only based on $\mathcal{R}_x$ and $\mathcal{R}_y$. We first estimate an initial transformation directly with SVD based on $\mathcal{I}_{k_g}^*$ and then refine the transformation by optimizing it with a factor graph.

\begin{figure}[h!]
  \centering
   \includegraphics[width=\linewidth]{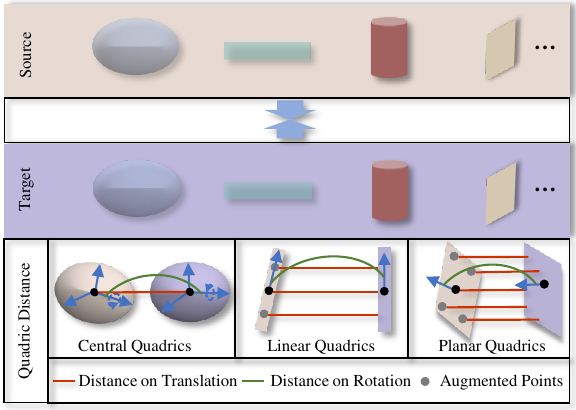}
   \caption{The distance between quadrics is composed of translational and rotational components. Leveraging the degeneracy of quadrics, we compute quadric distance among diverse geometric primitives in a unified manner. The translational distances for central, linear-center, and planar-center quadrics are measured by point-to-point, point-to-line, and point-to-plane distances, respectively. Additionally, sampling on the source surface is used for non-central quadrics to compensate for degeneracy. Rotational distances are measured along non-symmetric axes. 
   }   
   \label{fig: Illustration of quadrics distance}
\end{figure}
According to the correspondence pairs $\{(\mathbf{r}_x, \mathbf{r}_y) \in \mathcal{R}_x \times \mathcal{R}_y \mid \text{associated in } \mathcal{I}_{k_g}^*\}$, we estimate the initial transformation $\mathbf{T}^0_{k_g}$ by SVD, based on the full center $\mathbf{t}_f$ of each $\mathbf{r}$. A nonlinear factor graph is constructed to further optimize the transformation $\mathbf{T}_{k_g}(\mathbf{R}_{k_g},\mathbf{t}_{k_g})$, starting from $\mathbf{T}^0_{k_g}$. Before adding the factors, we filter out irregular key structures in the scene $\mathcal{R}_x$ and $\mathcal{R}_y$. Specifically, we remove structures for which the direction is neither perpendicular nor parallel to the ground:
\begin{equation}
f_{pp}(\mathbf{r}_s) = 
\text{\scriptsize 
$(\theta > \delta_g^e) \lor (|\theta - 180^\circ| > \delta_g^e) \lor (|\theta - 90^\circ| > \delta_g^e)
$},
\end{equation}
where $\theta = \arccos \left(\hat{\mathbf{R}}_{q}[:,i] \cdot \hat{\mathbf{v}}_g \right)$ for $\mathbf{I}_{\mathbf{R}}[i] = 1$, and $\delta_g^r$ is the threshold of this removal. Here, $\hat{\mathbf{R}}_q$ represents the rotation, and $\mathbf{I_{R}}$ is the indicator of the rotation degeneracy, they all inferred from the quadric parameters $\mathbf{q}$ in $\mathbf{r}$. The subscripts $s$ and $g$ denote the key structure and ground, as determined through semantic label $l$ in $\mathbf{r}$. $\hat{\mathbf{v}}_g$ is the normal vector of the ground. 

The distance $\mathbf{e} = [\mathbf{e_R},\mathbf{e_t}]^{\mathrm{T}} \in \mathbb{R}^{12}$ between quadrics in each pair $(\mathbf{r}_x, \mathbf{r}_y)$ is used as the error function in Eq. \ref{eq: formulation}:
\begin{equation}
\label{eq: quadrics distance}
\begin{bmatrix}
 \mathbf{e_R}\\
\mathbf{e_t}
\end{bmatrix} = 
\begin{bmatrix}
\mathrm{diag}(\mathbf{I}_{\mathbf{R},y})(\mathbf{R}_{k_g}\hat{\mathbf{R}}_{q,x} \otimes \hat{\mathbf{R}}_{q,y}) \\
\mathrm{diag}(\mathbf{I}_{\mathbf{t},y})\hat{\mathbf{R}}_{q,y}^{\mathrm{T}}( \mathbf{R}_{k_g} \mathbf{t}_{f,x}+\mathbf{t}_{k_g} - \mathbf{t}_{f,y})
\end{bmatrix},
\end{equation}
where $\mathbf{I_{t}}$ indicates the translation degeneracy, $\otimes$ denotes the column-wise cross product, and $\mathbf{t}_{f}$ is the fully center in $\mathbf{r}$. The error is measured from both rotation $\mathbf{e_R}\in \mathbb{R}^{9}$ and translation $\mathbf{e_t}\in \mathbb{R}^{3}$ aspects, leveraging the geometric properties of quadrics fully. As illustrated in Fig. \ref{fig: Illustration of quadrics distance}, this strategy eliminates the influence of degenerate geometric attributes. Specifically, we achieve principal axis alignment for rotation, as well as point-to-point, point-to-line, and point-to-plane measurements for translation. The Jacobian matrix for the translation component is given by $\frac{\partial \mathbf{e_R}}{\partial \mathbf{t}} = \mathbf{0} \in \mathbb{R}^{9\times 3}$ and $\frac{\partial \mathbf{e_t}}{\partial \mathbf{t}} = \mathrm{diag}(\mathbf{I}_{\mathbf{t},y}) \hat{\mathbf{R}}_{q,y}^{\mathrm{T}} \in \mathbb{R}^{3\times 3}$. The remaining Jacobian terms $\frac{\partial \mathbf{e_R}}{\partial \mathbf{R}} \in \mathbb{R}^{9\times 3}$ and $\frac{\partial \mathbf{e_t}}{\partial \mathbf{R}} \in \mathbb{R}^{3\times 3}$ are derived from the right perturbation model:
\begin{equation}
\label{eq: Jacobian of quadrics distance}
\begin{aligned}
 &\frac{\partial \mathbf{e_R}}{\partial \mathbf{R}} = \mathrm{diag}(\mathbf{I}_{\mathbf{R},y}^{'})
\frac{
\partial \left[\begin{array}{c}
\cdots \\
\mathbf{R}_{k_g}\hat{\mathbf{R}}_{q,x}[:,i] \times \hat{\mathbf{R}}_{q,y}[:,i]\\
\cdots
\end{array}\right]
}
{\partial \mathbf{R}} \\
&= \mathrm{diag}(\mathbf{I}_{\mathbf{R},y}^{'})
\left[\begin{array}{c}
\cdots \\
-\mathbf{R}_{k_g} [\hat{\mathbf{R}}_{q,x}[:,i] \times \hat{\mathbf{R}}_{q,y}[:,i]]_{\times}\\
\cdots
\end{array}\right], \\
& \frac{\partial \mathbf{e_t}}{\partial \mathbf{R}} = - \mathrm{diag}(\mathbf{I}_{\mathbf{t},y}) \hat{\mathbf{R}}_{q,y}^{\mathrm{T}} \mathbf{R}_{k_g} [\mathbf{t}_{f,x} ]_{\times},
\end{aligned}
\end{equation}
where $\mathbf{I}_{\mathbf{R},y}^{'} \in \{0,1\}^{9}$ is formed by repeating each element of $\mathbf{I}_{\mathbf{R},y}$ three times, $\times$ represents the cross product, and $[\cdot]_{\times}$ denotes the skew-symmetric matrix representation. Finally, the Jacobi matrix is combined as $\mathbf{J} = \left[\frac{\partial \mathbf{e_R}}{\partial \mathbf{R}}, \frac{\partial \mathbf{e_R}}{\partial \mathbf{t}} ; \frac{\partial \mathbf{e_t}}{\partial \mathbf{R}}, \frac{\partial \mathbf{e_t}}{\partial \mathbf{t}} \right] \in \mathbb{R}^{12\times 6}$.

For non-central quadrics such as planes, lines, and cylinders, which are ubiquitous in urban environments, even though the translation error formulation accounts for degeneracy, two challenges may arise. First, eliminating the error along the degenerate translation axis may diminish the contribution of these quadrics during the optimization. Second, if the center points of non-central quadrics lie near their intersection, even when the two quadrics are not perfectly aligned, the translation error can still be minimal. To address this, as illustrated in Fig. \ref{fig: Illustration of quadrics distance}, for $\mathbf{r}_x$ of such quadrics, we augment the number of points along the central line or central plane (2 samples for linear quadrics, 4 for planar, with sampling radius equal to the mean scale). These augmented points in the source are associated with the original target quadrics $\mathbf{r}_y$  as augmented correspondences.

The errors $e(\mathbf{r}_x, \mathbf{r}_y)$ for each correspondence or augmented correspondence in $\mathcal{I}_{k_g}^*$, along with their Jacobian matrices, are added to the factor graph. We then perform optimization using the Levenberg-Marquardt algorithm to refine the transformation. This process is repeated for all correspondence candidates in the set $\{\mathcal{I}_1^*,\mathcal{I}_2^*,\dots,\mathcal{I}_{K_g}^* \}$, resulting in the estimation of transformation candidates $\mathcal{T} = \{ \mathbf{T}_1, \mathbf{T}_2, \dots, \mathbf{T}_{K_g}\}$.

\subsubsection{Optimal Transformation Selection.}
Based on the representations $\mathcal{R}_x$ and $\mathcal{R}_y$, we aim to find the optimal transformation $\mathbf{T}^*$ within the set $\mathcal{T}$. Eq. \ref{eq: formulation} is rewritten as
\begin{equation}
    \mathbf{T}^* = 
    \argmin_{\mathbf{T}_{k_g} \in \mathcal{T}} \frac{1}{\left | \mathcal{I}_{snn} \right | }  \sum_{\left(\mathbf{r}_x, \mathbf{r}_y\right) \in \mathcal{I}_{snn}} \rho\left( \left \| \mathbf{e_R} \right \| + \left \| \mathbf{e_t} \right \| \right),
\end{equation}
where $\mathbf{e_R}$ and $\mathbf{e_t}$ represent the distances between quadrics as defined in Eq. \ref{eq: quadrics distance}, and $\mathcal{I}_{snn}$ denotes the one-to-one correspondence established by performing a nearest neighbor search for each semantic type (SNN) in $\mathcal{L}_{\cap}$ on full centers from $f_t(\mathcal{R}_x|\mathbf{T}_{k_g})$ to $\mathcal{R}_y$. Specifically, $f_t(\mathcal{R}_x|\mathbf{T}_{k_g}) = \{\mathbf{R}_{k_g}\mathbf{t}_x+\mathbf{t}_{k_g} | \mathbf{t}_f\in \mathbf{r}_x, \mathbf{r}_x \in \mathbf{R}_x \}$ denotes the transformation of $\mathcal{R}_x$ using $\mathbf{T}_{k_g}$. We also filter out irregular structures with the threshold $\delta_g^e$ and perform correspondence augmentation for non-central quadrics.

\section{Experiments and Analysis}
\label{sec5:experiment}
\subsection{Dataset}
To evaluate the performance of global point cloud registration methods, we conduct experiments using partially overlapping point clouds from both loop closure (LC) and odometry (ODO) scenarios. The loop closure scenario provides a broader distribution of transformations, whereas the odometry scenario is commonly used as a benchmark for evaluating global point cloud registration \citep{Lim2024quatro++,Qiao2024G3reg,Yin2023Segregator,Bai2021Pointdsc,Choy2020DGR,Choy2019FCGF}. 

\begin{table}[h]
    \centering
    \captionsetup{font=footnotesize}
    \caption{\color{black}Datasets of experiments.}
    \label{table: dataset}
    {
    \scriptsize
        \centering
            \begin{tabular}{c|l|c|c|c}
                \toprule
                 & Dataset & LiDAR-Beam & Points/Frame & Evaluation \\ 
                \cline{1-5}
                \parbox[t]{2mm}{\multirow{5}{*}{\centering\rotatebox[origin=c]{90}{Public}}} & KITTI & Velodyne-64 & 120K & LC, ODO \\
                & KITTI-360 & Velodyne-64 & 120K & LC, ODO \\
                & Apollo-SouthBay & Velodyne-64 & 120K & LC, ODO \\
                & Waymo & Spinning-64$^{*}$ & 177K & ODO \\
                & nuScenes & Spinning-32$^{*}$ & 34K & ODO \\
                \cline{1-5}
                \parbox[t]{2mm}{\multirow{3}{*}{\centering\rotatebox[origin=c]{90}{Self.}}} & \multirow{3}{*}{Hetero-Reg} & Velodyne-16 & 30K & \multirow{3}{*}{LC, MM} \\ 
                & & Livox \emph{Mid-360} & 20K & \\
                & & Livox Avia & 24K & \\
               \bottomrule
            \end{tabular}
    }
    \begin{flushleft}
        \footnotesize{
        $*$: Spinning-state LiDAR (exact model undisclosed).
        }
    \end{flushleft}
\end{table}
\begin{figure}[th!]
  \centering
   \includegraphics[width=\linewidth]{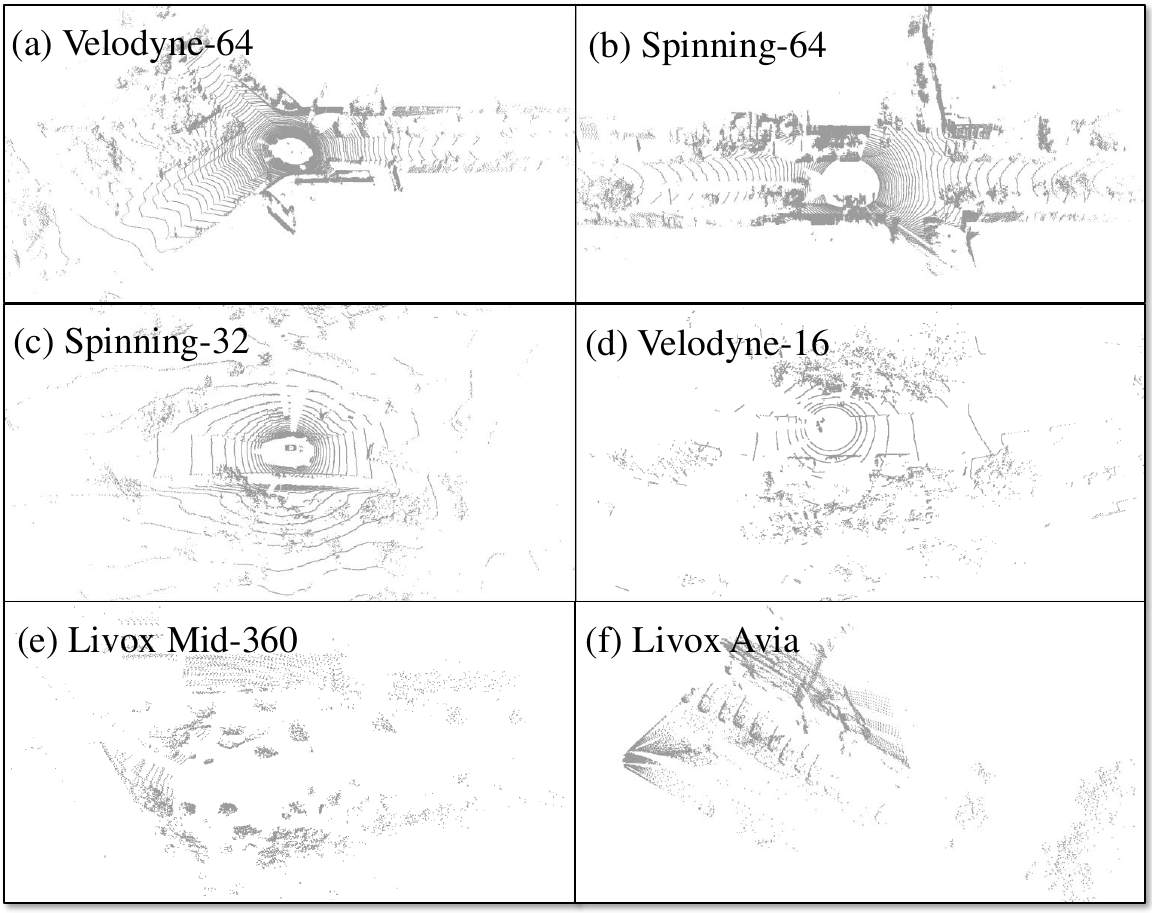}
   \caption{Point clouds collected by different LiDAR sensors in the datasets. We select LiDAR sensors of spinning-state, semi-solid-state, and solid-state types, with the collected point clouds exhibiting differences in spatial arrangement, density, and FOV.}   
   \label{fig: Illustration of sensor data}
\end{figure}

As summarized in Table \ref{table: dataset}, experiments are performed on five public datasets and one self-collected dataset. Fig. \ref{fig: Illustration of sensor data} illustrates the point clouds collected by different LiDAR sensors. KITTI \citep{Geiger2012KITTI}, KITTI-360 \citep{Yiyi2022KITTI-360}, and Apollo-SouthBay \citep{Lu2019L3-netApollo} datasets are large-scale SLAM datasets that provide abundant loops for evaluation. Although Waymo \citep{Sun2020ScalabilityWaymo} and nuScenes \citep{caesar2020nuscenes} datasets contain sequences of only about 20s duration, they encompass thousands of sequences, allowing a comprehensive evaluation of odometry performance across diverse environments. The nuScenes dataset, which uses a 32-beam LiDAR sensor, is suitable for evaluating the generalization ability of global registration methods across different LiDAR sensors. Furthermore, to validate the real-world applicability of QuadricsReg, we build 3 heterogeneous robot platforms mounted with 3 types of LiDAR to collect the Hetero-Reg dataset. QuadricsReg is applied to this real-world dataset for SLAM loop closure and multi-session mapping (MM) scenarios. 

\subsubsection{Dataset from Loop Closure.}
For the loop closure scenario, the set of loop pairs is defined as follows:
\begin{equation}
\label{eq: loop pair gen}
\text{\footnotesize $
\left\{(i, j) \mid d^{\min }_{lc} \leq\left\|\bar{\mathbf{t}}_{i}-\bar{\mathbf{t}}_{j}\right\| \leq d^{\max }_{lc},|i-j| \geq t_{lc}, \forall i, j \in \mathbb{N}_p\right\},
$}
\end{equation}
where $\mathbb{N}_p$ represents the set of frame indices for point clouds within the same sequence, and $\bar{\mathbf{t}}$ denotes the ground truth (GT) position of the point cloud. $d^{\min }_{lc}$ and $d^{\max }_{lc}$ denote the minimum and maximum allowable distances between two frames, respectively, while $t_{lc}$ is a threshold ensuring sufficient temporal separation between selected frames to avoid consecutive selections. Generally, point cloud pairs with greater distances exhibit lower overlap rates, resulting in higher registration difficulty. By adjusting $d^{\min }_{lc}$ and $d^{\max }_{lc}$, we control the difficulty of the loop closure set. In this study, we define three difficulty levels of the loop closure: Easy (0-10 m), medium (10-20 m), and hard (20-30 m). 

For the KITTI dataset, loops are primarily found in sequences 00, 02, 05, 06, and 08. Following the outlined strategy, we processed these five sequences to generate the KITTI-LC dataset. The GT poses for the KITTI dataset are obtained from SemanticKITTI \citep{Behley2019SemanticKITTI}. Compared to KITTI, the KITTI-360 dataset contains more frequent loops, so we selected the sequences with the highest number of loops: 00, 02, 04, 06, and 09, to construct the KITTI-360-LC dataset. Apollo-SouthBay also features numerous loops, and we created the Apollo-SouthBay-LC dataset using five sequences: SunnyvaleBigloop, MathildaAVE, SanJoseDowntown, BaylandsToSeafood, and ColumbiaPark. To ensure consistency, we apply uniform spatial sampling to maintain approximately 1,000 pairs at each difficulty level across all public datasets.

\subsubsection{Dataset from Odometry.}
For the odometry scenario, the set of registration pairs is defined as follows:
\begin{equation}
\text{\footnotesize $
\left\{ (i, j) \mid j = \arg\min_{k} \left( \left |  \left\| \bar{\mathbf{t}}_i - \bar{\mathbf{t}}_k \right\| - d_{odo} \right |  \right), \forall i, j \in \mathbb{N}_p \right\}
$},
\end{equation}
where $d_{odo}$ represents the distance between two point clouds. We varied the distance from 1 m to 10 m, generating 10 difficulty levels of point cloud registration datasets to analyze the effect of viewpoint shift on the performance of registration methods.

For testing in odometry, we aim to cover as many diverse scenes as possible to validate the generalization ability of our method. Following the above-mentioned strategy, we spatially sample approximately 500 pairs from each dataset for each testing level. We use sequences 08-10 for testing on the KITTI dataset, following the deep learning method \citep{Choy2019FCGF,Choy2020DGR,Bai2021Pointdsc}. The KITTI-360 and KITTI are datasets collected using identical LiDAR in similar scenes, and therefore we conduct experiments only on the KITTI dataset. The Apollo-SouthBay dataset involves sequences HighWay237, SunnyvaleBigloop, MathildaAVE, SanJoseDowntown, BaylandsToSeafood, and ColumbiaPark for testing. For the Waymo and nuScenes datasets, we sample testing pairs from the first 100 sequences. As a result, we generate datasets for the odometry scenario named KITTI-ODO, KITTI-360-ODO, Apollo-SouthBay-ODO, Waymo-ODO, and nuScenes-ODO. 
\subsection{Experimental Settings and Evaluation Metrics}
\subsubsection{Implementation Details.}
\label{sect: Implementation Details}
In the experiments, during the quadrics modeling of the scene, $K_p = K_l = K_o = 60$ key elements are extracted, with ground extraction and clustering performed using the default parameters of TRAVEL \citep{Oh2022TRAVEL} and G3Reg \citep{Qiao2024G3reg}. For datasets where semantics can be reliably predicted, PVKD \citep{Hou2022PVKD} is employed to predict semantic labels. The threshold $\delta_p$ for adopting the fitting quadrics of QuadricsNet is set to $0.5$, where the network configuration we follow \citep{Wu2024QuadricsNet}. In the final quadrics representation $\mathcal{R}$, $K_e = 50$ quadrics are retained for each semantic type. If the total number of quadrics in $\mathcal{R}$ is less than $\delta_a = 60$, the top $K_a = 10$ largest quadrics from each semantic type are down-sampled using a voxel size of $v_a = 0.5$ to serve as augmented points. In the correspondence establishment, up to $K_s = 20$ putative correspondences are established for each quadric, and a four-level compatibility graph is constructed using thresholds $[0.2, 0.4, 0.6, 0.8]$. During transformation estimation, irregular structures are removed based on a threshold of $\delta_g^r = 5^{\circ}$. We adopt the dynamic covariance scaling \citep{Agarwal2013DCS} as the robust kernel function $\rho$ in optimal transformation selection.

\begin{table*}[htbp]
    \captionsetup{font=footnotesize}
    \centering
    \caption{Evaluation of Correspondence Establishment on KITTI-LC dataset (Num.: Number of correspondences,  IR: Inlier ratio, unit: \%, CR: Correspondence recall, unit: \%, SR: Registration success rate, unit: \%).}
    \label{table: Evaluation of Correspondence Establishment}
    \setlength{\tabcolsep}{6pt}
    {\scriptsize
        \centering
        \begin{tabular}{l|l|l|cccc|cccc|cccc}
            \toprule
            \parbox[t]{2mm}{\multirow{9}{*}{\rotatebox[origin=c]{90}{KITTI-LC}}} && Level & \multicolumn{4}{c|}{Easy} & \multicolumn{4}{c|}{Medium} & \multicolumn{4}{c}{Hard} \\ \cline{3-15}
    
    
          & & Metrics & Num. & IR $\uparrow$ & CR $\uparrow$  & SR $\uparrow$  & Num. & IR $\uparrow$ & CR $\uparrow$ & SR $\uparrow$  & Num. & IR $\uparrow$ & CR $\uparrow$  & SR $\uparrow$  \\ \cline{2-15}
            
           & \parbox[t]{2mm}{\multirow{2}{*}{\rotatebox[origin=c]{90}{Low.}}} 
           & FPFH & 707.20 & 9.15 & \textbf{100.00} & \textbf{99.38} & 633.12 & 2.55 & 95.60 & 82.37 & 595.63 & 0.89 & 66.20 & 45.17  \\
           && FCGF & 2483.55 & \textbf{17.70} & 80.89 & 73.74 & 1824.31  & \textbf{6.74} & 60.06 & 54.12 & 1568.15 & \underline{2.30} & 43.55 & 39.09\\ \cline{2-15}
           
           &\parbox[t]{2mm}{\multirow{5}{*}{\rotatebox[origin=c]{90}{High.}}}
           & Segregator & 351.98 & \underline{14.52} & 98.64 & 91.61 & 235.56 & \underline{6.18} & 86.13 & 80.71 & 187.19 & \textbf{2.92} & 55.84 & 54.11  \\
           && G3Reg & 1353.92 & 1.52 & \underline{99.20} & 97.53 & 1326.64 & 0.76 & 94.80 & 90.60 & 1313.10 & 0.40 & 76.40 & 69.20  \\
            && $\text{QuadricsReg-15}^{*}$ & 1107.18 & 2.50 & \textbf{100.00} & \underline{98.77} & 1082.75 & 1.26 & 97.40 & 92.91 & 1053.49 & 0.71 & 83.80 & 78.20  \\
          && $\text{QuadricsReg-20}^{*}$ & 1456.42 & 2.05 & \textbf{100.00} & \underline{98.77} & 1427.26 & 1.05 & \underline{98.00} & \underline{93.22} & 1395.62 & 0.60 & \underline{85.60} & \underline{80.00}  \\ 
          && $\text{QuadricsReg-25}^{*}$ & 1787.60 & 1.75 & 99.88 & 98.64 & 1769.24 & 0.87 & \textbf{98.33} & \textbf{93.64} & 1741.44 & 0.50 & \textbf{88.43} & \textbf{81.42} \\ \bottomrule
        \end{tabular}
    }
    
    \begin{flushleft}
        \footnotesize{
        $*$: QuadricsReg-$K_s$, where $K_s$ denotes the upper limit for the initial correspondences of each quadric.
        }
    \end{flushleft}
\end{table*}

\begin{table*}[t!]
    \captionsetup{font=footnotesize}
    \centering
    \caption{ Global registration test for loop closure on KITTI-LC, KITTI-360-LC and Apollo-SouthBay-LC datasets (registration success rate, unit: \%).}
    \label{table: Evaluation of global registration for loop closure test}
    \setlength{\tabcolsep}{4pt}
    {\scriptsize
        \centering
        \begin{tabular}{l|l|l|ccc|ccc|ccc|ccc|ccc}
            \toprule
            \parbox[t]{2mm}{\multirow{14}{*}{\rotatebox[origin=c]{90}{KITTI-LC}}} && Sequence & \multicolumn{3}{c|}{00} & \multicolumn{3}{c|}{02} & \multicolumn{3}{c|}{05} & \multicolumn{3}{c|}{06} & \multicolumn{3}{c}{08}  \\ \cline{3-18}
            
           & & Level & Easy & Medium & Hard & Easy & Medium & Hard & Easy & Medium & Hard & Easy & Medium & Hard & Easy & Medium & Hard \\ \cline{2-18}

           &\parbox[t]{2mm}{\multirow{4}{*}{\rotatebox[origin=c]{90}{Learning}}}
            & FCGF & 89.50 & 67.50 & 48.00 & 62.50 & 49.00 & 32.50 & 82.95 & 69.85 & 51.00 & 93.50 & 49.50 & 37.00  & 2.24 & 0.61 & 0.00  \\
           && DGR & 83.00 & 61.50 & 40.50 & 62.50 & 50.50 & 33.00 & 78.75 & 66.83 & 47.00 & \underline{96.90} & 49.50 & 25.50 & 2.99 & 1.23 & 0.00  \\
           && PointDSC & 79.50 & 61.00 & 45.00 & 53.48 & 46.46 & 37.50 & 77.27 & 62.81 & 49.00 & 89.92 & 49.00 & 36.14 & 2.24 & 0.00 & 0.00  \\
           && LCDNet & 95.00 & - & - & 68.60 & - & - & 95.45 & - & - & \textbf{100.00} & - & - & 88.81 & - & -  \\ \cline{2-18}

           & \parbox[t]{2mm}{\multirow{4}{*}{\rotatebox[origin=c]{90}{Handcraft}}} 
           & RANSAC & \underline{99.50} & 74.50 & 27.50 & \underline{98.83} & 70.20 & 32.00 & 98.86 & 78.89 & 35.50 & \textbf{100.00} & \underline{96.50} & 80.50 & \textbf{100.00} & 93.82 & 50.81  \\
           && TEASER++ & \underline{99.50} & 79.50 & 32.00 & 97.67 & \underline{71.21} & 27.00 & 99.43 & 86.93 & 42.00 & \textbf{100.00} & 98.50 & 87.50 & \textbf{100.00} & 96.29 & 59.45  \\
           && Quadro & 98.50 & 91.00 & 51.00 & \textbf{100.00} & \textbf{74.74} & 39.00 & \textbf{100.00} & 92.96 & 50.00 & \textbf{100.00} & 98.50 & 85.00 & \underline{99.25} & 95.06 & 63.78  \\
           && 3DMAC & \underline{99.50} & 66.00 & 15.50 & 95.34 & 51.51 & 15.50 & 98.86 & 64.32 & 20.00 & \textbf{100.00} & 89.00 & 58.00 & \textbf{100.00} & 79.01 & 32.43  \\ \cline{2-18}
           
           &\parbox[t]{2mm}{\multirow{3}{*}{\rotatebox[origin=c]{90}{High.}}}
           & Segregator & 99.00 & 91.00 & 48.00 & 62.21 & 48.48 & 26.00 & \underline{99.43} & 74.37 & 44.00 & \textbf{100.00} & \textbf{99.50} & 87.50 & \textbf{100.00} & 93.80 & 69.73  \\
           && G3Reg & \textbf{100.00} & \underline{95.00} & \underline{56.00} & 88.37 & 67.17 & \underline{50.00} & \textbf{100.00} & \underline{96.48} & \underline{73.50} & \textbf{100.00} & \textbf{100.00} & \textbf{97.50} & \textbf{100.00} & \underline{98.14} & \underline{78.91}   \\
           && QuadricsReg & \underline{99.50} & \textbf{99.50} & \textbf{80.00} & 96.51 & \underline{71.21} & \textbf{53.50} & \textbf{100.00} & \textbf{98.49} & \textbf{87.50} & \textbf{100.00} & \textbf{100.00} & \underline{96.00} & \textbf{100.00} & \textbf{98.77} & \textbf{84.32} \\ \hline 

\parbox[t]{2mm}{\multirow{14}{*}{\rotatebox[origin=c]{90}{KITTI-360-LC}}} && Sequence & \multicolumn{3}{c|}{00} & \multicolumn{3}{c|}{02} & \multicolumn{3}{c|}{04} & \multicolumn{3}{c|}{06} & \multicolumn{3}{c}{09}  \\ \cline{3-18}
            
           & & Level & Easy & Medium & Hard & Easy & Medium & Hard & Easy & Medium & Hard & Easy & Medium & Hard & Easy & Medium & Hard \\ \cline{2-18}

           &\parbox[t]{2mm}{\multirow{4}{*}{\rotatebox[origin=c]{90}{Learning}}}
            & FCGF & 38.50 & 29.00 & 25.50 & 21.00 & 18.00 & 16.50 & 14.00 & 8.00 & 10.00 & 41.00 & 35.00 & 24.50 & 56.00 & 47.50 & 31.00 \\
           && DGR & 39.00 & 26.50 & 22.00 & 20.50 & 16.00 & 14.00 & 16.00 & 9.00 & 6.50 & 40.00 & 34.00 & 20.50 & 56.00 & 42.00 & 23.00 \\
            && PointDSC & 37.20 & 26.50 & 21.00 & 18.50 & 16.50 & 16.00 & 14.50 & 9.50 & 8.00 & 38.50 & 35.00 & 27.00 & 54.00 & 42.50 & 28.50  \\
            && LCDNet & 92.00 & - & - & 89.00 & - & - & 94.50 & - & - & 96.00 & - & - & 94.00 & - & -  \\ \cline{2-18}

           & \parbox[t]{2mm}{\multirow{4}{*}{\rotatebox[origin=c]{90}{Handcraft}}} 
           & RANSAC & \underline{99.50} & 89.50 & 49.00 & \textbf{99.00} & \textbf{87.00} & 55.00 & 97.50 & 77.00 & 32.50 & 97.00 & 71.00 & 34.50 & 98.00 & 78.00 & 35.50  \\
           && TEASER++ & \textbf{100.00} & 91.50 & 56.00 & \underline{98.50} & 83.50 & 54.50 & 97.00 & 75.00 & 36.50 & 96.50 & 70.00 & 35.00 & \underline{99.50} & 83.00 & 34.00  \\
           && Quadro & \textbf{100.00} & 93.50 & 66.00 & 97.50 & 81.00 & \underline{59.50} & \textbf{99.00} & 86.50 & 49.00 & 97.00 & 76.00 & 41.00 & \underline{99.50} & 89.50 & 50.00  \\
           && 3DMAC & 99.50 & 72.00 & 33.50 & 96.00 & 72.50 & 33.00 & 98.00 & 61.50 & 21.00 & 93.00 & 51.50 & 19.50 & 98.50 & 58.00 & 15.50  \\ \cline{2-18}
           
           &\parbox[t]{2mm}{\multirow{3}{*}{\rotatebox[origin=c]{90}{High.}}}
          & Segregator & 97.00 & 86.50 & 62.00 & 94.00 & 67.50 & 41.50 & 91.00 & 64.00 & 34.00 & 87.00 & 52.00 & 20.00 & 97.50 & 74.50 & 30.50  \\
           && G3Reg & \textbf{100.00} & \underline{94.00} & \underline{73.50} & \underline{98.50} & \underline{84.00} & \textbf{64.00} & \textbf{100.00} & \underline{88.00} & \underline{49.50} & \textbf{99.50} & \textbf{79.50} & \underline{43.00} & \textbf{100.00} & \underline{91.50} & \underline{59.50}  \\
           && QuadricsReg & \underline{99.50} & \textbf{95.50} & \textbf{81.00} & 96.50 & 80.00 & 55.50 & \textbf{100.00} & \textbf{91.00} & \textbf{64.50} & \underline{99.00} & \underline{78.00} & \textbf{45.50} & \textbf{100.00} & \textbf{95.50} & \textbf{67.00}  \\ \hline 

            \parbox[t]{2mm}{\multirow{13}{*}{\rotatebox[origin=c]{90}{Apollo-SouthBay-LC}}} && Sequence & \multicolumn{3}{c|}{SunnyvaleBigloop} & \multicolumn{3}{c|}{MathildaAVE} & \multicolumn{3}{c|}{SanJoseDowntown} & \multicolumn{3}{c|}{BaylandsToSeafood} & \multicolumn{3}{c}{ColumbiaPark}  \\ \cline{3-18}
            
           & & Level & Easy & Medium & Hard & Easy & Medium & Hard & Easy & Medium & Hard & Easy & Medium & Hard & Easy & Medium & Hard \\ \cline{2-18}

           &\parbox[t]{2mm}{\multirow{4}{*}{\rotatebox[origin=c]{90}{Learning}}}
            & FCGF & 85.50 & 74.50 & 61.50 & \textbf{100.00} & 4.50 & 4.00 & 92.50 & 90.00 & 77.50 & \underline{93.75} & 4.00 & 5.50 & 66.50 & 65.00 & 50.50  \\
           && DGR & 82.50 & 71.50 & 41.00 & 93.10 & 7.50 & 4.00 & 89.00 & 84.00 & 45.00 & \underline{93.75} & 5.50 & 4.00 & 65.00 & 63.50 & 44.00  \\
           && PointDSC & 78.50 & 63.50 & 36.50 & \textbf{100.00} & 4.00 & 4.00 & 89.00 & 77.50 & 49.50 & 75.00 & 4.00 & 5.00 & 67.00 & 65.00 & 45.00  \\
           && LCDNet & 79.00 & - & - & 89.66 & - & - & 77.00 & - & - & \underline{93.75} & - & - & 92.00 & - & -  \\ \cline{2-18}

           & \parbox[t]{2mm}{\multirow{4}{*}{\rotatebox[origin=c]{90}{Handcraft}}} 
           & RANSAC & \underline{99.50} & 94.50 & 60.50 & \textbf{100.00} & 96.50 & 73.00 & \underline{99.50} & 92.00 & 60.00 & \textbf{93.75} & 88.50 & 75.50 & \textbf{100.00} & \textbf{99.00} & 76.50  \\
           && TEASER++ & \textbf{100.00} & 93.00 & 70.50 & \textbf{100.00} & 96.00 & \underline{81.50} & \textbf{100.00} & 97.00 & 72.00 & \textbf{100.00} & \textbf{92.00} & 85.00 & \textbf{100.00} & 98.50 & 80.50  \\
           && Quadro & \textbf{100.00} & 92.50 & 70.00 & \textbf{100.00} & 92.50 & 80.00 & 99.00 & 96.50 & 76.00 & \textbf{100.00} & 89.50 & \textbf{87.00} & \underline{99.50} & \textbf{100.00} & \underline{84.00}  \\
           && 3DMAC & 98.50 & 82.50 & 32.50 & \underline{96.55} & 83.00 & 50.50 & \underline{99.50} & 81.50 & 38.50 & \underline{93.75} & 76.50 & 60.50 & \textbf{100.00} & 93.50 & 55.00  \\ \cline{2-18}
           
           &\parbox[t]{2mm}{\multirow{2}{*}{\rotatebox[origin=c]{90}{High.}}}
           & G3Reg & \textbf{100.00} & \textbf{100.00} & \textbf{97.50} & \textbf{100.00} & \textbf{99.50} & \textbf{96.50} & \textbf{100.00} & \textbf{100.00} & \textbf{96.50} & \textbf{100.00} & \underline{90.50} & \underline{86.50} & \textbf{100.00} & \textbf{100.00} & \textbf{97.00}  \\
           && QuadricsReg & \textbf{100.00} & \underline{99.50} & \underline{97.05} & \textbf{100.00} & \underline{98.50} & \textbf{96.50} & \textbf{100.00} & \underline{99.00} & \underline{93.50} & \textbf{100.00} & 88.00 & 84.50 & \textbf{100.00} & \textbf{100.00} & \textbf{97.00}  \\
           \bottomrule

        \end{tabular}
    }
\end{table*}

All experiments are conducted on a workstation with an Intel Xeon Platinum 8336C CPU running at 2.30 GHz, 128 GB RAM, and a Nvidia RTX4090 GPU.

\subsubsection{Evaluation Metrics.}
We evaluate the global registration methods from four aspects: Scene representation, correspondence establishment, transformation estimation, and global registration.

For the evaluation of correspondence establishment, the number of correspondences, the inlier ratio (IR), and the correspondence recall (CR) are utilized. Given a pair of point clouds with an established correspondence set $\mathcal{L}$, the number of correspondences is denoted as $|\mathcal{L}|$. The inlier ratio is defined as
\begin{equation}
 \frac{1}{|\mathcal{L}|} \sum_{(i, j) \in \mathcal{L}} \mathds{1}\left(\left\|\bar{\mathbf{R}} \mathbf{t}_{x} + \bar{\mathbf{t}} -\mathbf{t}_{y}\right\| \le 0.5\,\mathrm{m} \right),
\end{equation}
where $(\bar{\mathbf{R}},\bar{\mathbf{t}})$ represents the GT transformation. The GT transformations of all datasets are refined using the Iterative Closest Point (ICP) algorithm. The threshold of 0.5 m defines the distance tolerance within which correspondence is considered an inlier. If the total number of inliers reaches or exceeds 3, the point cloud pair is considered successfully matched. Based on this criterion, the match recall rate can be computed for the dataset.


For the evaluation of transformation estimation, we use the relative translation error ($RTE$) and relative rotation error ($RRE$) as metrics:
\begin{equation}
\begin{aligned}
RTE &= \left\|\bar{\mathbf{t}}-\mathbf{t}\right\|, \\
RRE &= \arccos \left[\frac{\operatorname{tr}\left(\mathbf{R}^\mathrm{T} \bar{\mathbf{R}}\right)-1}{2} \right].
\end{aligned}
\end{equation}

Furthermore, We leverage the registration success rate (SR) to evaluate global registration. In the SLAM application, global point cloud registration is typically employed for loop closure. After obtaining an initial estimate of the transformation, a precise local registration algorithm, such as ICP, is often utilized to refine the pose to a high level of accuracy. Consequently, the primary requirement for global registration is to ensure that the estimated transformation error remains within an acceptable range, enabling the subsequent local registration to converge effectively to the optimal solution. Therefore, we adopt the registration success rate as the metric: If $RTE \leq 2\,\mathrm{m}$ and $RRE \leq 5 ^{\circ}$, the global registration is considered successful.

To evaluate the efficiency of different scene representations for global registration, we assess both storage overhead and computational runtime. Storage overhead is quantified by measuring the storage size per frame of the point cloud representation, while computational runtime is evaluated by the total time required to execute the global point cloud registration algorithm. 

\begin{figure*}[th!]
  \centering
   \includegraphics[width=\linewidth]{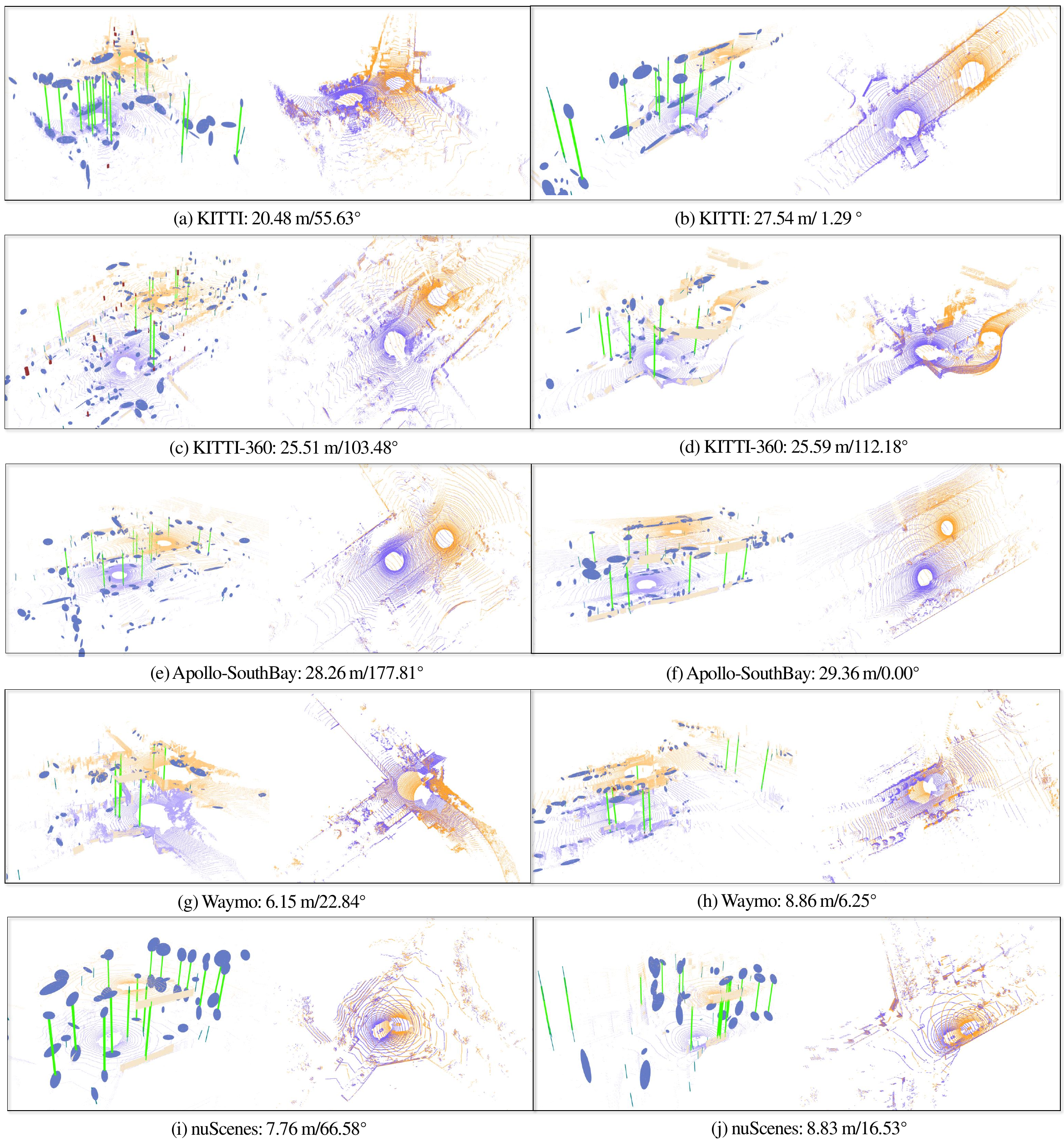}
   \caption{Global point cloud registration results on different datasets. QuadricsReg demonstrates robust performance across diverse datasets, which encompass a variety of scenes and are collected using different LiDAR sensors. Even in cases with considerable viewpoint differences, it ensures adequate matching and achieves accurate alignment by effectively representing, matching, and estimating transformations based on quadrics.}   
   \label{fig: Global point cloud registration results on different datasets.}
\end{figure*}
\subsection{Evaluation of Correspondence Establishment}
In this section, we evaluate the performance of our method in establishing 3D feature correspondences, which involves feature description and matching. We compare the correspondence establishment based on two category representations: low-level and high-level representations. For low-level representations, we employ the handcrafted feature FPFH \citep{Rusu2009FPFH} and learning-based point feature FCGF \citep{Choy2019FCGF}, as they are commonly employed as the front-end in global point cloud registration methods. The final correspondences for these low-level representations are determined using a mutual nearest-neighbor search strategy. 
For high-level representations, we adopt semantic-based Segregator \citep{Yin2023Segregator} and geometric-based G3Reg \citep{Qiao2024G3reg}. For all these learning-based methods, we utilize the officially provided pre-trained models. For QuadricsReg, increasing the upper limit $K_s$ for initial correspondences of each quadric results in a more conservative approximation of all-to-all matching. We set $K_s$ to 15, 20, and 25 to evaluate the effectiveness of our matching strategy based on the quadric similarity.

Table \ref{table: Evaluation of Correspondence Establishment} lists the results of correspondence establishment on the KITTI-LC dataset. In general, methods based on low-level representations achieve relatively higher inlier rates (IR) due to their dense features and the use of a one-to-one matching strategy, resulting in fewer mismatches. In contrast, high-level representation methods tend to have lower inlier rates because of the feature sparsity, which necessitates a one-to-many matching strategy to maintain the number of inliers. Nevertheless, QuadricsReg obtains more competitive correspondence recalls (CR) and registration success rates (SR) for all levels regardless of the lower inlier rates.
As the viewpoint distance increases from easy to hard level, changes in the surrounding points make it challenging for low-level features to establish reliable matches, resulting in reduced correspondence recall and a lower registration success rate. High-level representations, leveraging object-level geometric information, are robust to viewpoint changes, achieving higher correspondence recalls and registration success rates at medium and hard difficulty levels. QuadricsReg shows the best performance for hard levels because of the expressive capability and fitting accuracy of quadrics. Notably, as $K_s$ increases in QuadricsReg, the number of correspondences increases, but the variations in correspondence recall and registration success rate remain minimal. This indicates that the quadric similarity-based matching method effectively establishes potential correspondences without requiring a conservative strategy.

\subsection{Evaluation of Global Registration}

\begin{figure*}[h]
  \centering
\includegraphics[width=0.95\linewidth]{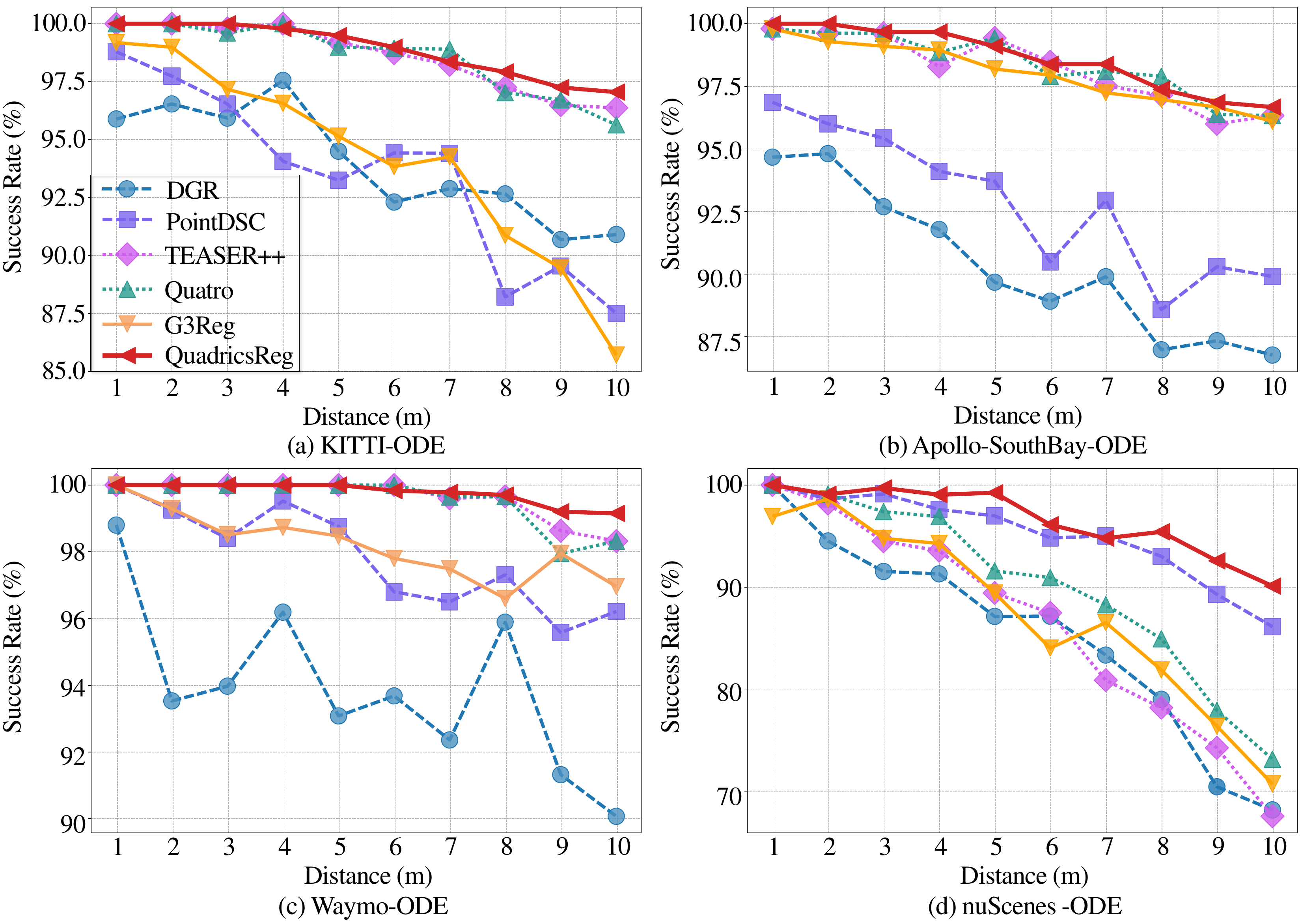}
   \caption{Evaluation of the global registration success rate in LiDAR odometry test on KITTI-ODE, Apollo-SouthBay-ODE, Waymo-ODE, and nuScenes-ODE datasets. } 
   \label{fig: Evaluation of global registration for odometry test.}
\end{figure*}

\subsubsection{Loop Closure Test.}
\label{sect. Loop Closure Test.}
In this section, we assess the performance of various global point cloud registration methods in the loop closure test. We compare three categories of methods: Learning based on low-level representation, handcraft based on low-level representation, and high-level representation-based. For learning methods, we select FCGF, DGR \citep{Choy2020DGR}, PointDSC \citep{Bai2021Pointdsc}, and LCDNet \citep{Cattaneo2022LCDNet}. FCGF employs RANSAC as the backend, whereas DGR, PointDSC, and LCDNet are all end-to-end methods, with LCDNet specifically designed for loop detection. For handcraft methods, we evaluate RANSAC, TEASER++ \citep{Yang2021TEASER}, Quatro \citep{Lim2022Quatro}, and 3DMAC \citep{Zhang20233DMAC} on the FPFH-based matching results. We compare high-level representations with the semantic-based Segregator and the geometry-based G3Reg. 

The results are presented in Table \ref{table: Evaluation of global registration for loop closure test}. Learning-based methods generally exhibit inferior performance compared to handcraft and high-level object-based approaches. This is primarily because the learning-based methods are trained on registration datasets involving continuous viewpoint variations. In contrast, loop closure scenarios feature greater diversity in translational and rotational changes. At the easy level, point-based handcraft-designed methods achieve favorable registration success rates around $100\%$. As viewpoint distance increases, although the registration success rate decreases for all methods, those based on high-level representations generally exhibit a smaller performance decline than point-based learning and handcraft-designed methods. This is because the integration of object-level semantic and geometric information by high-level representations mitigates the challenges of matching under significant viewpoint differences. Consequently, these high-level object-based methods establish more stable correspondences under challenging conditions, thereby ensuring robust point cloud registration. 

Among the high-level representation methods, G3Reg and QuadricsReg outperform Segregator, highlighting the advantages of initial matching based on object geometric attributes and the filtering of outliers through multi-level compatibility graphs. QuadricsReg demonstrates superior performance, particularly at larger viewpoint differences for hard levels. As shown in Fig. \ref{fig: Global point cloud registration results on different datasets.}a-h, it models different semantic elements in the scene using distinct geometric primitives, which not only reduces the matching search space but also provides reliable geometric similarity measures for later matching. In sparse-object environments, such as the KITTI-LC sequence 02 (Fig. \ref{fig: Global point cloud registration results on different datasets.}b), QuadricsReg leverages the augmented points to effectively compensate for the reduced number of matches caused by the sparsity of key elements. In the KITTI-LC sequence 08 and KITTI-360-LC, most loops are reversed, where the opposite direction poses challenges to estimating the geometric properties of objects, QuadricsReg nonetheless maintains its robustness. The Apollo-SouthBay-LC dataset lacks semantic annotations, preventing reliable semantic prediction, which challenges the quadric representation of scenes. Nevertheless, due to the abundance of geometric features in this dataset, using geometric clustering categories as semantic labels still yields commendable performance.

\subsubsection{Odometry Test with Random Rotations.}
\begin{figure*}[h]
  \centering
   \includegraphics[width=\linewidth]{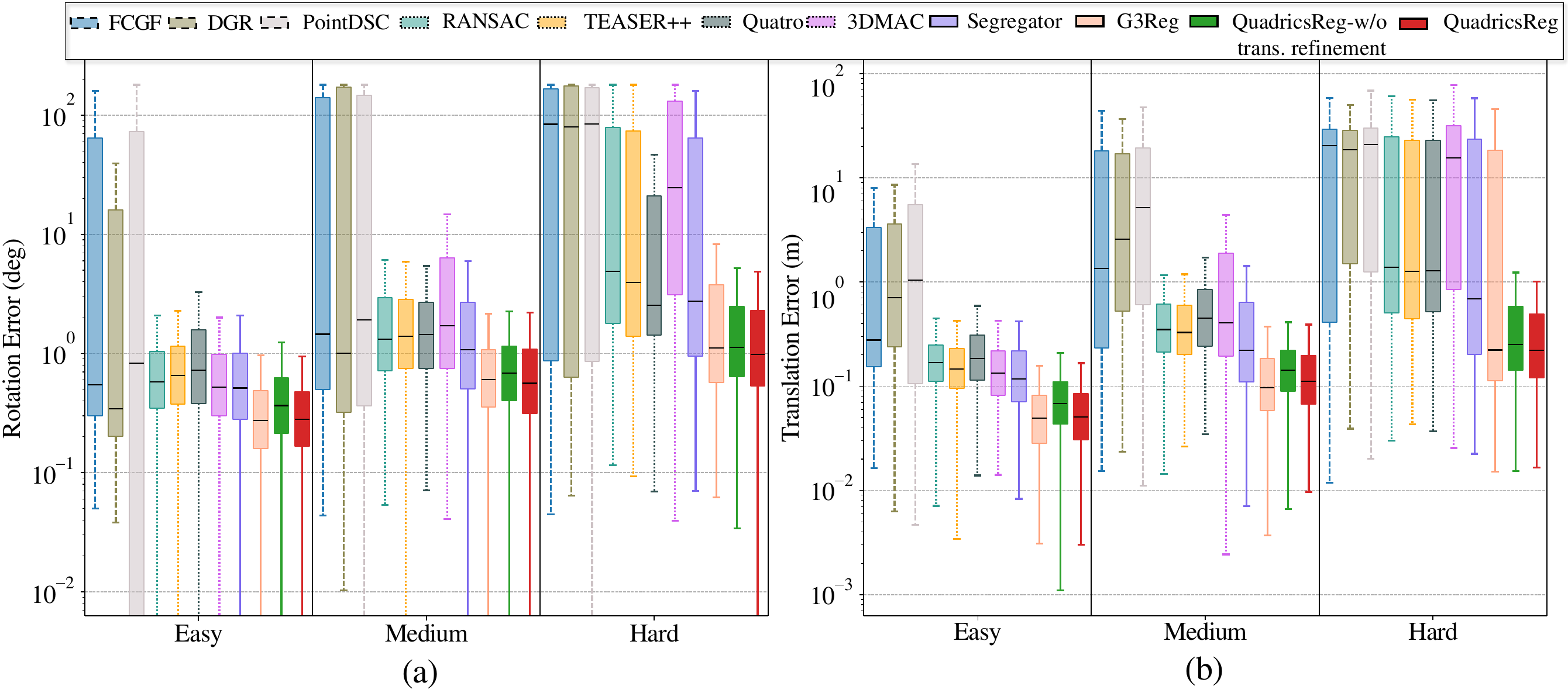}
   \caption{Evaluation of transformation estimation on KITTI-LC dataset (Left: RRE, right:RTE). }
   \label{fig: Evaluation of transformation estimation on KITTI-LC.}
\end{figure*}

In this section, we evaluate the performance of point cloud registration methods in the odometry test. Since variations on rotation are small in odometry scenarios, the estimated rotation may not represent the true performance of global registration \citep{Qiao2024G3reg}. Therefore, we augment the point clouds by applying random rotations in the yaw direction within the range of $[-45^{\circ}, 45^{\circ}]$. We select the learning-based methods DGR and PointDSC, the handcraft methods TEASER++ and Quatro, and the high-level representation-based methods G3Reg and QuadricsReg for comparison. 

The results are presented in Fig. \ref{fig: Evaluation of global registration for odometry test.}. The learning-based DGR and Point DSC show the lowest success rates on the 4 datasets. TEASER++ and Quatro have a good performance on the first 3 datasets but show much worse results on the last nuScenes dataset. This is because the nuScenes is collected using a 32-beam LiDAR, while the other 3 use 64-beam LiDAR. G3Reg has unstable performance across different datasets, the reason is that G3Reg is not robust to the random rotation argumentation. Among these methods, QuadricsReg obtains the best performance in all 4 datasets. Additionally, its registration success rate remains stable with the increase of the viewpoint distance between paired point clouds. These results demonstrate the effectiveness of QuadricsReg on the odometry configurations.


Furthermore, we present the qualitative registration performance of QuadricsReg in the odometry test in Fig. \ref{fig: Global point cloud registration results on different datasets.}g-j. The addition of random rotations further impacts the geometric attributes of objects in the scene, such as their orientation, and the odometry test dataset also features numerous scenes with sparse objects, both issues pose significant challenges for the proposed QuadricsReg. Despite these challenges, QuadricsReg demonstrates robust performance due to its capability to accurately estimate the geometric attributes of underlying surfaces from point clouds and effectively fit these surfaces as quadrics.  Moreover, experiments conducted on the nuScenes-ODO dataset, collected using a 32-beam LiDAR, further validate QuadricsReg's robustness to variations in spatial arrangement and point cloud density due to different sensor types, as also illustrated in Fig. \ref{fig: Global point cloud registration results on different datasets.}i-j.

\subsection{Evaluation of Transformation Estimation}

In this section, we assess the transformation estimation performance of various global registration methods. We continue to compare the three categories of methods. The RRE and RTE results are shown in Fig. \ref{fig: Evaluation of transformation estimation on KITTI-LC.}. The transformation errors generally grow higher from easy to hard viewpoint distance for all methods. The learning-based methods exhibit the largest error variances no matter for RRT or RTE. Methods based on high-level representations generally obtain lower rotation and translation errors compared to those utilizing lower-level representations. QuadricsReg achieves competitive performance even before transformation optimization, primarily due to the effectiveness of its quadric-based representation and matching, resulting in more accurate correspondences and subsequently precise transformation estimation. After incorporating transformation optimization, the errors are further reduced, surpassing the performance of other methods. This outcome not only demonstrates the effectiveness of the optimization process but also proves the robustness of quadrics to the geometric degeneracy. During the quadrics distance computation, distances between quadrics are evaluated only for non-degenerate attributes, thereby enhancing the validity of both the optimization objective and the criteria for selecting the optimal transformation.

\subsection{Evaluation of Scene Representation}
\begin{table*}[t!]
    \captionsetup{font=footnotesize}
    \centering
    \caption{Ablation study on KITTI-LC dataset (registration success rate, unit: \%).}
    \label{table: Ablation_study}
    \setlength{\tabcolsep}{3pt}
    {\scriptsize
        \centering
        \begin{tabular}{l|l|ccc|ccc|ccc|ccc|ccc}
            \toprule
            & Sequence & \multicolumn{3}{c|}{00} & \multicolumn{3}{c|}{02} & \multicolumn{3}{c|}{05} & \multicolumn{3}{c|}{06} & \multicolumn{3}{c}{08}  \\ \cline{2-17}
            
            & Level & Easy & Medium & Hard & Easy & Medium & Hard & Easy & Medium & Hard & Easy & Medium & Hard & Easy & Medium & Hard \\ \cline{1-17}

           \parbox[t]{2mm}{\multirow{8}{*}{\rotatebox[origin=c]{90}{Representation}}}
            & W/O semantics-aided & \textbf{100.00} & 93.50 & 45.00 & \underline{95.93} & 69.70 & 51.50 & \textbf{100.00} & 93.97 & 62.50 & \textbf{100.00} & \underline{99.50} & 94.00 & \textbf{100.00} & 96.91 & 70.27  \\ \cline{2-17}
           & W/O central quadrics & 98.50 & 80.00 & 40.50 & 61.05 & 50.51 & 38.00 & 89.20 & 57.79 & 28.00 & \textbf{100.00} & 99.00 & 85.00 & 97.01 & 83.33 & 58.38  \\ 
            & W/O linear-center quadrics & \underline{99.50} & 95.50 & 55.00 & 92.44 & 56.57 & 30.50 & \textbf{100.00} & 95.48 & 76.00 & \textbf{100.00} & \textbf{100.00} & 96.00 & \textbf{100.00} & 95.06 & 67.57  \\
            & W/O planar-center quadrics & \underline{99.50} & 98.00 & 68.50 & 94.77 & 61.11 & 40.00 & \textbf{100.00} & 95.98 & 79.50 & \textbf{100.00} & \textbf{100.00} & 88.50 & \textbf{100.00} & 97.53 & 76.22   \\ \cline{2-17}
            & W/O augmented points & \underline{99.50} & 98.00 & 71.00 & 86.63 & 66.67 & 50.50 & \textbf{100.00} & \textbf{98.49} & 84.50 & \textbf{100.00} & \textbf{100.00} & 95.00 & \textbf{100.00} & \underline{98.77} & 83.24  \\ \cline{2-17}
           & $K_e=30$ & \textbf{100.00} & 97.50 & 70.50 & 94.77 & \textbf{71.21} & 52.50 & \textbf{100.00} & 97.49 & 78.50 & \textbf{100.00} & \textbf{100.00} & 88.00 & \textbf{100.00} & 98.15 & 77.84  \\
           & $K_e=40$ & \underline{99.50} & 98.00 & 72.50 & 94.77 & \textbf{71.21} & 52.50 & \textbf{100.00} & 97.49 & 80.00 & \textbf{100.00} & \textbf{100.00} & 93.50 & \textbf{100.00} & \underline{98.77} & 80.00 \\
           & $K_e=60$ & \underline{99.50} & 98.00 & 72.00 & 94.77 & \underline{70.71} & 52.50 & \textbf{100.00} & \underline{97.99} & 84.00 & \textbf{100.00} & \textbf{100.00} & 95.00 & \textbf{100.00} & \underline{98.77} & \textbf{86.49}  \\ \cline{1-17}

            \parbox[t]{2mm}{\multirow{7}{*}{\rotatebox[origin=c]{90}{Corr. Estab.}}} 
           & $K_s=15$ & \underline{99.50} & 98.00 & 71.50 & 94.77 & 70.20 & 52.50 & \textbf{100.00} & 98.49 & 84.00 & \textbf{100.00} & \textbf{100.00} & 95.00 & \textbf{100.00} & \underline{98.77} & 82.70  \\
           & $K_s=25$ & \textbf{100.00} & \underline{99.00} & \textbf{84.00} & 93.60 & \textbf{72.21} & \underline{53.00} & \textbf{100.00} & 96.98 & \textbf{88.00} & \textbf{100.00} & \textbf{100.00} & \textbf{98.00} & \textbf{100.00} & \textbf{100.00} & \textbf{86.49}  \\ \cline{2-17}
           
           & $\delta_m: [0.2]$ & \underline{99.50} & 91.00 & 47.00 & 87.79 & 60.10 & 34.00 & \textbf{100.00} & 91.46 & 56.50 & \textbf{100.00} & 99.00 & 82.00 & \underline{99.25} & 94.44 & 65.41  \\
           & $\delta_m: [0.8]$ & 99.00 & 97.50 & 61.50 & 94.19 & \underline{70.71} & 49.50 & \textbf{100.00} & 95.48 & 77.50 & \textbf{100.00} & \textbf{100.00} & 88.50 & \textbf{100.00} & \underline{98.77} & 78.92  \\ 
           & $\delta_m: [0.2,0.8]$ & \underline{99.50} & 97.00 & 70.00 & 94.77 & 68.18 & 52.50 & \textbf{100.00} & \underline{97.99} & 80.50 & \textbf{100.00} & \textbf{100.00} & 94.50 & \textbf{100.00} & \underline{98.77} & 81.08  \\ 
           & $\delta_m: [0.2,0.6,0.8]$ & \underline{99.50} & 97.50 & 71.50 & 94.77 & 69.70 & \underline{53.00} & \textbf{100.00} & \textbf{98.49} & 83.50 & \textbf{100.00} & \textbf{100.00} & 93.50 & \textbf{100.00} & \underline{98.77} & 82.70  \\ 
           & $\delta_m: [0.2,0.4,0.6,0.8,1.0]$ & 99.50 & 98.00 & 76.00 & 94.77 & \textbf{71.21} & \underline{53.00} & \textbf{100.00} & \underline{97.99} & 84.50 &\textbf{ 100.00} & \textbf{100.00} & 95.00 & \textbf{100.00} & \underline{98.77} & \underline{84.86}  \\ \cline{1-17}
           
           \parbox[t]{2mm}{\multirow{4}{*}{\rotatebox[origin=c]{90}{Trans. Est.}}}
           & W/O $\mathbf{e_R}$ & \underline{99.50} & 98.00 & 71.50 & 95.35 & \underline{70.71} & 52.50 & \textbf{100.00} & \textbf{98.49} & 84.00 & \textbf{100.00} & \textbf{100.00} & 95.00 & \textbf{100.00} & \underline{98.77} & 83.24   \\
           & W/O $\mathbf{e_t}$ & \underline{99.50} & 98.00 & \underline{73.00} & 94.19 & \underline{70.71} & \underline{53.00} & \textbf{100.00} & \underline{97.99} & 84.00 & \textbf{100.00} & \textbf{100.00} & 95.00 & \textbf{100.00} & \underline{98.77} & 82.16 \\ 
           & W/O augmented points & \underline{99.50} & 98.00 & 72.00 & 94.19 & 69.19 & \underline{53.00} & \textbf{100.00} & 97.49 & 84.50 & \textbf{100.00} & \textbf{100.00} & 95.00 & \textbf{100.00} & \underline{98.77} & 82.70 \\
           & W/O trans. refinement & \underline{99.50} & 98.00 & 72.00 & 93.60 & 69.70 & 52.50 & \textbf{100.00} & \underline{97.99} & 85.00 & \textbf{100.00} & \textbf{100.00} & 94.50 & \textbf{100.00} & \underline{98.77} & 82.16  \\ \cline{1-17}

           & Default & \underline{99.50} & \textbf{99.50} & \underline{80.00} & \textbf{96.51} & \textbf{71.21} & \textbf{53.50} & \textbf{100.00} & \textbf{98.49} & \underline{87.50} & \textbf{100.00} & \textbf{100.00} & \underline{96.00} & \textbf{100.00} & \underline{98.77} & 84.32  \\
           
           \bottomrule
        \end{tabular}
    }
\end{table*}

\begin{figure}[h]
  \centering
   \includegraphics[width=\linewidth]{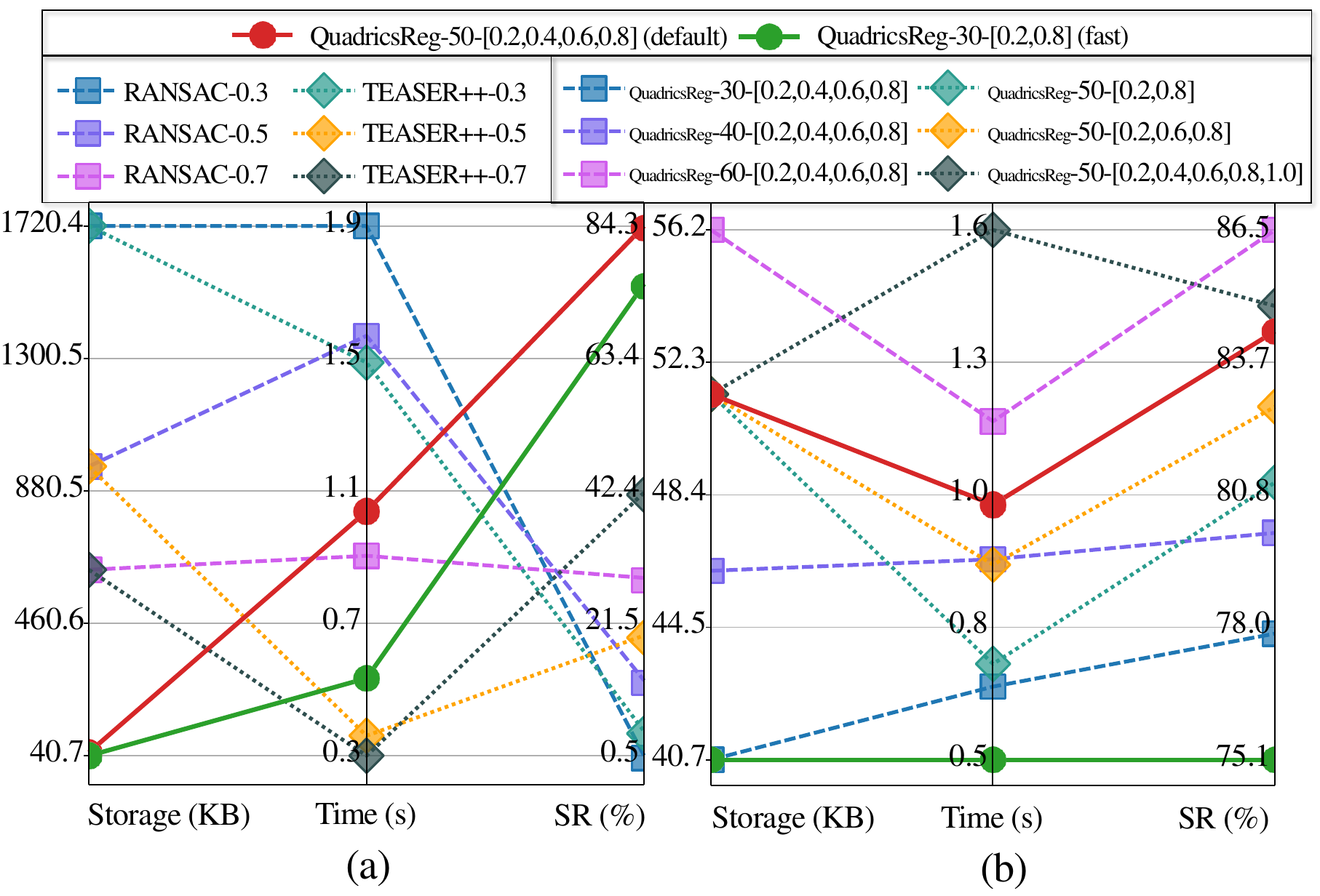}
   \caption{Efficiency evaluation of scene representation on the hard level of the KITTI-LC sequence 08. The storage results are measured by uniformly storing the point-based or quadric-based scene representation data in text format.}   
   \label{fig: Efficiency evaluation of scene representation on the KITTI-LC 08 sequence.}
\end{figure}

In this section, we evaluate the efficiency of different scene representations for global point cloud registration. We select RANSAC and TEASER++, both of which utilize low-level point representations with the FPFH descriptor matching. To assess the impact of point density on registration efficiency and success rate, we set the voxel down-sampling rate of 0.3, 0.5, and 0.7, respectively. For QuadricsReg, we set the upper limit $K_e$ of quadrics for each semantic type to 30, 40, 50, and 60 to investigate the influence of quadrics density. Furthermore, various thresholds are configured during the consistency-checking process to construct compatibility graphs at levels 1, 2, 3, 4, and 5, which are used to evaluate the impact of the number of graphs. In addition to the default configuration of QuadricsReg, we also configure a fast version with reduced quadrics density and fewer compatibility graph levels. All methods are implemented in Python on the same hardware to mitigate the efficiency differences caused by programming languages. For RANSAC with FPFH, we use the implementation available in the Open3D library \citep{Qian-Yi2018Open3D}, while for TEASER++, we utilize the official implementation. The comparison results are shown in Fig. \ref{fig: Efficiency evaluation of scene representation on the KITTI-LC 08 sequence.}. 

In terms of storage overhead, the quadric-based representation exhibits significantly reduced requirements compared to point-based representation, with each frame's storage size being less than one-twentieth of that of point representation at a 0.5 down-sampling rate. Regarding computational efficiency, the default QuadricsReg demonstrates performance comparable to RANSAC-0.7, while the fast is slightly slower than TEASER++-0.5. Notably, both versions of QuadricsReg outperform the other methods in terms of registration accuracy.

For QuadricsReg, given the same compatibility graph level, an increase in quadrics density $K_e$ leads to greater storage and time overhead but also results in an improved registration success rate. When the quadric density is fixed, increasing the number of compatibility graph levels requires more computational time to iteratively determine the maximum clique. Particularly, as the consistency check threshold increases, the number of vertices in the graph also increases, which necessitates a more extensive maximum cliques search cost. Considering the trade-off between storage and time overhead versus registration success rate, we configure both default and fast versions of QuadricsReg to accommodate varying speed requirements. In the SLAM framework, global point cloud registration is primarily employed for loop closure and is executed only after loop detection. The QuadricsReg is capable of running efficiently for loop closure and pose optimization.   

\subsection{Ablation Study}

This section presents the ablation study focusing on representation, correspondence establishment, and transformation estimation to demonstrate the contributions of key modules in QuadricsReg and validate the rationality of parameter settings. Results on the KITTI-LC dataset are shown in Table \ref{table: Ablation_study}, with the last row representing the default settings.

\begin{figure*}[th!]
  \centering
   \includegraphics[width=\linewidth]{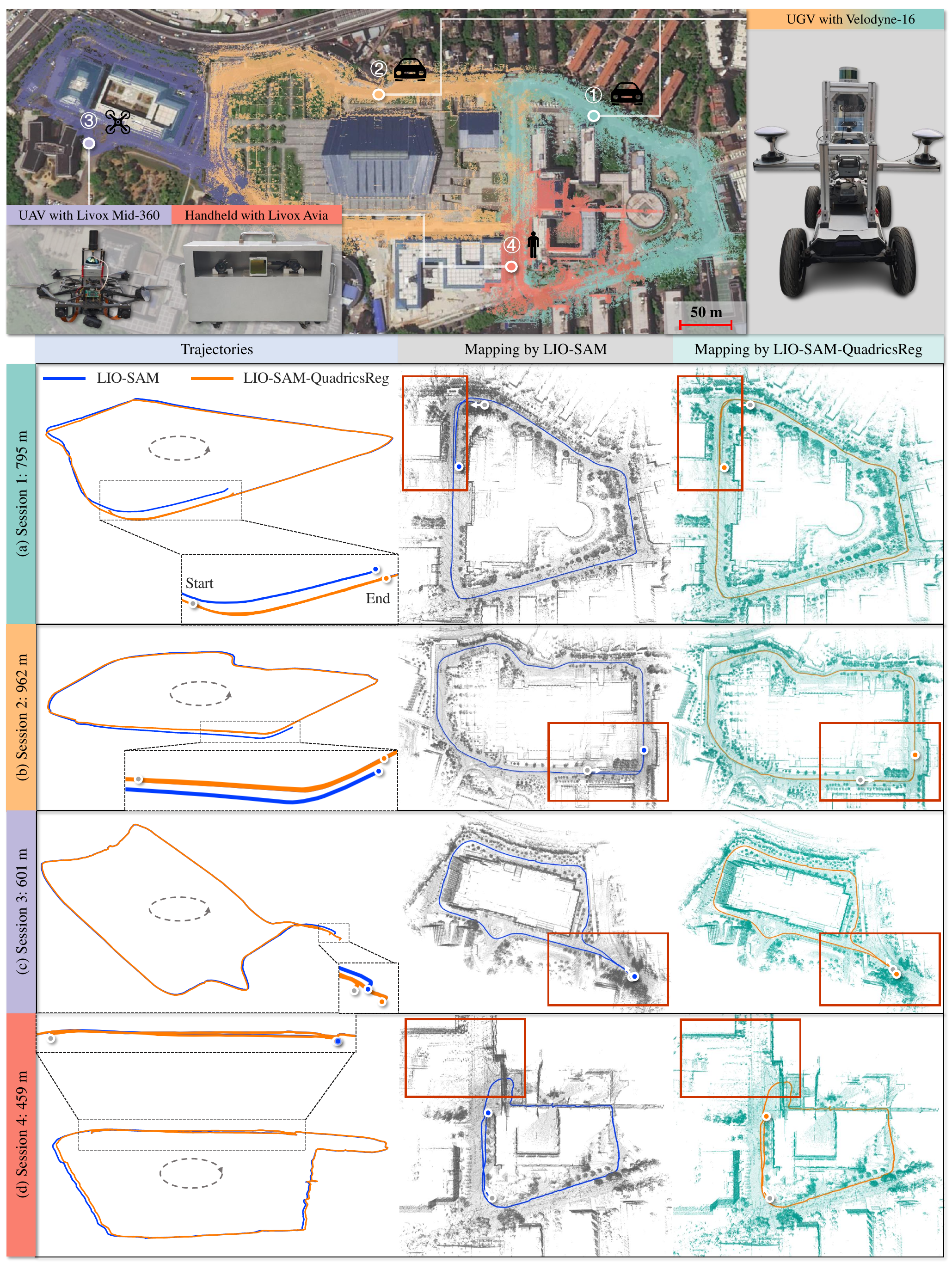}
   \caption{Localization and mapping results on the self-collected Hetero-Reg dataset with QuadricsReg as the loop closure module. We collect four sessions, totaling approximately 2.8 kilometers within a campus environment using a UGV, UAV, and handheld platform, each equipped with different LiDAR sensors. QuadricsReg effectively corrects trajectory drift caused by long-term errors in LiDAR odometry, resulting in more accurate localization and mapping performance.
   }   
   \label{fig:Hetero-Reg}
\end{figure*}

\begin{figure*}[!t]
  \centering
   \includegraphics[width=\linewidth]{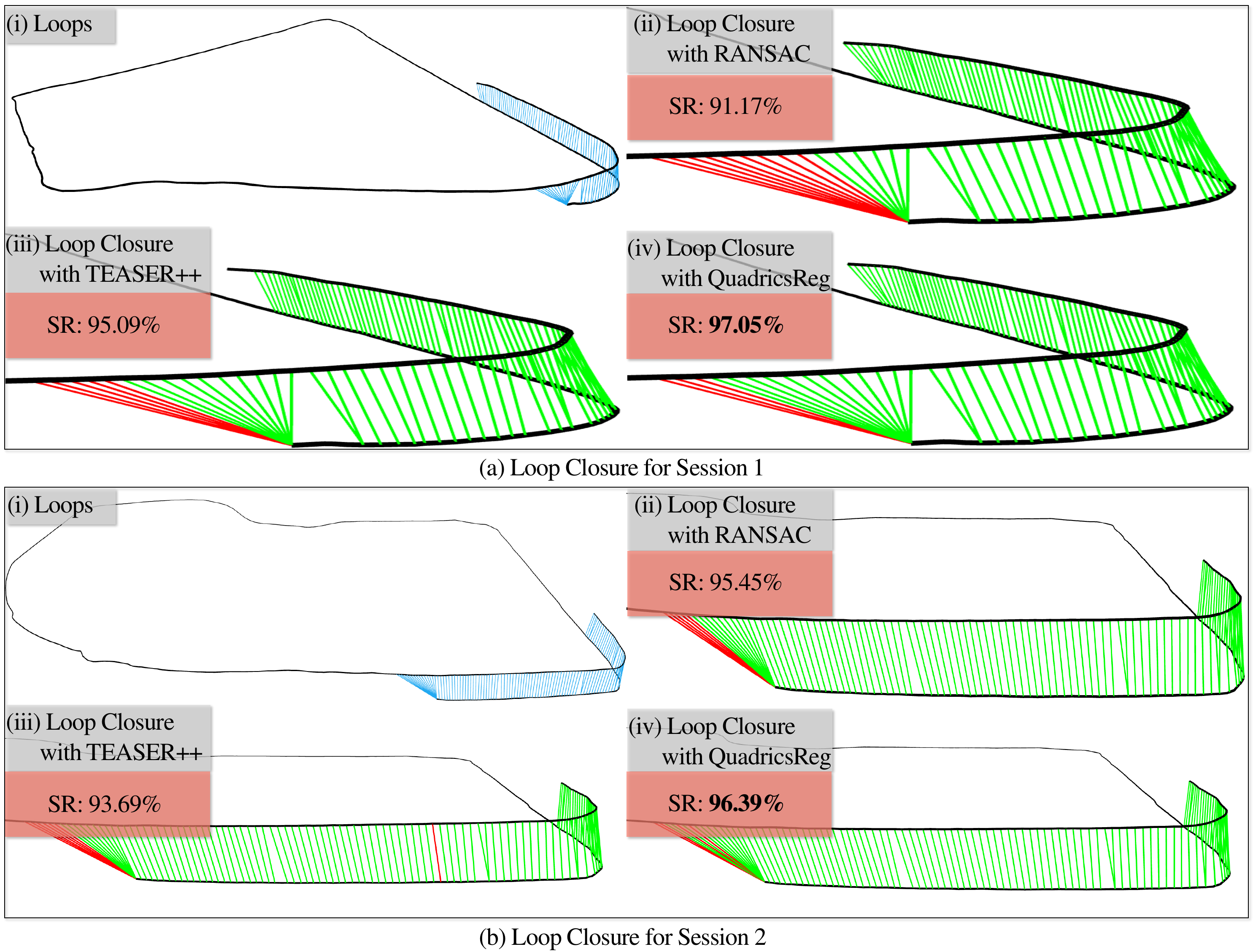}
   \caption{Evaluation of loop closure on session 1 and session 2 of Hetero-Reg dataset. The black lines represent the trajectories, elevated for clarity in loop visualization. Sub-figure (i) depicts the complete trajectory of sessions, with blue lines indicating loop frames identified by GT position. In the remaining sub-figures, zoomed-in views of loop regions are presented, where green lines denote successful loop closure by the global registration method, and red lines indicate failures. }   
   \label{fig: Evaluation of loop closure on session 1 and session 2 of Hetero-Reg dataset.}
\end{figure*}
For representation, we examine the impact of semantic assistance, different center types of quadrics, point augmentation, and density of quadrics. Comparing the version without semantics and the default one, it can be observed that incorporating semantic information outputs more stable success rates from easy to hard levels, this is because the semantic information can improve the primitive extraction and reduce mismatches. We further evaluate the importance of different quadric types on global registration. Central quadrics effectively represent distinct, isolated elements, providing constraints in three directions, thus significantly impacting registration. Linear-center and planar-center quadrics provide constraints in two and one direction respectively, leading to a lesser effect. However, the omission of any type results in inadequate scene representation, which adversely downgrades registration performance, particularly at considerable viewpoint distances for the hard level. Notably, the point augmentation in representation significantly benefits KITTI-LC sequence 02, which consists of highway scenes with sparse, extended, and repetitive elements. Furthermore, increasing the upper limit on the number of semantic quadrics enhances the comprehensiveness of the scene representation, leading to a higher registration success rate. However, as illustrated in Fig. \ref{fig: Efficiency evaluation of scene representation on the KITTI-LC 08 sequence.}, this also incurs increased storage and computational overhead. Consequently, $K_e = 50$ is selected as a balanced default configuration.

For correspondence establishment, we discuss the impact of the upper matching candidate bound $K_s$ for each quadric and the consistency check threshold $\delta_m$. Increasing the upper bound $K_s$ generally enhances registration performance. As discussed in Table \ref{table: Evaluation of Correspondence Establishment}, a larger $K_s$ results in more correspondences, and we find that $K_s=20$ is sufficient with high efficiency. The consistency check threshold is typically challenging to determine, and a single-level consistency graph derived from a single threshold often fails to identify inliers optimally. The multi-level graph provides adequate tolerance for observation errors in key elements, thereby contributing to a higher registration success rate. However, as shown in Fig. \ref{fig: Efficiency evaluation of scene representation on the KITTI-LC 08 sequence.}, this also leads to increased computational time overhead associated with maximum clique searches. Furthermore, larger consistency check thresholds may lead to suboptimal outlier pruning, which could result in erroneous correspondences and, as a consequence, reduced registration performance. To balance the efficiency and accuracy, we employ four consistency check thresholds, with the maximum threshold capped at 0.8.

For transformation estimation, we investigate the influence of rotation distance, translation distance between quadrics, and point augmentation within the optimization process. Additionally, we evaluated the effect of the refinement process itself on transformation estimation. Experimental results demonstrate that rotation and translation distance are all important for maintaining a high registration success rate. Including augmented points helps mitigate the reduced contribution of degenerate quadrics during the optimization process. Furthermore, if the transformation optimization component is removed from the default QuadricsReg, the registration success rates are decreased, which is more obvious for hard levels. The transformation errors are reported in Fig. \ref{fig: Evaluation of transformation estimation on KITTI-LC.} for QuadricsReg without pose refinement. We can observe that the refinement process can reduce the transformation errors for all levels. 

\begin{figure*}[th!]
  \centering
   \includegraphics[width=\linewidth]{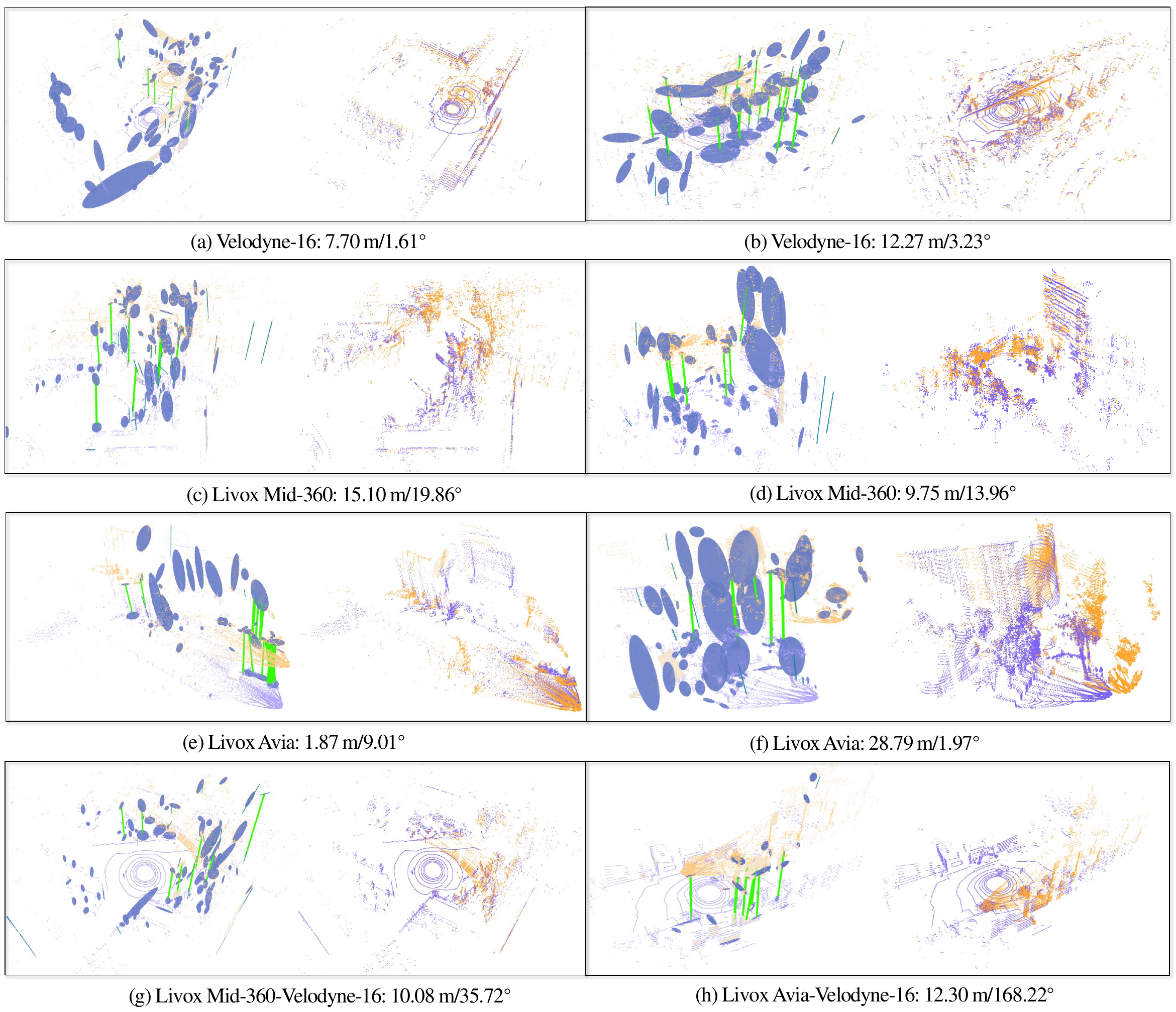}
   \caption{Real-world global registration results on point clouds scanned by different LiDAR sensors. QuadricsReg robustly conducts quadric-based representation, matching, and transformation estimation across various viewpoints, densities, and FOVs.}   
   \label{fig: Global registration results on point clouds scanned by different LiDAR sensors.}
\end{figure*}

\begin{figure*}[th!]
  \centering
   \includegraphics[width=\linewidth]{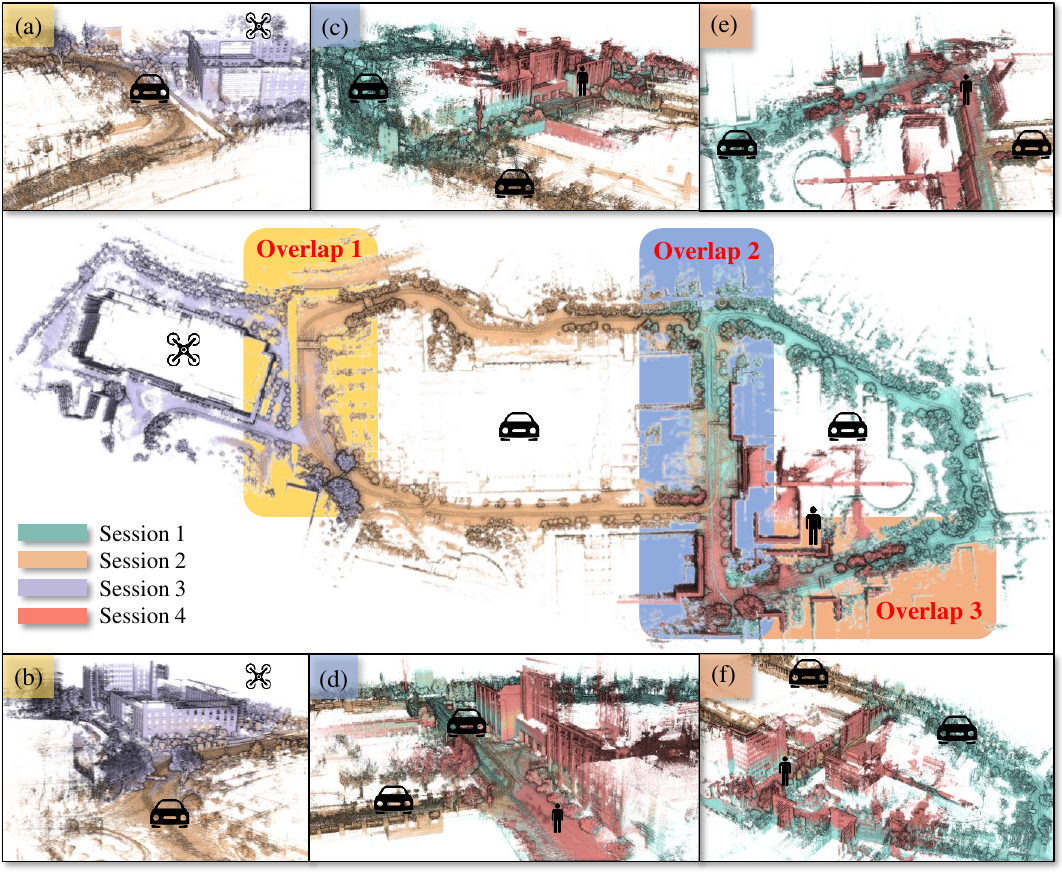}
   \caption{Multi-session mapping on Hetero-Reg dataset. QuadricsReg is utilized for global point cloud registration within the same session for the same LiDAR and across sessions for different LiDARs. Accurate map merging results demonstrate the effectiveness of QuadricsReg.}   
   \label{fig: Multi-session Mapping on Hetero-Reg dataset.}
\end{figure*}
\subsection{Real-World Applications}

\label{sect: Real-World Applications}
To validate the applicability of QuadricsReg in real-world applications, as illustrated on the top of Fig. \ref{fig:Hetero-Reg}, we collected four map sessions on the Wuhan University campus with approximately 2.8 kilometers scenarios, to form the Hetero-Reg dataset, which features various LiDAR sensors and robot platforms. Specifically, the Velodyne-16 is mounted on the UGV, the Livox Mid-360 is on the UAV, and the Livox Avia is on the handheld platform. The UGV is equipped with the Real-Time Kinematic (RTK) system for GT position. We conduct single-session loop closure tests in SLAM and multi-session mapping tests to evaluate the mapping capabilities of the proposed QuadricsReg on challenging scenarios.

\subsubsection{Loop Closure in SLAM.}
We integrate QuadricsReg as a loop closure module into LIO-SAM \citep{liosam2020shan} to evaluate its effectiveness for localization and mapping. LIO-SAM utilizes a radius search strategy to identify revisited locations and generate loop candidates. Upon loop detection, QuadricsReg estimates the transformation between point clouds. The loop closure is then incorporated as constraints into the pose graph for further map optimization. However, inaccuracies in pose estimation can lead to the rejection of the loop closure constraints, resulting in suboptimal optimization. As shown in Fig. \ref{fig:Hetero-Reg}, we conduct tests on all four sessions of the Hetero-Reg dataset. The results indicate that incorporating QuadricsReg for loop closure allows the trajectory to be correctly aligned from start to end, effectively eliminating map inconsistencies in revisited areas. In particular, the significant z-axis drift in the LIO-SAM odometry for sessions 1 and 2 is successfully corrected by applying QuadricsReg for pose graph optimization. In sessions 3 and 4, the shorter distances result in minimal drift along the odometry. However, map inconsistencies still exist in revisited areas, which are also eliminated after using the QuadricsReg-based pose graph optimization.

To quantitatively evaluate the loop closure performance of QuadricsReg on the real robot SLAM application, we generate loop candidates for sessions 1 and 2 using RTK-provided positioning information with a search radius of 5 m. We select two typical competitors, \emph{i.e.}, RANSAC and TEASER++. The results are shown in Fig. \ref{fig: Evaluation of loop closure on session 1 and session 2 of Hetero-Reg dataset.}. QuadricsReg obtains the highest success rate (SR) for two sessions, demonstrating more robust performance even with relatively sparse point clouds. 

Furthermore, we test the generalization ability of QuadricsReg on different LiDAR sensors. The registration examples of QuadricsReg on different LiDAR types are presented in Fig. \ref{fig: Global registration results on point clouds scanned by different LiDAR sensors.}a-f. QuadricsReg effectively represents point cloud scenes acquired from various LiDAR types, accounting for spatial arrangement, density, and FOV differences, while establishing precise correspondences and achieving robust registration. Specifically, the narrow view of the Livox Avia (horizontal FOV of approximately $70^\circ$) poses challenges for object clustering, leading to slight inaccuracies in scene representation. Nevertheless, QuadricsReg successfully achieves robust point cloud registration, demonstrating the robustness of its backend matching and pose estimation module.

\subsubsection{Multi-session Mapping.}
Finally, we apply QuadricsReg to multi-session mapping, focusing on evaluating the performance of the global registration algorithm across heterogeneous robotic platforms and different LiDAR sensors. Fig. \ref{fig: Global registration results on point clouds scanned by different LiDAR sensors.}g-h present examples of QuadricsReg registration across different LiDAR types, indicating that representing the scene using quadrics mitigates the registration challenges posed by differing point cloud distributions from various LiDAR sensors. Based on these registration results, we conduct the multi-session mapping, as shown in Fig. \ref{fig: Multi-session Mapping on Hetero-Reg dataset.}. The local session maps are constructed using LIO-SAM, integrating QuadricsReg for loop closure. Fig. \ref{fig: Multi-session Mapping on Hetero-Reg dataset.}a-f illustrate the map merging results in three overlapping regions, demonstrating that maps from heterogeneous robotic platforms align effectively despite discrepancies in their operational workspaces. Therefore, QuadricsReg can provide effective transformations for map merging tasks on heterogeneous robotic platforms equipped with different sensors, even with significant viewpoint differences and inconsistent point distributions.
For merging multiple sessions, QuadricsReg is triggered when overlaps between robots' trajectories are detected. The estimated transformation using QuadricsReg is incorporated into the pose graph for global map optimization. Finally, we obtain the merged large-scale map of 4 sessions under different LiDAR sensors and heterogeneous robot platforms, demonstrating the capability of QuadricsReg on point cloud registration of large-scale scenarios.

\section{Conclusion}
\label{sec6:conclusion}

This paper introduces QuadricsReg, a robust point cloud registration method utilizing the concise quadric representation. It models the primary primitives of large-scale point clouds as quadrics, transforming point cloud registration into a graph-matching problem based on these primitives. QuadricsReg identifies correspondences by leveraging intrinsic characteristics and geometric consistency between quadrics. It estimates the final transformation between aligned point clouds, further optimized in a factor graph using quadric degeneracy-aware distance. Extensive experiments of correspondence establishment and global point cloud registration are conducted on five public datasets. The exceptional registration success rates and minimal registration errors demonstrate the effectiveness and robustness of QuadricsReg. The real-world testings for loop closure and multi-session mapping on our self-collected heterogeneous dataset demonstrate the generalization ability and robustness of QuaricsReg on different LiDAR sensors and robot platforms.


There are still limitations to the proposed QuatricsReg. Pure LiDAR-based semantic and geometric formulation for quadrics detection and matching is not robust for all scenarios, particularly when significant disparities in point distribution arise due to notably low overlaps or differences in LiDAR types. Therefore, the fusion of rich image texture to enhance LiDAR geometry for better quadrics detection and matching will be a promising future direction. Additionally, although quadrics can still model moving objects in dynamic environments, the geometric consistency between quadrics no longer exists, yielding noisy connections in the matching graph. Therefore, another direction is to estimate the state of dynamic objects and remove them in the quadrics matching graph for better registration robustness to dynamic scenes.

\section*{Declaration of conflicting interests}

The author(s) declared no potential conflicts of interest with respect to the research, authorship, and/or publication of this article.

\section*{Funding}
The author(s) disclosed receipt of the following financial support for the research, authorship, and/or publication of this article: This work was partially supported by the NSFC grants under contracts Nos. 62301370, 62325111 and U22B2011, and the Wuhan University-Huawei Geoinformatics Innovation Laboratory.

{
\bibliographystyle{SageH}
\bibliography{main}

\begin{thebibliography}{76}
\providecommand{\natexlab}[1]{#1}
\providecommand{\url}[1]{\texttt{#1}}
\providecommand{\urlprefix}{URL }
\expandafter\ifx\csname urlstyle\endcsname\relax
  \providecommand{\doi}[1]{DOI:\discretionary{}{}{}#1}\else
  \providecommand{\doi}{DOI:\discretionary{}{}{}\begingroup \urlstyle{rm}\Url}\fi

\bibitem[{Agarwal et~al.(2013)Agarwal, Tipaldi, Spinello, Stachniss and Burgard}]{Agarwal2013DCS}
Agarwal P, Tipaldi GD, Spinello L, Stachniss C and Burgard W (2013) Robust map optimization using dynamic covariance scaling.
\newblock In: \emph{IEEE International Conference on Robotics and Automation (ICRA)}. pp. 62--69.

\bibitem[{Ao et~al.(2021)Ao, Hu, Yang, Markham and Guo}]{9577271}
Ao S, Hu Q, Yang B, Markham A and Guo Y (2021) Spinnet: Learning a general surface descriptor for 3d point cloud registration.
\newblock In: \emph{IEEE/CVF Conference on Computer Vision and Pattern Recognition (CVPR)}. pp. 11748--11757.

\bibitem[{Arun et~al.(1987)Arun, Huang and Blostein}]{Arun1987Least}
Arun KS, Huang TS and Blostein SD (1987) Least-squares fitting of two 3-d point sets.
\newblock \emph{IEEE Transactions on Pattern Analysis and Machine Intelligence} 9(5): 698--700.

\bibitem[{Bai et~al.(2021)Bai, Luo, Zhou, Chen, Li, Hu, Fu and Tai}]{Bai2021Pointdsc}
Bai X, Luo Z, Zhou L, Chen H, Li L, Hu Z, Fu H and Tai CL (2021) Pointdsc: Robust point cloud registration using deep spatial consistency.
\newblock In: \emph{IEEE/CVF Conference on Computer Vision and Pattern Recognition (CVPR)}. pp. 15859--15869.

\bibitem[{Bai et~al.(2020)Bai, Luo, Zhou, Fu, Quan and Tai}]{Bai2020D3Feat}
Bai X, Luo Z, Zhou L, Fu H, Quan L and Tai CL (2020) D3feat: Joint learning of dense detection and description of 3d local features.
\newblock In: \emph{IEEE/CVF Conference on Computer Vision and Pattern Recognition (CVPR)}. pp. 6358--6366.

\bibitem[{Behley et~al.(2019)Behley, Garbade, Milioto, Quenzel, Behnke, Stachniss and Gall}]{Behley2019SemanticKITTI}
Behley J, Garbade M, Milioto A, Quenzel J, Behnke S, Stachniss C and Gall J (2019) {SemanticKITTI: A dataset for semantic scene understanding of LiDAR sequences}.
\newblock In: \emph{IEEE International Conference on Computer Vision (CVPR)}. pp. 9297--9307.

\bibitem[{Besl and McKay(1992)}]{Besl1992ICP}
Besl P and McKay ND (1992) A method for registration of 3-d shapes.
\newblock \emph{IEEE Transactions on Pattern Analysis and Machine Intelligence} 14(2): 239--256.

\bibitem[{Caesar et~al.(2020)Caesar, Bankiti, Lang, Vora, Liong, Xu, Krishnan, Pan, Baldan and Beijbom}]{caesar2020nuscenes}
Caesar H, Bankiti V, Lang AH, Vora S, Liong VE, Xu Q, Krishnan A, Pan Y, Baldan G and Beijbom O (2020) nuscenes: A multimodal dataset for autonomous driving.
\newblock In: \emph{IEEE/CVF conference on computer vision and pattern recognition (CVPR)}. pp. 11621--11631.

\bibitem[{Cattaneo et~al.(2022)Cattaneo, Vaghi and Valada}]{Cattaneo2022LCDNet}
Cattaneo D, Vaghi M and Valada A (2022) {LCDNet: Deep loop closure detection and point cloud registration for LiDAR SLAM}.
\newblock \emph{IEEE Transactions on Robotics} 38(4): 2074--2093.

\bibitem[{Censi(2008)}]{Censi2008point-to-line}
Censi A (2008) An icp variant using a point-to-line metric.
\newblock In: \emph{IEEE International Conference on Robotics and Automation (ICRA)}. pp. 19--25.

\bibitem[{Chen et~al.(2022)Chen, Wang, Xie, Zhai, Wang, Bao and Zhang}]{Chen2022VIP-SLAM}
Chen D, Wang S, Xie W, Zhai S, Wang N, Bao H and Zhang G (2022) Vip-slam: An efficient tightly-coupled rgb-d visual inertial planar slam.
\newblock In: \emph{International Conference on Robotics and Automation (ICRA)}. pp. 5615--5621.

\bibitem[{Chen et~al.(2020{\natexlab{a}})Chen, Nan, Xia, Zhao and Wonka}]{8936527}
Chen S, Nan L, Xia R, Zhao J and Wonka P (2020{\natexlab{a}}) Plade: A plane-based descriptor for point cloud registration with small overlap.
\newblock \emph{IEEE Transactions on Geoscience and Remote Sensing} 58(4): 2530--2540.

\bibitem[{Chen et~al.(2020{\natexlab{b}})Chen, Nardari, Lee, Qu, Liu, Romero and Kumar}]{Chen2020SLOAM}
Chen SW, Nardari GV, Lee ES, Qu C, Liu X, Romero RAF and Kumar V (2020{\natexlab{b}}) Sloam: Semantic lidar odometry and mapping for forest inventory.
\newblock \emph{IEEE Robotics and Automation Letters} 5(2): 612--619.

\bibitem[{Chen et~al.(2019)Chen, Milioto, Palazzolo, Giguère, Behley and Stachniss}]{chen2019SuMa++}
Chen X, Milioto A, Palazzolo E, Giguère P, Behley J and Stachniss C (2019) {SuMa++: Efficient LiDAR-based Semantic SLAM}.
\newblock In: \emph{IEEE/RSJ International Conference on Intelligent Robots and Systems (IROS)}. pp. 4530--4537.

\bibitem[{Choy et~al.(2020)Choy, Dong and Koltun}]{Choy2020DGR}
Choy C, Dong W and Koltun V (2020) Deep global registration.
\newblock In: \emph{IEEE/CVF conference on computer vision and pattern recognition (CVPR)}. pp. 2514--2523.

\bibitem[{Choy et~al.(2019)Choy, Park and Koltun}]{Choy2019FCGF}
Choy C, Park J and Koltun V (2019) Fully convolutional geometric features.
\newblock In: \emph{IEEE/CVF international conference on computer vision (CVPR)}. pp. 8958--8966.

\bibitem[{Cramariuc et~al.(2021)Cramariuc, Tschopp, Alatur, Benz, Falck, Brühlmeier, Hahn, Nieto and Siegwart}]{Cramariuc2021SemSegMap}
Cramariuc A, Tschopp F, Alatur N, Benz S, Falck T, Brühlmeier M, Hahn B, Nieto J and Siegwart R (2021) Semsegmap – 3d segment-based semantic localization.
\newblock In: \emph{IEEE/RSJ International Conference on Intelligent Robots and Systems (IROS)}. pp. 1183--1190.

\bibitem[{Dong et~al.(2018)Dong, Yang, Liang, Huang and Scherer}]{DONG201861}
Dong Z, Yang B, Liang F, Huang R and Scherer S (2018) Hierarchical registration of unordered tls point clouds based on binary shape context descriptor.
\newblock \emph{ISPRS Journal of Photogrammetry and Remote Sensing} 144: 61--79.

\bibitem[{Dubé et~al.(2017)Dubé, Dugas, Stumm, Nieto, Siegwart and Cadena}]{Dubé2017SegMatch}
Dubé R, Dugas D, Stumm E, Nieto J, Siegwart R and Cadena C (2017) Segmatch: Segment based place recognition in 3d point clouds.
\newblock In: \emph{IEEE International Conference on Robotics and Automation (ICRA)}. pp. 5266--5272.

\bibitem[{Fischler and Bolles(1981)}]{Fischler1981Random}
Fischler MA and Bolles RC (1981) Random sample consensus: a paradigm for model fitting with applications to image analysis and automated cartography.
\newblock \emph{Communications of the ACM} 24(6): 381–395.

\bibitem[{Geiger et~al.(2012)Geiger, Lenz and Urtasun}]{Geiger2012KITTI}
Geiger A, Lenz P and Urtasun R (2012) Are we ready for autonomous driving? the kitti vision benchmark suite.
\newblock In: \emph{IEEE Conference on Computer Vision and Pattern Recognition (CVPR)}. pp. 3354--3361.

\bibitem[{Geneva et~al.(2018)Geneva, Eckenhoff, Yang and Huang}]{8594463}
Geneva P, Eckenhoff K, Yang Y and Huang G (2018) Lips: Lidar-inertial 3d plane slam.
\newblock In: \emph{IEEE/RSJ International Conference on Intelligent Robots and Systems (IROS)}. pp. 123--130.

\bibitem[{Horn(1987)}]{Horn1987Closed}
Horn B (1987) Closed-form solution of absolute orientation using unit quaternions.
\newblock \emph{Journal of the Optical Society A} 4: 629--642.

\bibitem[{Hornung et~al.(2013)Hornung, Wurm, Bennewitz, Stachniss and Burgard}]{hornung2013octomap}
Hornung A, Wurm KM, Bennewitz M, Stachniss C and Burgard W (2013) Octomap: An efficient probabilistic 3d mapping framework based on octrees.
\newblock \emph{Autonomous robots} 34: 189--206.

\bibitem[{Hou et~al.(2022)Hou, Zhu, Ma, Loy and Li}]{Hou2022PVKD}
Hou Y, Zhu X, Ma Y, Loy CC and Li Y (2022) Point-to-voxel knowledge distillation for lidar semantic segmentation.
\newblock In: \emph{IEEE Conference on Computer Vision and Pattern Recognition (CVPR)}. pp. 8479--8488.

\bibitem[{Kong et~al.(2020)Kong, Yang, Zhai, Zhao, Zeng, Wang, Liu, Li and Wen}]{Kong2020Semantic}
Kong X, Yang X, Zhai G, Zhao X, Zeng X, Wang M, Liu Y, Li W and Wen F (2020) Semantic graph based place recognition for 3d point clouds.
\newblock In: \emph{IEEE/RSJ International Conference on Intelligent Robots and Systems (IROS)}. pp. 8216--8223.

\bibitem[{Li et~al.(2019)Li, Sung, Dubrovina, Yi and Guibas}]{Lingxiao2019Supervised}
Li L, Sung M, Dubrovina A, Yi L and Guibas LJ (2019) Supervised fitting of geometric primitives to 3d point clouds.
\newblock In: \emph{IEEE/CVF International Conference on Computer Vision and Pattern Recognition (CVPR)}. pp. 2652--2660.

\bibitem[{Liao et~al.(2022)Liao, Xie and Geiger}]{Yiyi2022KITTI-360}
Liao Y, Xie J and Geiger A (2022) {KITTI}-360: A novel dataset and benchmarks for urban scene understanding in 2d and 3d.
\newblock \emph{IEEE Transactions on Pattern Analysis and Machine Intelligence} 45(3): 3292--3310.

\bibitem[{Lim et~al.(2024)Lim, Kim, Kim, Mason~Lee and Myung}]{Lim2024quatro++}
Lim H, Kim B, Kim D, Mason~Lee E and Myung H (2024) Quatro++: Robust global registration exploiting ground segmentation for loop closing in lidar slam.
\newblock \emph{The International Journal of Robotics Research} 43(5): 685--715.

\bibitem[{Lim et~al.(2022)Lim, Yeon, Ryu, Lee, Kim, Yun, Jung, Lee and Myung}]{Lim2022Quatro}
Lim H, Yeon S, Ryu S, Lee Y, Kim Y, Yun J, Jung E, Lee D and Myung H (2022) A single correspondence is enough: Robust global registration to avoid degeneracy in urban environments.
\newblock In: \emph{2022 International Conference on Robotics and Automation (ICRA)}. IEEE, pp. 8010--8017.

\bibitem[{Lin et~al.(2023)Lin, Yuan, Cai, Li, Ren, Zou, Hong and Zhang}]{Lin2023ImMesh}
Lin J, Yuan C, Cai Y, Li H, Ren Y, Zou Y, Hong X and Zhang F (2023) Immesh: An immediate lidar localization and meshing framework.
\newblock \emph{IEEE Transactions on Robotics} 39(6): 4312--4331.

\bibitem[{Low(2004)}]{low2004linear}
Low KL (2004) Linear least-squares optimization for point-to-plane icp surface registration.
\newblock \emph{Chapel Hill, University of North Carolina} 4(10): 1--3.

\bibitem[{Lu et~al.(2019)Lu, Zhou, Wan, Hou and Song}]{Lu2019L3-netApollo}
Lu W, Zhou Y, Wan G, Hou S and Song S (2019) L3-net: Towards learning based lidar localization for autonomous driving.
\newblock In: \emph{IEEE/CVF conference on computer vision and pattern recognition (CVPR)}. pp. 6389--6398.

\bibitem[{Lusk et~al.(2021)Lusk, Fathian and How}]{lusk2021clipper}
Lusk PC, Fathian K and How JP (2021) Clipper: A graph-theoretic framework for robust data association.
\newblock In: \emph{IEEE International Conference on Robotics and Automation (ICRA)}. IEEE, pp. 13828--13834.

\bibitem[{Nicholson et~al.(2018)Nicholson, Milford and S{\"u}nderhauf}]{nicholson2018quadricslam}
Nicholson L, Milford M and S{\"u}nderhauf N (2018) Quadricslam: Dual quadrics from object detections as landmarks in object-oriented slam.
\newblock \emph{IEEE Robotics and Automation Letters} 4(1): 1--8.

\bibitem[{Oh et~al.(2022)Oh, Jung, Lim, Song, Hu, Lee, Park, Kim, Lee and Myung}]{Oh2022TRAVEL}
Oh M, Jung E, Lim H, Song W, Hu S, Lee EM, Park J, Kim J, Lee J and Myung H (2022) Travel: Traversable ground and above-ground object segmentation using graph representation of 3d lidar scans.
\newblock \emph{IEEE Robotics and Automation Letters} 7(3): 7255--7262.

\bibitem[{Olsson et~al.(2009)Olsson, Kahl and Oskarsson}]{Olsson2009Branch}
Olsson C, Kahl F and Oskarsson M (2009) Branch-and-bound methods for euclidean registration problems.
\newblock \emph{IEEE Transactions on Pattern Analysis and Machine Intelligence} 31(5): 783--794.

\bibitem[{Papazov et~al.(2012)Papazov, Haddadin, Parusel, Krieger and Burschka}]{Chavdar2012Rigid}
Papazov C, Haddadin S, Parusel S, Krieger K and Burschka D (2012) Rigid 3d geometry matching for grasping of known objects in cluttered scenes.
\newblock \emph{The International Journal of Robotics Research} 31(4): 538--553.

\bibitem[{Poiesi and Boscaini(2023)}]{9775606}
Poiesi F and Boscaini D (2023) Learning general and distinctive 3d local deep descriptors for point cloud registration.
\newblock \emph{IEEE Transactions on Pattern Analysis and Machine Intelligence} 45(3): 3979--3985.

\bibitem[{Pramatarov et~al.(2022)Pramatarov, De~Martini, Gadd and Newman}]{Pramatarov2022BoxGraph}
Pramatarov G, De~Martini D, Gadd M and Newman P (2022) Boxgraph: Semantic place recognition and pose estimation from 3d lidar.
\newblock In: \emph{IEEE/RSJ International Conference on Intelligent Robots and Systems (IROS)}. pp. 7004--7011.

\bibitem[{Prokop et~al.(2020)Prokop, Shaikh and Kim}]{rs12010061}
Prokop M, Shaikh SA and Kim KS (2020) Low overlapping point cloud registration using line features detection.
\newblock \emph{Remote Sensing} 12(1).

\bibitem[{Qiao et~al.(2024)Qiao, Yu, Jiang, Yin and Shen}]{Qiao2024G3reg}
Qiao Z, Yu Z, Jiang B, Yin H and Shen S (2024) G3reg: Pyramid graph-based global registration using gaussian ellipsoid model.
\newblock \emph{IEEE Transactions on Automation Science and Engineering} : 1--17.

\bibitem[{Qiao et~al.(2023)Qiao, Yu, Yin and Shen}]{Qiao2023Pyramid}
Qiao Z, Yu Z, Yin H and Shen S (2023) Pyramid semantic graph-based global point cloud registration with low overlap.
\newblock In: \emph{IEEE/RSJ International Conference on Intelligent Robots and Systems (IROS)}. IEEE, pp. 11202--11209.

\bibitem[{Rosinol et~al.(2021)Rosinol, Violette, Abate, Hughes, Chang, Shi, Gupta and Carlone}]{Antoni2021Kimera}
Rosinol A, Violette A, Abate M, Hughes N, Chang Y, Shi J, Gupta A and Carlone L (2021) Kimera: From slam to spatial perception with 3d dynamic scene graphs.
\newblock \emph{The International Journal of Robotics Research} 40(12-14): 1510--1546.

\bibitem[{Rossi et~al.(2015)Rossi, Gleich and Gebremedhin}]{Rossi2015PMC}
Rossi RA, Gleich DF and Gebremedhin AH (2015) Parallel maximum clique algorithms with applications to network analysis.
\newblock \emph{SIAM Journal on Scientific Computing} 37(5): C589--C616.

\bibitem[{Ruan et~al.(2023)Ruan, Li, Wang and Sun}]{Ruan2023SLAMesh}
Ruan J, Li B, Wang Y and Sun Y (2023) Slamesh: Real-time lidar simultaneous localization and meshing.
\newblock In: \emph{IEEE International Conference on Robotics and Automation (ICRA)}. pp. 3546--3552.

\bibitem[{Rusu et~al.(2009)Rusu, Blodow and Beetz}]{Rusu2009FPFH}
Rusu RB, Blodow N and Beetz M (2009) Fast point feature histograms (fpfh) for 3d registration.
\newblock In: \emph{IEEE International Conference on Robotics and Automation (ICRA)}. pp. 3212--3217.

\bibitem[{Rusu et~al.(2008)Rusu, Blodow, Marton and Beetz}]{rusu2008aligning}
Rusu RB, Blodow N, Marton ZC and Beetz M (2008) Aligning point cloud views using persistent feature histograms.
\newblock In: \emph{IEEE/RSJ international conference on intelligent robots and systems (IROS)}. IEEE, pp. 3384--3391.

\bibitem[{Salti et~al.(2014)Salti, Tombari and Di~Stefano}]{salti2014shot}
Salti S, Tombari F and Di~Stefano L (2014) Shot: Unique signatures of histograms for surface and texture description.
\newblock \emph{Computer Vision and Image Understanding} 125: 251--264.

\bibitem[{Schnabel et~al.(2007)Schnabel, Wahl and Klein}]{schnabel2007efficient}
Schnabel R, Wahl R and Klein R (2007) Efficient ransac for point-cloud shape detection.
\newblock In: \emph{Computer graphics forum}, volume~26. pp. 214--226.

\bibitem[{Shan et~al.(2020)Shan, Englot, Meyers, Wang, Ratti and Daniela}]{liosam2020shan}
Shan T, Englot B, Meyers D, Wang W, Ratti C and Daniela R (2020) Lio-sam: Tightly-coupled lidar inertial odometry via smoothing and mapping.
\newblock In: \emph{IEEE/RSJ International Conference on Intelligent Robots and Systems (IROS)}. pp. 5135--5142.

\bibitem[{Sharma et~al.(2020)Sharma, Liu, Maji, Kalogerakis, Chaudhuri and Mech}]{Gopal2020ParSeNet}
Sharma G, Liu D, Maji S, Kalogerakis E, Chaudhuri S and Mech R (2020) Parsenet: A parametric surface fitting network for 3d point clouds.
\newblock In: \emph{European Conference on Computer Vision (ECCV)}, volume 12352. pp. 261--276.

\bibitem[{Shiratori et~al.(2015)Shiratori, Berclaz, Harville, Shah, Li, Matsushita and Shiller}]{shiratori2015efficient}
Shiratori T, Berclaz J, Harville M, Shah C, Li T, Matsushita Y and Shiller S (2015) Efficient large-scale point cloud registration using loop closures.
\newblock In: \emph{International Conference on 3D Vision}. pp. 232--240.

\bibitem[{Sun et~al.(2020)Sun, Kretzschmar, Dotiwalla, Chouard, Patnaik, Tsui, Guo, Zhou, Chai, Caine et~al.}]{Sun2020ScalabilityWaymo}
Sun P, Kretzschmar H, Dotiwalla X, Chouard A, Patnaik V, Tsui P, Guo J, Zhou Y, Chai Y, Caine B et~al. (2020) Scalability in perception for autonomous driving: Waymo open dataset.
\newblock In: \emph{IEEE/CVF conference on computer vision and pattern recognition (CVPR)}. pp. 2446--2454.

\bibitem[{Taubin(1991)}]{Taubin1991Estimation}
Taubin G (1991) Estimation of planar curves, surfaces, and nonplanar space curves defined by implicit equations with applications to edge and range image segmentation.
\newblock \emph{IEEE Transactions on Pattern Analysis and Machine Intelligence} 13(11): 1115--1138.

\bibitem[{Tombari et~al.(2010)Tombari, Salti and Di~Stefano}]{Tombari2010Unique}
Tombari F, Salti S and Di~Stefano L (2010) Unique shape context for 3d data description.
\newblock In: \emph{ACM Workshop on 3D Object Retrieval}. p. 57–62.

\bibitem[{Wang et~al.(2023)Wang, Liu, Hu, Wang, Chen, Dong, Guo, Wang and Yang}]{wang2023roreg}
Wang H, Liu Y, Hu Q, Wang B, Chen J, Dong Z, Guo Y, Wang W and Yang B (2023) Roreg: Pairwise point cloud registration with oriented descriptors and local rotations.
\newblock \emph{IEEE Transactions on Pattern Analysis and Machine Intelligence} 45(8): 10376--10393.

\bibitem[{Wu et~al.(2024)Wu, Yu, Yang and Xia}]{Wu2024QuadricsNet}
Wu J, Yu H, Yang W and Xia GS (2024) Quadricsnet: Learning concise representation for geometric primitives in point clouds.
\newblock In: \emph{IEEE International Conference on Robotics and Automation (ICRA)}. pp. 4060--4066.

\bibitem[{Xia et~al.(2023)Xia, Xu, Rim, Ding, Zheng, Keutzer, Tomizuka and Zhan}]{Xia2023Quadric}
Xia C, Xu C, Rim P, Ding M, Zheng N, Keutzer K, Tomizuka M and Zhan W (2023) Quadric representations for lidar odometry, mapping and localization.
\newblock \emph{IEEE Robotics and Automation Letters} 8(8): 5023--5030.

\bibitem[{Xu et~al.(2022)Xu, Cai, He, Lin and Zhang}]{Xu2022FAST-LIO2}
Xu W, Cai Y, He D, Lin J and Zhang F (2022) Fast-lio2: Fast direct lidar-inertial odometry.
\newblock \emph{IEEE Transactions on Robotics} 38(4): 2053--2073.

\bibitem[{Yang et~al.(2020)Yang, Antonante, Tzoumas and Carlone}]{Yang2020Graduated}
Yang H, Antonante P, Tzoumas V and Carlone L (2020) Graduated non-convexity for robust spatial perception: From non-minimal solvers to global outlier rejection.
\newblock \emph{IEEE Robotics and Automation Letters} 5(2): 1127--1134.

\bibitem[{Yang et~al.(2021)Yang, Shi and Carlone}]{Yang2021TEASER}
Yang H, Shi J and Carlone L (2021) Teaser: Fast and certifiable point cloud registration.
\newblock \emph{IEEE Transactions on Robotics} 37(2): 314--333.

\bibitem[{Yang et~al.(2024)Yang, Zhang, Wang, Guo, Sun, Wu, Zhang and Zhang}]{Yang2024MAC}
Yang J, Zhang X, Wang P, Guo Y, Sun K, Wu Q, Zhang S and Zhang Y (2024) Mac: Maximal cliques for 3d registration.
\newblock \emph{IEEE Transactions on Pattern Analysis and Machine Intelligence} 46(12): 10645--10662.

\bibitem[{Yang and Scherer(2019)}]{Yang2019CubeSLAM}
Yang S and Scherer S (2019) Cubeslam: Monocular 3-d object slam.
\newblock \emph{IEEE Transactions on Robotics} 35(4): 925--938.

\bibitem[{Yew and Lee(2020)}]{9157132}
Yew ZJ and Lee GH (2020) Rpm-net: Robust point matching using learned features.
\newblock In: \emph{IEEE/CVF Conference on Computer Vision and Pattern Recognition (CVPR)}. pp. 11821--11830.

\bibitem[{Yin et~al.(2024)Yin, Xu, Lu, Chen, Xiong, Shen, Stachniss and Wang}]{Yin2024}
Yin H, Xu X, Lu S, Chen X, Xiong R, Shen S, Stachniss C and Wang Y (2024) A survey on global lidar localization: Challenges, advances and open problems.
\newblock \emph{International Journal of Computer Vision} 132: 3139--3171.

\bibitem[{Yin et~al.(2023)Yin, Yuan, Cao, Ji, Zhang and Xie}]{Yin2023Segregator}
Yin P, Yuan S, Cao H, Ji X, Zhang S and Xie L (2023) Segregator: Global point cloud registration with semantic and geometric cues.
\newblock In: \emph{2023 IEEE International Conference on Robotics and Automation (ICRA)}. pp. 2848--2854.

\bibitem[{Yu et~al.(2020)Yu, Zhen, Yang, Zhang and Scherer}]{yu2020monocular}
Yu H, Zhen W, Yang W, Zhang J and Scherer S (2020) Monocular camera localization in prior lidar maps with 2d-3d line correspondences.
\newblock In: \emph{IEEE/RSJ International Conference on Intelligent Robots and Systems (IROS)}. IEEE, pp. 4588--4594.

\bibitem[{Zhang and Singh(2014)}]{zhang2014loam}
Zhang J and Singh S (2014) Loam: Lidar odometry and mapping in real-time.
\newblock In: \emph{Robotics: Science and Systems (RSS)}, volume~2. pp. 1--9.

\bibitem[{Zhang et~al.(2023)Zhang, Yang, Zhang and Zhang}]{Zhang20233DMAC}
Zhang X, Yang J, Zhang S and Zhang Y (2023) 3d registration with maximal cliques.
\newblock In: \emph{IEEE/CVF Conference on Computer Vision and Pattern Recognition (CVPR)}. pp. 17745--17754.

\bibitem[{Zhen et~al.(2022)Zhen, Yu, Hu and Scherer}]{Zhen2022Unified}
Zhen W, Yu H, Hu Y and Scherer S (2022) Unified representation of geometric primitives for graph-slam optimization using decomposed quadrics.
\newblock In: \emph{International Conference on Robotics and Automation (ICRA)}. pp. 5636--5642.

\bibitem[{Zhou et~al.(2022)Zhou, Huang, Mao, Yu, Wang and Kaess}]{9787712}
Zhou L, Huang G, Mao Y, Yu J, Wang S and Kaess M (2022) $\mathcal{PLC}$-lislam: Lidar slam with planes, lines, and cylinders.
\newblock \emph{IEEE Robotics and Automation Letters} 7(3): 7163--7170.

\bibitem[{Zhou et~al.(2016)Zhou, Park and Koltun}]{zhou2016fast}
Zhou QY, Park J and Koltun V (2016) Fast global registration.
\newblock In: \emph{European Conference on Computer Vision (ECCV)}. pp. 766--782.

\bibitem[{Zhou et~al.(2018)Zhou, Park and Koltun}]{Qian-Yi2018Open3D}
Zhou QY, Park J and Koltun V (2018) {Open3D}: {A} modern library for {3D} data processing.
\newblock \emph{arXiv:1801.09847} .

\bibitem[{Zhou et~al.(2023)Zhou, Feng, Di and Zhou}]{zhou2023lidar}
Zhou Z, Feng X, Di S and Zhou X (2023) A lidar mapping system for robot navigation in dynamic environments.
\newblock \emph{IEEE Transactions on Intelligent Vehicles} .

\bibitem[{Zuo et~al.(2017)Zuo, Xie, Liu and Huang}]{zuo2017robust}
Zuo X, Xie X, Liu Y and Huang G (2017) Robust visual slam with point and line features.
\newblock In: \emph{IEEE/RSJ International Conference on Intelligent Robots and Systems (IROS)}. IEEE, pp. 1775--1782.

\end{thebibliography}
}

\end{document}